\documentclass[10pt,twocolumn,letterpaper]{article}

\usepackage[pagenumbers]{cvpr} 

\usepackage{amsthm}
\usepackage{multirow}
\usepackage{algorithm}
\usepackage{algpseudocode}
\usepackage{tikz}
\usetikzlibrary{arrows.meta,positioning,calc,fit,backgrounds}
\usepackage{pgfplots}
\pgfplotsset{compat=1.18}
\usepgfplotslibrary{groupplots}

\newcommand{\method}{DGSP\xspace}

\newcommand{\draft}{\mathcal{C}}            
\newcommand{\full}{\mathcal{F}}             
\newcommand{\merged}{\mathcal{M}}           
\newcommand{\Sup}{\mathsf{Sup}}   
\newcommand{\field}{\mathcal{Z}}            
\newcommand{\foot}{\mathcal{R}_{\field}}    
\newcommand{\supcrops}{\mathcal{R}_{\Sup}}  
\newcommand{\detL}{\mathsf{Det}^{L}} 
\newcommand{\detH}{\mathsf{Det}^{H}} 
\newcommand{\policy}{\pi}         
\newcommand{\dem}{\mathsf{d}}              
\newcommand{\vclose}{v_{\mathrm{close}}}
\newcommand{\scene}{\mathcal{S}}       
\newcommand{\reg}{\mathcal{R}}         
\newcommand{\state}{\mathbf{x}}        
\newcommand{\img}{\mathbf{I}}          
\newcommand{\ver}{\mathcal{V}}          

\newtheorem{theorem}{Theorem}
\newtheorem{corollary}{Corollary}
\newtheorem{lemma}{Lemma}

\theoremstyle{definition}
\newtheorem{definition}{Definition}
\theoremstyle{remark}
\newtheorem{remark}{Remark}

\usepackage{mathtools}
\usepackage{float}
\usepackage{placeins}
\theoremstyle{plain}
\newtheorem*{restatedinner}{\restatedname}
\newcommand{\restatedname}{}
\newenvironment{restated}[3]{\renewcommand{\restatedname}{#1~#2}\begin{restatedinner}[#3]}{\end{restatedinner}}
\theoremstyle{definition}
\newtheorem*{restateddefinner}{\restatedname}
\newenvironment{restateddef}[3]{\renewcommand{\restatedname}{#1~#2}\begin{restateddefinner}[#3]}{\end{restateddefinner}}
\theoremstyle{plain}
\newenvironment{strategy}{\par\noindent\textit{Proof strategy.}\ }{\par}

\usepackage{colortbl}
\definecolor{oursbg}{RGB}{232,240,252}   
\definecolor{gaingreen}{RGB}{0,128,64}   
\newcommand{\ci}[2]{[#1,\,#2]}

\makeatletter 
\newcommand{\oursfirst}{\gdef\CT@cell@color{\CT@color{oursbg}\@tempdimb\z@\global\let\CT@cell@color\relax}}
\newcommand{\ourslast}{\gdef\CT@cell@color{\CT@color{oursbg}\@tempdimc\z@\global\let\CT@cell@color\relax}}
\makeatother
\newcommand{\oursrow}{\rowcolor{oursbg}\oursfirst}
\makeatletter\newcommand{\oursfull}{\gdef\CT@cell@color{\CT@color{oursbg}\@tempdimb\col@sep\@tempdimc\col@sep\global\let\CT@cell@color\relax}}\makeatother 
\newcolumntype{g}{>{\columncolor{gray!10}}c}   
\newcolumntype{G}{>{\columncolor{gray!22}[\tabcolsep][0pt]}c} 
\newcolumntype{M}{>{\columncolor{gray!10}}c} 
\newcolumntype{N}{>{\columncolor{gray!10}[0pt][\tabcolsep]}c} 
\newcommand{\na}{--}
\newcommand{\rf}[1]{#1} 
\newcommand{\sgu}{\ensuremath{^{\textcolor{black!45}{\dagger}}}} 
\newcommand{\sgd}{\ensuremath{^{\textcolor{black!45}{\ast}}}} 
\newcommand{\up}{\ensuremath{\uparrow}}\newcommand{\dn}{\ensuremath{\downarrow}}
\newcommand{\rankkey}{\textbf{Best}, \underline{second}}
\definecolor{rkA}{RGB}{250,214,105}\definecolor{rkB}{RGB}{214,216,222}\definecolor{rkC}{RGB}{233,188,150} 
\newcommand{\fst}[1]{\textbf{#1}}\newcommand{\scd}[1]{\underline{#1}}\newcommand{\thd}[1]{#1} 

\definecolor{oursc}{RGB}{200,40,40}
\definecolor{unic}{RGB}{60,60,60}
\definecolor{adac}{RGB}{31,119,180}
\definecolor{cropc}{RGB}{44,160,44}
\definecolor{fieldc}{RGB}{240,150,60}

\definecolor{cvprblue}{rgb}{0.21,0.49,0.74}
\usepackage[pagebackref,breaklinks,colorlinks,allcolors=cvprblue]{hyperref}
\AtBeginDocument{}
\newcommand{\supp}[1]{Appendix~\ref{S-#1}} 
\AtEndPreamble{%
\Crefname{assumption}{Assumption}{Assumptions}%
\Crefname{proposition}{Proposition}{Propositions}%
\Crefname{theorem}{Theorem}{Theorems}%
\Crefname{corollary}{Corollary}{Corollaries}%
\Crefname{lemma}{Lemma}{Lemmas}%
\Crefname{definition}{Definition}{Definitions}%
\Crefname{algorithm}{Algorithm}{Algorithms}%
\crefname{lemma}{Lemma}{Lemmas}\crefname{remark}{Remark}{Remarks}\Crefname{remark}{Remark}{Remarks}%
}

\def\confName{CVPR}
\def\confYear{2027}

\title{Check Your Blind Spots: Decision-Guided Perception\\by Probing the Planner with Hypothetical Objects}

\author{Hoang Pham Cong\textsuperscript{1}\qquad Long D. Nguyen\textsuperscript{2}\qquad Ho Viet Duc Luong\textsuperscript{1}\\
Nguyen Tran Minh Nhat\textsuperscript{1}\qquad Binh P. Nguyen\textsuperscript{2}\\[0.6em]
\textsuperscript{1}Hanoi University of Science and Technology\qquad
\textsuperscript{2}Victoria University of Wellington
}
\hypersetup{pdftitle={Check Your Blind Spots: Decision-Guided Perception by Probing the Planner with Hypothetical Objects},pdfauthor={Hoang Pham Cong; Long D. Nguyen; Ho Viet Duc Luong; Nguyen Tran Minh Nhat; Binh P. Nguyen}}

\begin{document}
\setlength{\textfloatsep}{12pt plus 2pt minus 4pt}\setlength{\dbltextfloatsep}{12pt plus 2pt minus 4pt}
\makeatletter\@clubpenalty=10000\makeatother\clubpenalty=10000 \widowpenalty=10000 
\setcounter{dbltopnumber}{1} 
\hyphenation{Table Tables Figure Figures Section Sections Equation Corollary Theorem} 
\maketitle

\begin{abstract}
Not every perception error changes the planner's action.
Yet adaptive-resolution perception chooses where to spend high-resolution computation from image cues, which can neither reliably point to an object that low resolution missed nor tell which objects matter to the planner.
We observe that the planner can say where an object would matter before anything is seen: inserting a hypothetical object at a location tests whether its action would change.
Decision-Guided Speculative Perception (\method) treats the low-resolution scene and its action as a draft and probes the planner in this way to obtain a decision-sensitivity field, the locations where an unseen object would change the action.
Only this field and the support, the detections on which the current action rests, are verified at high resolution, repeatedly until the action is stable.
For two planner classes, we prove that \method returns exactly the action of whole-frame high-resolution perception if crop detections match whole-frame ones and the crops cover the field, conditions that real images meet approximately.
On three driving datasets, it reproduces this action on about 98\% of frames with about a quarter of the pixels. At the pixel cost of uniform downscaling, it repairs more of the draft's disagreements with this action, brakes falsely less often, and recovers many more of the decision-relevant objects that the draft misses.
\par\smallskip\noindent Code: \url{https://github.com/kongwoang/DGSP}
\end{abstract}

\section{Introduction}
\label{sec:intro}

\begin{figure}[t]
\centering
\includegraphics[width=\linewidth]{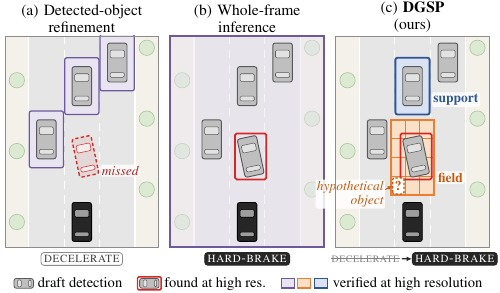}
\caption{\textbf{Probing the planner shows where high resolution is needed.}
Schematic top view: a low-resolution draft detects a lead vehicle (\textsc{decelerate}) but misses a cut-in vehicle requiring \textsc{hard-brake} (dashed red; its location is given to no method).
(a)~Detected-object refinement leaves it unseen.
(b)~Whole-frame inference detects it at full-image cost.
(c)~\method probes the planner with hypothetical objects and verifies the image regions of the field (orange) and support (blue), recovering \textsc{hard-brake}.}
\label{fig:teaser}
\end{figure}

\looseness=-1 An autonomous vehicle perceives its surroundings in order to act on them.
Megapixel cameras let it detect small and distant objects~\cite{nuscenes,av2}, but running a detector on every frame at native resolution strains the compute budget of onboard hardware.
Adaptive-resolution methods therefore spend high resolution selectively: on image regions chosen by object priors~\cite{fovea,lzu}, learned saliency~\cite{esod,zoomdet}, or coarse detections~\cite{dzn,ufpmp}, or on whole frames, by adapting the input scale~\cite{adascale}, the detector configuration~\cite{chanakya}, or the choice of model~\cite{bgtada}.
Their signals come from the image or the system state, not from the planner, which limits them in two ways.
They cannot reliably point to what low resolution failed to show: a missed object leaves no detection to refine, and locating it would require the very inference we try to avoid.
Nor do they know what matters: an object far from the vehicle's path can receive as much computation as one forcing it to brake.
In \Cref{fig:teaser}(a), detected-object refinement sharpens the vehicles it already sees and leaves a nearer cut-in vehicle unseen, retaining \textsc{decelerate} where \textsc{hard-brake} is required; whole-frame inference in \Cref{fig:teaser}(b) removes this blind spot only at full cost.

A human driver resolves this differently.
Before changing lanes, a driver checks the blind spot, not because a car is likely to be there, but because a car there would change what the driver does; when driving straight ahead, the same spot goes unchecked.
The intended action thus decides where to look, before anything is seen.
Our key observation is that a planner can make this decision explicitly: whether an object at a given location would change the action is a question about the planner, not about the image, and the planner can answer it for a \emph{hypothetical} object placed there.
The choice of where to perceive thereby moves from the image to the planner, and an unseen object need not be found in order to know where it would matter.

We turn this insight into \textbf{D}ecision-\textbf{G}uided \textbf{S}pec\-u\-la\-tive \textbf{P}er\-cep\-tion (\method), which treats the scene and action of low-resolution perception as a draft and verifies only what could overturn that action.
Probing the planner with hypothetical objects yields a \emph{decision-sensitivity field}, the locations where an unseen object would change the action (orange in \Cref{fig:teaser}(c)); it covers the cut-in vehicle although the draft never detected it.
Because the draft can also be wrong about what it does show, \method additionally identifies a \emph{support}, the detections on which the current action rests (blue).
Only these regions are processed at high resolution, and the check repeats until the action is stable.

Deriving the regions from the planner also changes what can be guaranteed: the target is no longer detection accuracy, but the reference action that whole-frame high-resolution perception would produce, and we prove that, under explicit conditions, \method returns exactly this action for certain braking controllers and fixed-trajectory planners.
Experiments on nuScenes~\cite{nuscenes}, Argoverse~2~\cite{av2}, and Argoverse-HD~\cite{streaming} test this on real images: with a 960-pixel reference, \method matches its action on 97.5--98.7\% of frames with 22--27\% of its detector-input pixels (6--8\% of the native frame).
The saving is not that of a lower resolution: at the pixel cost of uniform downscaling, \method repairs 7--9 points more of the draft's disagreements with the reference and raises braking F1 by 6--7 points, an advantage that grows with the range the planner must cover.

\noindent\textbf{Contributions. }Our contributions are as follows:
\begin{itemize}[leftmargin=*,itemsep=1pt,topsep=2pt]
  \item We propose to decide where to perceive by querying the planner, not the image: probing it with hypothetical objects yields, without detections, a decision-sensitivity field marking where unseen objects change the action.
  \item We develop an iterative high-resolution verification algorithm that combines this field with a support set of detections sufficient to preserve the current action, and prove that, under explicit conditions, it returns exactly the reference action for certain braking controllers and fixed-trajectory planners.
  \item On three datasets, \method reproduces the reference action at about a quarter of its pixels and applies to other planners, even a black-box one, and to closed-loop braking.
\end{itemize}

\section{Related Work}
\label{sec:related}

\noindent\textbf{Spatially selective detection.}
These methods concentrate high-resolution computation on selected image locations.
They refine regions chosen by learned policies~\cite{dzn,patchdrop}, cluster proposals~\cite{clusdet}, or coarse detections~\cite{ufpmp}, process high-resolution features sparsely at predicted object locations~\cite{querydet,esod}, or magnify salient areas using object priors, previous detections, or learned offsets~\cite{fovea,lzu,zoomdet}; SAHI~\cite{sahi} instead slices the image into overlapping tiles.
\method shares their premise that most pixels need no high resolution, but not their selection signal: visual and detection cues are absent where low resolution misses objects.

\noindent\textbf{Frame-level adaptive inference.}
These methods apply one setting to the whole frame at run time: its input resolution~\cite{adascale}, the detector configuration~\cite{chanakya}, or the choice of model~\cite{bgtada}; recent work also adapts to runtime slack~\cite{slackdrive} or LiDAR scene complexity~\cite{thinkasneeded}.
They trade accuracy against cost per frame, not per decision-relevant region.

\noindent\textbf{Decision relevance in driving.}
Planning-oriented models such as UniAD~\cite{uniad} and PlanT~\cite{plant} organize scene representations around downstream planning, and planner-centric metrics~\cite{pkl,tip}, decision-value analysis~\cite{deep}, and reachability-based perception zones~\cite{perceptionzones,maneuverzones} evaluate perception offline by the decisions its errors affect.
\method applies this criterion at run time, to allocate computation, in the spirit of the value of information~\cite{howard}.
Object-importance studies~\cite{goaloie,droid} identify agents that influence driving, and counterfactual importance~\cite{cfimportance} removes or perturbs objects in the scene; our support plays this role for detections, but removal cannot point to an undetected object, the case our field addresses.
Hypothetical objects have precedents: PKL~\cite{pkl} inserts false positives on a grid to map planner sensitivity offline, occlusion-aware planners place phantom agents where nothing is observed~\cite{hubmann,occlusionblind}, and a learned relevance model scores objects on a grid to guide sensing~\cite{cfrelevance}; \method asks the planner itself where to look.

\noindent\textbf{Planner-guided computation.}
Closest to our work are systems in which planning objectives direct perception, an idea dating back to robot driving~\cite{reece}.
AnA~\cite{ana} directs perception to agents of interest that the system already tracks, QuAD~\cite{quad} queries occupancy at locations relevant to candidate plans, VeteranAD~\cite{veteranad} gathers scene information along anchored trajectories, PAD~\cite{pad} learns spatial attention for planning, and DriveAgent-R1~\cite{driveagentr1} actively requests more visual evidence.
There, relevance follows tracked agents, candidate plans, or learned attention. \method defines it counterfactually, as the object states whose insertion would change the current action, adds the detections sustaining it, and targets a reference, the action of whole-frame high-resolution perception, which it provably reproduces under explicit conditions, like speculative decoding~\cite{specdec}.

\section{Decision-Guided Speculative Perception}
\label{sec:method}

\begin{figure*}[t]
\centering
\includegraphics[width=0.93\textwidth]{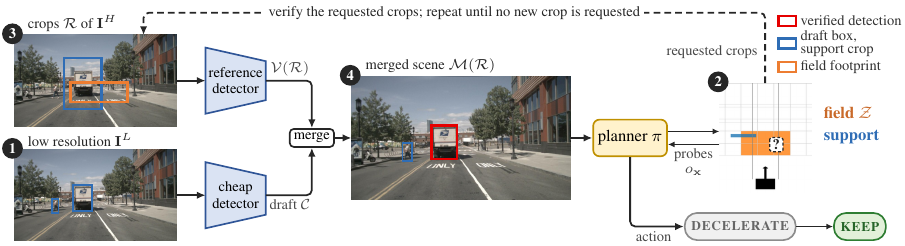}
\caption{\textbf{Overview of \method on a nuScenes frame.}
(1)~The cheap detector finds a truck and a pedestrian; the truck demands \textsc{decelerate} and forms the support.
(2)~Probing the planner with hypothetical objects (``?'')\ gives the field (orange): at this ego speed, empty at \textsc{decelerate} and a short corridor at \textsc{keep}.
(3)~The reference detector runs on the requested crops only: first the support crop, where the truck closes more slowly than the draft estimated, so the action drops to \textsc{keep}; then the field footprint that \textsc{keep} opens (guard crops not shown).
(4)~No new crop is requested; the planner returns the required \textsc{keep}. \Cref{fig:qual}(a) shows the field recovering a missed object.}
\label{fig:method}
\end{figure*}

\subsection{Problem formulation}
\label{sec:setting}

Scenes differing in many detections can share an action, whereas one missed object can change it.
We therefore aim for \emph{decision fidelity}: reproducing the action of whole-frame high-resolution perception with less detector computation.

\noindent\textbf{Objects and planner.}
We treat one frame $\img$ at a time, with all boxes and crops in its native pixel coordinates.
A detected object $o$ has an image box and a state $\state(o)$---its range, lateral extent, and time to collision in the ego frame---which a \emph{lift} estimates from the box, the calibration, and earlier frames.
A \emph{scene} $\scene$ is a finite set of objects.
A \emph{planner} $\policy$ is any deterministic downstream decision rule that maps the object states of a scene to an action for the fixed ego state, from a braking controller to a selector over fixed trajectories; our guarantees cover specific classes only.

\noindent\textbf{Draft and reference.}
Let $\img^L$ and $\img^H$ be copies of $\img$ at a low resolution and at the higher \emph{verification resolution}, which may lie below native.
For a region $w$ of the frame, a crop or the whole frame $\img$, let $\detL(w)$ and $\detH(w)$ denote the lifted objects that the \emph{cheap} and the \emph{reference} detector return on the parts of $\img^L$ and $\img^H$ that show $w$, respectively.
The whole frame, $w=\img$, gives the \emph{draft} scene $\draft=\detL(\img)$ and the \emph{reference} scene $\full=\detH(\img)$, with \emph{reference action} $\policy(\full)$.
$\full$ and $\policy(\full)$ define the goal, whatever the quality of $\full$, and are not computed at run time.

\noindent\textbf{Verification.}
A \emph{crop} is an axis-aligned image rectangle, a \emph{verified region} $\reg$ is a finite set of crops, and $\scene|_{\reg}$ denotes the objects of $\scene$ with box centers in $\reg$.
Verifying $\reg$ runs the reference detector on its crops only: $\ver(\reg)=\bigcup_{w\in \reg}\detH(w)$, with duplicates from overlapping crops merged.
In general, $\ver(\reg)\neq\full|_{\reg}$, the part of the reference scene in $\reg$: detection does not commute with cropping, since a crop changes the context and truncates boxes at its border.
The detections $\ver(\reg)$ replace the draft inside $\reg$, giving the \emph{merged} scene
\begin{equation}
\merged(\reg)=(\draft\setminus\draft|_{\reg})\cup \ver(\reg).
\label{eq:merged}
\end{equation}

\noindent\textbf{Goal.}
We seek a region $\reg$ with $\policy(\merged(\reg))=\policy(\full)$ that uses fewer detector-input pixels than the reference, counting the draft and every crop; neither $\merged(\reg)=\full$ nor a minimum-pixel $\reg$ is required.
The difficulty is that $\reg$ must be chosen without knowing $\full$: it must reach the action-changing objects that the draft misses and the detections that the current action rests on, some possibly spurious.

\subsection{Decision-sensitivity field and support}
\label{sec:field}

\noindent\textbf{Sensitivity field.}
Rather than predicting where an unseen object may be, \method asks where one would change the action.
A \emph{probe} $o_{\state}$ is a hypothetical object with state $\state$, and the \emph{decision-sensitivity field} of a scene $\scene$ holds the states whose insertion changes the action:
\begin{equation}
\field(\scene) \;=\; \{\,\state \,:\, \policy(\scene\cup\{o_{\state}\})\neq\policy(\scene)\,\}.
\label{eq:field}
\end{equation}
The field measures action sensitivity, not object likelihood, and needs no image evidence, so it exists even where the draft detects nothing.
In practice, it is evaluated on a finite ground-plane grid of ranges and lateral offsets with a fixed-size probe that closes at a fixed speed, which can miss sensitive states.
Some planners need no explicit calls: for a braking controller, a probe is sensitive when its demand reaches the threshold of the next stronger action.
The calibration projects each sensitive probe to an image box, and the \emph{field footprint} $\foot(\scene)$ is the single crop bounding them.

\noindent\textbf{Support.}
The field cannot correct a spurious or mislocalized detection that sustains the current action: \eg, one that triggers \textsc{hard-brake} leaves the field empty, since no stronger action exists.
\method therefore also verifies a \emph{support} $\Sup(\scene)\subseteq \scene$, detections sufficient to preserve the current action (not necessarily minimal): removing the others leaves the action unchanged, whereas verifying the support may change it.
Its crops $\supcrops(\scene)$ are the object boxes, enlarged for context.
For the braking controller above, the support holds every object whose own demand reaches the threshold of the current action, and it is empty at \textsc{keep}.
It can thus hold objects that are not \emph{pivots}~\cite{cfimportance} (objects whose removal changes the action): if two objects each justify \textsc{hard-brake}, removing either leaves it unchanged, although removing both may change it.

\subsection{Iterative verification and decision exactness}
\label{sec:loop}

Verifying a region can change the action, and with it the field and the support, so regions selected from the draft alone may not suffice: \eg, removing a spurious \textsc{hard-brake} detection lowers the action, and a previously empty field requests a new region.
\Cref{alg:loop} therefore iterates from the draft (\Cref{fig:method}): it requests the field footprint and the support crops of the current merged scene and runs the reference detector only on those not yet inside $\reg$, reusing earlier detections.
It stops when every requested crop lies inside $\reg$.
Every non-final round adds pixels to $\reg$, so the loop terminates; the experiments cap it at three rounds, which cuts off no braking test frame (\supp{sec:supp_results}).

\begin{algorithm}[t]
\caption{Verification loop of \method}
\label{alg:loop}
\small
\begin{algorithmic}[1]
\Require frame $\img$, planner $\policy$, detectors $\detL$, $\detH$
\State $\merged \gets \draft \gets \detL(\img)$;\; $\reg \gets \varnothing$
\While{\textbf{true}}
  \State $\mathcal{N} \gets$ crops of $\foot(\merged)\cup\supcrops(\merged)$ not inside $\reg$
  \State \textbf{if} $\mathcal{N}=\varnothing$ \textbf{then return} $\policy(\merged)$
  \State $\reg \gets \reg\cup \mathcal{N}$ \Comment{run $\detH$ on $\mathcal{N}$ only}
  \State $\merged \gets (\draft\setminus\draft|_{\reg})\cup \ver(\reg)$ \Comment{reuse earlier crops}
\EndWhile
\end{algorithmic}
\end{algorithm}

\noindent\textbf{Why a stable action is the reference action.}
When the loop stops, it has verified the detections that sustain the action and every location where one unseen object would change it.
What remains are draft detections outside the support and reference objects outside the field; the action is exact if these cannot matter, which two properties of the planner ensure.
First, the support must be \emph{certifying}: the action survives removing any subset ${\mathcal{D}\subseteq\scene\setminus\Sup(\scene)}$ of the other detections, with or without one inserted object outside the field, ${\mathcal{O}\in\mathcal{Y}(\scene)}$,
\begin{equation}
\policy\big((\scene\setminus \mathcal{D})\cup \mathcal{O}\big)=\policy(\scene),
\label{eq:support}
\end{equation}
where $\mathcal{Y}(\scene)$ contains $\varnothing$ and each singleton of an object with state outside $\field(\scene)$.
The insertion matters because a removal can make an unseen object relevant: once a spurious obstacle near an alternative trajectory is removed, an unseen obstacle on the current trajectory can make the alternative the better choice.
Second, the field probes unseen objects one at a time, which suffices only if individually irrelevant objects cannot change the action jointly, as additive demands can.
A planner is \emph{insertion-stable} if
$\policy(\scene\cup \mathcal{X})=\policy(\scene)$
for every finite $\mathcal{X}$ with states outside $\field(\scene)$.

\begin{theorem}[Decision exactness]
Let \Cref{alg:loop} stop with $\reg$ and $\merged$.
It returns $\policy(\full)$ if (i)~$\policy$ is insertion-stable; (ii)~$\Sup$ is certifying; (iii)~\emph{locality}: $\ver(\reg)=\full|_{\reg}$, with identical boxes and states; and (iv)~\emph{coverage}: every object of $\full$ with state in $\field(\merged)$ has its box center in $\reg$.\label{thm:exact}
\end{theorem}

\noindent\emph{Proof sketch.}
By locality, $\merged=(\draft\setminus\draft|_{\reg})\cup\full|_{\reg}$, and the support lies in $\full|_{\reg}$ since $\reg$ contains its crops.
The leftover draft objects $\mathcal{D}=\draft\setminus\draft|_{\reg}$ lie outside the support, and, by coverage, the states of the leftover reference objects $\bar\full=\full\setminus\full|_{\reg}$ lie outside $\field(\merged)$.
Certification removes $\mathcal{D}$, alone or with any single $o\in\bar\full$ inserted: $\policy(\full|_{\reg})=\policy(\merged)=\policy(\full|_{\reg}\cup\{o\})$.
Each object of $\bar\full$ is thus individually irrelevant at $\full|_{\reg}$, and insertion stability adds them all: $\policy(\full)=\policy(\full|_{\reg})$ (\supp{sec:supp_theory}).
\hfill$\square$

\noindent\textbf{Scope of the guarantee.}
Conditions (i)--(ii) concern the planner and hold exactly for the two classes below; (iii)\nobreakdash--(iv) concern verification, hold only approximately on real images, and are not checked at run time: the footprint aims at coverage without proving it.
The theorem thus covers an idealized verification and traces any residual disagreement to locality or coverage, which \Cref{sec:practice,sec:setup} narrow; idealizing both gives 99.7\nobreakdash--99.9\% agreement with $\policy(\full)$ (\Cref{sec:exp_exact}).
Even under locality, each mechanism is needed: dropping the field, the support, its certification (pivots only), or the iteration admits counterexamples (\supp{sec:supp_theory}); \Cref{tab:ablation} measures the effect.

In the first class, as in emergency braking, each object imposes a demand and the largest sets the action.

\begin{definition}[Max-composable planner]
Let actions $a\in\{0,\dots,K\}$ be intensity-ordered, with thresholds $0=\tau_0<\tau_1<\dots<\tau_K<\tau_{K+1}=\infty$.
$\policy$ is \emph{max-composable} if
\begin{equation}
\policy(\scene)=\mathsf{q}\big(\max_{o\in \scene}\dem(\state(o))\big),\ \ \mathsf{q}(u)=\max\{a:u\ge\tau_a\},
\label{eq:maxcomp}
\end{equation}
for a per-object demand $\dem\ge 0$, where $\max\varnothing=0$.\label{def:maxcomp}
\end{definition}

\begin{corollary}[Max-composable planners]
Such a planner is insertion-stable with $\field(\scene)=\{\state:\dem(\state)\ge\tau_{a+1}\}$, where $a=\policy(\scene)$, and the support $\Sup(\scene)=\{o\in \scene:\dem(\state(o))\ge\tau_a\}$ for $a\ge1$, with $\Sup(\scene)=\varnothing$ for $a=0$, is certifying.\label{cor:maxcomp}
\end{corollary}

Insertions only raise the largest demand; removals keep the action while it stays at least $\tau_a$.
For braking, the field is a stopping corridor: the part of the ego corridor where a probe's demand reaches the next threshold.
It follows from the planner alone and thus changes with the action, ego speed, and path: it shortens at \textsc{decelerate} and is empty at \textsc{hard-brake}, where only the support is verified; other planners give other fields (\Cref{sec:exp_ablation}).

The second class selects among fixed candidates $c$ (trajectories), each scored by its preference capped by its worst object, \eg, its smallest clearance.

\begin{corollary}[Bottleneck selection]
Let $\policy(\scene)$ select, from a fixed ordered set of candidates $c$, the first maximizer of
\begin{equation*}
\mathsf{s}(c,\scene)=\min\Big(\beta(c),\,\min_{o\in \scene}\mathsf{h}\big(c,\state(o)\big)\Big),\quad \min\varnothing=+\infty,
\end{equation*}
with a preference $\beta$ and a per-object score $\mathsf{h}$.
Then $\policy$ is insertion-stable, and $\Sup(\scene)=\{o\in \scene:\mathsf{h}(c,\state(o))\le \mathsf{s}(\policy(\scene),\scene)\text{ for some }c\neq\policy(\scene)\}$ is certifying.\label{cor:bottleneck}
\end{corollary}

The support holds the objects that keep some competing candidate at or below the selected score, such as an obstacle that blocks only an alternative trajectory.
Scores compose by the minimum, which gives insertion stability, and field and support need per-candidate scores but no re-planning.

\subsection{Practical considerations}
\label{sec:practice}

\noindent\textbf{Approximating the reference.}
Three mechanisms improve locality (\supp{sec:supp_impl}).
\emph{Grid-aligned crops} are cut from the reference's resized input on the detector's stride grid, so the detector sees the same pixels in a crop as in the whole frame.
The \emph{truncation guard} re-detects, in an enlarged crop, a box cut by an interior crop boundary, unless its demand is zero however the box is completed.
\emph{Same-scale history} estimates closing rates with a learned \emph{looming estimate} from observations at one scale; the first round re-requests the previous support crops, and an object newly verified near the path is also re-detected in the previous frame.

\noindent\textbf{Lift.}
The lift estimates states by monocular geometry and the looming estimate and, once the loop has stopped, corrects the range of the verified support objects with a learned \emph{range refiner}.
The reference uses the same lift; \Cref{thm:exact} covers the loop, not this final correction.

\noindent\textbf{General planners.}
Without closed forms, probes and removals approximate the field and a support of pivots, which need not be certifying: \method still applies, at the cost of planner calls, but outside \Cref{thm:exact} (\Cref{sec:exp_ablation}).

\section{Experiments}
\label{sec:exp}

\subsection{Experimental setup}
\label{sec:setup}

\begin{table}[t]
\centering
\scriptsize
\setlength{\tabcolsep}{3.49pt}
\begin{tabular}{@{}lgccgccgcc@{}}
\toprule
 & \multicolumn{3}{c}{nuScenes} & \multicolumn{3}{c}{Argoverse 2} & \multicolumn{3}{c}{Argoverse-HD} \\
\cmidrule(lr){2-4}\cmidrule(lr){5-7}\cmidrule(l){8-10}
\raisebox{6.2pt}[0pt][0pt]{Method} & Agr.\up & Rep.\up & px\dn & Agr.\up & Rep.\up & px\dn & Agr.\up & Rep.\up & px\dn \\
\midrule
\rf{u320 (draft)} & \rf{92.7\sgu{}} & \rf{0\sgu{}} & \rf{4.3} & \rf{85.3\sgu{}} & \rf{0\sgu{}} & \rf{2.6} & \rf{78.8\sgu{}} & \rf{0\sgu{}} & \rf{3.1} \\
\rf{u320$^{+}$} & \rf{96.5\sgu{}} & \rf{59.9\sgu{}} & \rf{4.3} & \rf{94.7\sgu{}} & \rf{70.2\sgu{}} & \rf{2.6} & \rf{91.4\sgu{}} & \rf{63.8\sgu{}} & \rf{3.1} \\
\rf{u512} & \rf{97.7\sgu{}} & \rf{79.6\sgu{}} & \rf{10.2} & \rf{96.4\sgu{}} & \rf{84.0\sgu{}} & \rf{6.7} & \rf{94.5\sgu{}} & \rf{82.6\sgu{}} & \rf{7.1} \\
\rf{u640} & \rf{98.2\sgu{}} & \rf{84.7} & \rf{17.1} & \rf{97.3\sgu{}} & \rf{88.3\sgu{}} & \rf{10.3} & \rf{95.8\sgu{}} & \rf{87.4\sgu{}} & \rf{11.6} \\
\rf{u768} & \rf{98.6} & \rf{88.6} & \rf{23.9} & \rf{97.8\sgu{}} & \rf{90.4} & \rf{14.7} & \rf{96.1\sgu{}} & \rf{88.5\sgu{}} & \rf{16.0} \\
\rf{u1280} & \rf{98.3\sgu{}} & \rf{87.8} & \rf{65.4} & \rf{97.6\sgu{}} & \rf{91.8} & \rf{40.0} & \rf{96.2\sgu{}} & \rf{90.0} & \rf{44.4} \\
\midrule
CF~\cite{cfimportance} & 97.5\sgu{} & 71.3\sgu{} & 5.3 & 96.1\sgu{} & 78.7\sgu{} & 3.8 & 91.5\sgu{} & 64.4\sgu{} & 4.5 \\
AnA~\cite{ana} & 98.1\sgu{} & 78.5\sgu{} & 8.3 & 97.4\sgu{} & 85.6\sgu{} & 5.8 & 94.7\sgu{} & 77.5\sgu{} & 11.7 \\
SAHI~\cite{sahi} & 97.4\sgu{} & \thd{85.9} & 188 & 93.6\sgu{} & 83.5\sgu{} & 165 & 91.4\sgu{} & 82.7\sgu{} & 154 \\
FOVEA~\cite{fovea} & 97.9\sgu{} & 85.0 & 69.7 & 96.5\sgu{} & 88.0\sgu{} & 42.6 & 95.6\sgu{} & 88.1\sgu{} & 47.6 \\
UFP~\cite{ufpmp} & \thd{98.0\sgu{}} & 83.2 & 18.1 & \thd{97.2\sgu{}} & 87.6\sgu{} & 9.0 & 95.3\sgu{} & 84.5\sgu{} & 12.8 \\
AdaScale~\cite{adascale} & \scd{98.4} & \fst{86.9} & 30.6 & \fst{99.1\sgd{}} & \fst{96.8\sgd{}} & 26.8 & \fst{98.2} & \fst{95.2} & 29.9 \\
\midrule
\method-fast & \fst{98.7} & \scd{86.8} & 8.1 & \scd{98.2} & \thd{91.5} & 5.6 & \thd{97.4\sgu{}} & \thd{91.6} & 6.5 \\
\oursrow\textbf{\method} & \fst{98.7} & \fst{86.9} & 8.4 & \scd{98.2} & \scd{91.7} & 5.9 & \scd{97.5} & \scd{92.0} & \ourslast 6.9 \\
\bottomrule
\end{tabular}
\caption{\textbf{Decision fidelity.} \method agrees with $\policy(\full)$ more often than every uniform resolution and every adaptive method but AdaScale, which uses 3.6--4.5$\times$ as many pixels. Agr., Rep.: agreement and repair rate (\%); px: pixels (\% of native; $\full$: 38.4, 23.2, and 25.3); CF: counterfactual importance. Top block: whole-frame inference. \rankkey{} among the adaptive methods; \method-fast: one round; \sgu{}\,/\,\sgd{}: \method significantly higher\,/\,lower.}
\label{tab:fid}
\end{table}

\begin{table*}[t]
\renewcommand{\rf}[1]{\textcolor{black!60}{#1}}
\centering
\scriptsize
\setlength{\tabcolsep}{3.90pt}
\begin{tabular}{@{}lccccgccccgccccg>{\hspace{4pt}}NMMMG@{}}
\toprule
 & \multicolumn{5}{c}{nuScenes} & \multicolumn{5}{c}{Argoverse 2} & \multicolumn{5}{c}{Argoverse-HD} & \multicolumn{5}{c}{\textbf{Mean}} \\
\cmidrule(lr){2-6}\cmidrule(lr){7-11}\cmidrule(lr){12-16}\cmidrule(l){17-21}
\raisebox{6.2pt}[0pt][0pt]{Method} & px\dn & GF\dn & P\up & R\up & F1\up & px\dn & GF\dn & P\up & R\up & F1\up & px\dn & GF\dn & P\up & R\up & F1\up & px\dn & GF\dn & P\up & R\up & F1\up \\
\midrule
\rf{u320 (draft)} & \rf{4.3} & \rf{4.3} & \rf{31.4\sgu{}} & \rf{65.3} & \rf{42.4\sgu{}} & \rf{2.6} & \rf{5.8} & \rf{29.3\sgu{}} & \rf{82.1} & \rf{43.2\sgu{}} & \rf{3.1} & \rf{5.0} & \rf{10.9\sgu{}} & \rf{70.4\sgd{}} & \rf{18.9\sgu{}} & \rf{3.3} & \rf{5.0} & \rf{23.9\sgu{}} & \rf{72.6\sgd{}} & \rf{34.8\sgu{}} \\
\rf{u320$^{+}$} & \rf{4.3} & \rf{9.6} & \rf{52.8\sgu{}} & \rf{57.1\sgu{}} & \rf{54.9\sgu{}} & \rf{2.6} & \rf{19.0} & \rf{62.9\sgu{}} & \rf{76.8} & \rf{69.2\sgu{}} & \rf{3.1} & \rf{20.6} & \rf{25.5\sgu{}} & \rf{50.7} & \rf{34.0\sgu{}} & \rf{3.3} & \rf{16.4} & \rf{47.1\sgu{}} & \rf{61.6} & \rf{52.7\sgu{}} \\
\rf{u512} & \rf{10.2} & \rf{17.8} & \rf{58.8\sgu{}} & \rf{66.0} & \rf{62.2\sgu{}} & \rf{6.7} & \rf{33.1} & \rf{64.0\sgu{}} & \rf{80.5} & \rf{71.3\sgu{}} & \rf{7.1} & \rf{32.6} & \rf{34.0\sgu{}} & \rf{47.9} & \rf{39.8\sgu{}} & \rf{8.0} & \rf{27.8} & \rf{52.3\sgu{}} & \rf{64.8} & \rf{57.8\sgu{}} \\
\rf{u640} & \rf{17.1} & \rf{26.1} & \rf{60.2\sgu{}} & \rf{66.0} & \rf{63.0\sgu{}} & \rf{10.3} & \rf{42.9} & \rf{71.8\sgu{}} & \rf{81.6\sgd{}} & \rf{76.4} & \rf{11.6} & \rf{43.6} & \rf{37.9} & \rf{54.9} & \rf{44.8} & \rf{13.0} & \rf{37.5} & \rf{56.6\sgu{}} & \rf{67.5} & \rf{61.4\sgu{}} \\
\rf{u768} & \rf{23.9} & \rf{33.6} & \rf{62.4\sgu{}} & \rf{66.7} & \rf{64.5\sgu{}} & \rf{14.7} & \rf{53.9} & \rf{71.7\sgu{}} & \rf{82.6\sgd{}} & \rf{76.8} & \rf{16.0} & \rf{52.5} & \rf{33.3\sgu{}} & \rf{47.9} & \rf{39.3\sgu{}} & \rf{18.2} & \rf{46.7} & \rf{55.8\sgu{}} & \rf{65.7} & \rf{60.2\sgu{}} \\
{u960 ($\full$)} & {38.4} & {48.9} & {61.1\sgu{}} & {69.4} & {65.0\sgu{}} & {23.2} & {74.9} & {70.6\sgu{}} & {82.1\sgd{}} & {75.9} & {25.3} & {70.8} & {33.6\sgu{}} & {52.1} & {40.9} & {29.0} & {64.8} & {55.1\sgu{}} & {67.9\sgd{}} & {60.6\sgu{}} \\
\rf{u1280} & \rf{65.4} & \rf{76.8} & \rf{62.5\sgu{}} & \rf{68.0} & \rf{65.1} & \rf{40.0} & \rf{115} & \rf{69.0\sgu{}} & \rf{82.1\sgd{}} & \rf{75.0} & \rf{44.4} & \rf{106} & \rf{33.0\sgu{}} & \rf{50.7} & \rf{40.0\sgu{}} & \rf{50.0} & \rf{99.1} & \rf{54.9\sgu{}} & \rf{66.9} & \rf{60.0\sgu{}} \\
\rf{native} & \rf{103} & \rf{115} & \rf{66.9} & \rf{64.6} & \rf{65.7} & \rf{101} & \rf{254} & \rf{68.9\sgu{}} & \rf{82.6\sgd{}} & \rf{75.1} & \rf{101} & \rf{201} & \rf{36.4\sgu{}} & \rf{54.9} & \rf{43.8} & \rf{102} & \rf{190} & \rf{57.4\sgu{}} & \rf{67.4} & \rf{61.6} \\
\midrule
Chanakya~\cite{chanakya} & 7.9 & 9.2 & 41.8\sgu{} & 60.5\sgu{} & 49.4\sgu{} & 4.9 & 14.8 & 36.2\sgu{} & 80.5 & 49.9\sgu{} & 6.3 & 17.2 & 16.2\sgu{} & \thd{54.9} & 25.1\sgu{} & 6.4 & 13.7 & 31.4\sgu{} & 65.3 & 41.5\sgu{} \\
BAR~\cite{bgtada} & 36.6 & 43.6 & 43.5\sgu{} & 63.9 & 51.8\sgu{} & 21.4 & 61.8 & 37.6\sgu{} & \scd{82.6}\sgd{} & 51.7\sgu{} & 23.1 & 52.3 & 15.2\sgu{} & \scd{56.3} & 24.0\sgu{} & 27.0 & 52.6 & 32.1\sgu{} & \scd{67.6} & 42.5\sgu{} \\
SAHI~\cite{sahi} & 188 & 202 & 46.7\sgu{} & \fst{68.0} & 55.4\sgu{} & 165 & 398 & 52.8\sgu{} & \fst{84.2\sgd{}} & 64.9\sgu{} & 154 & 288 & 23.2\sgu{} & \fst{63.4\sgd{}} & 34.0\sgu{} & 169 & 296 & 40.9\sgu{} & \fst{71.9\sgd{}} & 51.4\sgu{} \\
CF~\cite{cfimportance} & 5.3 & 6.2 & \thd{62.8\sgu{}} & 55.1\sgu{} & 58.7\sgu{} & 3.8 & 11.9 & 70.1\sgu{} & 75.3 & 72.6\sgu{} & 4.5 & 10.9 & 18.5\sgu{} & 43.7 & 25.9\sgu{} & 4.5 & 9.7 & 50.4\sgu{} & 58.0\sgu{} & 52.4\sgu{} \\
FOVEA~\cite{fovea} & 69.7 & 80.1 & 60.5\sgu{} & \thd{66.7} & \scd{63.4}\sgu{} & 42.6 & 117 & 63.8\sgu{} & \thd{81.6} & 71.6\sgu{} & 47.6 & 108 & 31.2\sgu{} & 49.3 & 38.3\sgu{} & 53.3 & 101 & 51.8\sgu{} & 65.8 & 57.8\sgu{} \\
AdaScale~\cite{adascale} & 30.6 & 40.0 & 59.1\sgu{} & 66.0 & 62.4\sgu{} & 26.8 & 82.4 & 69.5\sgu{} & 80.5 & 74.6 & 29.9 & 77.5 & 32.7\sgu{} & 52.1 & \thd{40.2} & 29.1 & 66.7 & 53.8\sgu{} & \thd{66.2} & 59.1\sgu{} \\
AnA~\cite{ana} & 8.3 & 10.9 & \scd{67.7} & 58.5\sgu{} & \thd{62.8\sgu{}} & 5.8 & 23.6 & \scd{75.8} & 77.4 & \scd{76.6} & 11.7 & 34.0 & \thd{34.0} & 45.1 & 38.8\sgu{} & 8.6 & 22.8 & \scd{59.2}\sgu{} & 60.3\sgu{} & \thd{59.4\sgu{}} \\
UFP~\cite{ufpmp} & 18.1 & 26.7 & 62.6\sgu{} & 62.6 & 62.6\sgu{} & 9.0 & 39.4 & \thd{72.1\sgu{}} & \thd{81.6} & \thd{76.5} & 12.8 & 47.5 & \scd{37.6} & 53.5 & \scd{44.2} & 13.3 & 37.8 & \thd{57.4\sgu{}} & 65.9 & \scd{61.1}\sgu{} \\
\midrule
\oursrow\textbf{\method} & 8.4 & 11.4 & \fst{70.7} & \scd{67.3} & \fst{69.0} & 5.9 & 21.3 & \fst{77.1} & 77.9 & \fst{77.5} & 6.9 & 25.1 & \fst{44.3} & 49.3 & \fst{46.7} & \oursfull 7.1 & 19.3 & \fst{64.0} & 64.8 & \ourslast \fst{64.4} \\
\bottomrule
\end{tabular}
\caption{\textbf{Decision quality and cost.} \method has the highest precision and F1 on every dataset with 6--8\% of the native pixels; the next-best mean F1 is that of native inference. px: pixels (\% of native); GF: GFLOPs per frame, detector and lift; P, R, F1: precision, recall, and F1 of the braking episodes (\%); Mean: over the datasets. Top block: whole-frame inference (u960 is $\full$); BAR: Budget-Adaptive Routing; CF: counterfactual importance. \rankkey{} among the adaptive methods, ordered by mean F1; \sgu{}\,/\,\sgd{}: \method significantly higher\,/\,lower.}
\label{tab:main}
\end{table*}

\noindent\textbf{Planner and metrics.}
The planner is the braking controller of~\cite{deep}, which is max-composable: the largest deceleration needed to stop short of an object in the ego corridor ($\pm$1.2\,m, up to 80\,m) selects \textsc{keep}, \mbox{\textsc{decelerate}} ($\ge{}$1.0\,m/s$^2$), or \textsc{hard-brake} ($\ge{}$3.5\,m/s$^2$).
\emph{Decision fidelity}, the primary target, is the agreement of the action with $\policy(\full)$, and the \emph{repair rate} is this agreement on the frames where the draft differs from $\policy(\full)$; both score a whole pipeline, lift included.
Since $\policy(\full)$ can itself be wrong, \emph{decision quality} scores the actions against the annotations by braking episode: a maximal run of frames that require braking is \emph{missed} if a pipeline never brakes in it, and a braking run with no required braking is \emph{false}.
We report episode precision, recall, and F1.
Cost is measured in \emph{detector-input pixels} (``pixels''): all pixels fed to a detector in a frame, draft and truncation guard included, in \% of the native frame; the detector's FLOPs are proportional to them (total FLOPs: \Cref{tab:main} and \supp{sec:supp_results}).

\noindent\textbf{Datasets and protocol.}
We use the front cameras of nuScenes~\cite{nuscenes} (228 held-out scenes of its trainval set, not the official test split), Argoverse~2~\cite{av2} (val, 150 logs; the ego hood is masked for every pipeline), and Argoverse-HD~\cite{streaming} (val, 24 sequences).
One configuration is selected on development data of nuScenes and Argoverse~2, frozen before test evaluation, and applied to Argoverse-HD without tuning (\supp{sec:supp_data}).
We test differences with 95\% paired bootstrap intervals (2,000 resamples over scenes, logs, or sequences), not adjusted for multiple comparisons unless stated; non-significance is not equivalence.

\noindent\textbf{Baselines.}
We compare whole-frame inference at a long side of $N$ pixels (u$N$) and \mbox{frame-,} region-, and object-level methods, built on official code where possible, all with the same draft, lift, and unmodified detector, our crops and guard if they crop, and a setting tuned on development data over a grid that includes the published one (\supp{sec:supp_impl}).

\noindent\textbf{Implementation.}
The cheap and reference detectors are the same COCO-pretrained~\cite{coco} YOLOv8s~\cite{yolov8}, applied to the frame at a long side of 320 pixels and to crops at the scale of the frame resized to a long side of 960 pixels, the verification resolution.
All pipelines but the draft share the lift of \Cref{sec:practice} (the monocular geometry of~\cite{deep}, the learned looming estimate, and the range refiner), which \method applies to the objects it verifies.
The reference $\full$ is whole-frame inference at the verification resolution (u960); native resolution is a separate baseline. u320 is the draft, and u320$^{+}$ adds the learned parts at the same pixels.
Only \method uses the field, with a vehicle-sized probe on a ground grid (both varied in \supp{sec:supp_grid}), a margin that counts a cell as sensitive from 60\% of the next threshold (a larger region, which \Cref{thm:exact} permits), and a cap of three rounds.

\subsection{Decision fidelity}
\label{sec:exp_exact}

\begin{figure}[t]
\centering
\includegraphics[width=\linewidth]{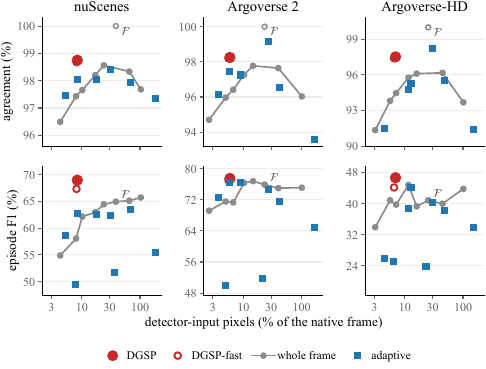}
\caption{\textbf{Fidelity and decision quality versus pixel cost.} No baseline has both fewer pixels and a higher agreement (top, \Cref{tab:fid}), and none has a significantly higher F1 (bottom, \Cref{tab:main}). $\full$: the reference (open marker: 100\% by definition).}
\label{fig:pareto}
\end{figure}

\begin{table}[t]
\centering
\scriptsize
\setlength{\tabcolsep}{3.76pt}
\begin{tabular}{@{}llccgccccc@{}}
\toprule
 & & & \multicolumn{2}{c}{Decisive object} & \multicolumn{2}{c}{Ego speed} & \multicolumn{3}{c}{Range of decisive object} \\
\cmidrule(lr){4-5}\cmidrule(lr){6-7}\cmidrule(l){8-10}
 & \raisebox{6.1pt}[0pt][0pt]{Method} & \raisebox{6.1pt}[0pt][0pt]{All} & Seen & Missed & Low & High & $<$20 & 20--40 & 40--60 \\
\midrule
 & u512 & 97.7\sgu{} & 91.5\sgu{} & 74.5\sgu{} & \underline{99.3}\sgd{} & 95.8\sgu{} & 92.5 & 88.1\sgu{} & 82.7\sgu{} \\
& u640 & \underline{98.2}\sgu{} & \underline{93.6} & \underline{81.6} & \textbf{99.5}\sgd{} & \underline{97.0}\sgu{} & \underline{93.2} & \underline{92.2} & \underline{89.0}\sgu{} \\
\oursrow\cellcolor{white}\multirow{-3}{*}{\rotatebox{90}{nuSc.}} & \textbf{\method} & \textbf{98.7} & \textbf{95.9} & \textbf{87.9} & 98.8 & \textbf{98.7} & \textbf{94.8} & \textbf{95.1} & \ourslast \textbf{95.3} \\
\midrule
 & u512 & 96.4\sgu{} & 92.3\sgu{} & 81.1\sgu{} & \underline{98.3} & 95.9\sgu{} & 92.8\sgu{} & 92.5\sgu{} & 88.8\sgu{} \\
& u640 & \underline{97.3}\sgu{} & \underline{94.3}\sgu{} & \underline{85.4}\sgu{} & \textbf{98.7} & \underline{96.8}\sgu{} & \underline{95.0} & \underline{93.4}\sgu{} & \underline{92.6}\sgu{} \\
\oursrow\cellcolor{white}\multirow{-3}{*}{\rotatebox{90}{AV2}} & \textbf{\method} & \textbf{98.2} & \textbf{96.6} & \textbf{94.3} & 98.2 & \textbf{98.8} & \textbf{96.5} & \textbf{96.8} & \ourslast \textbf{96.7} \\
\midrule
 & u512 & 94.5\sgu{} & 91.9\sgu{} & 77.2\sgu{} & 96.4\sgu{} & 90.2\sgu{} & \underline{95.5}\sgu{} & 85.5\sgu{} & 87.4\sgu{} \\
& u640 & \underline{95.8}\sgu{} & \underline{92.9}\sgu{} & \underline{86.1}\sgu{} & \underline{96.9} & \underline{93.2}\sgu{} & 95.3\sgu{} & \underline{89.6}\sgu{} & \underline{92.6}\sgu{} \\
\oursrow\cellcolor{white}\multirow{-3}{*}{\rotatebox{90}{A-HD}} & \textbf{\method} & \textbf{97.5} & \textbf{96.9} & \textbf{94.4} & \textbf{97.2} & \textbf{96.5} & \textbf{99.2} & \textbf{95.2} & \ourslast \textbf{96.7} \\
\bottomrule
\end{tabular}
\caption{\textbf{Decision fidelity by regime.} \method leads most where the draft misses the decisive object. Agreement (\%). Seen\,/\,Missed: frames requiring braking whose decisive object (largest demand) is in\,/\,not in the draft; Range: its range (m); Low\,/\,High: ego-speed terciles. \rankkey; \sgu{}\,/\,\sgd{}: \method significantly higher\,/\,lower.}
\label{tab:regime}
\end{table}

\noindent\textbf{\method reproduces the reference action at a fraction of its pixels.}
It agrees with $\policy(\full)$ on 97.5--98.7\% of frames---shifted by one pixel, the reference itself reaches only 98.7\nobreakdash--99.2\%---with 22--27\% of the pixels and 23--36\% of the FLOPs of the reference, lift included, and repairs 87\nobreakdash--92\% of the draft's disagreements (\Cref{tab:fid,tab:main}). The shared lift alone repairs 60--70\% (u320$^{+}$), and relative to u320$^{+}$, \method has 64--71\% fewer disagreements and u512 has 32--36\% fewer. Each remaining disagreement traces to coverage (35--42\%), locality (34--47\%), or the refiner outside \Cref{thm:exact} (11\nobreakdash--26\%); idealizing locality and coverage raises the agreement to 99.7\nobreakdash--99.9\% (\supp{sec:supp_results}).

\noindent\textbf{It is more faithful than uniform inference with more pixels.}
u512, the next uniform resolution above the pixels of \method (and 1.3--1.6$\times$ its FLOPs), repairs 7.3--9.4 points fewer and agrees on 1.1--3.0 points fewer frames, all significant after Holm correction (\supp{sec:supp_impl}).
Most frames require no braking; on the 10\nobreakdash--24\% that do, the gap widens to 5--9 points (95--96\% against 88--91\%).
On the hardest frames, where $\policy(\full)$ changes, \method still agrees on 84\%, 83\%, and 68\%, against 77\%, 75\%, and 54\% for u512.
The result holds with a second detector (YOLO11s), one \mbox{fine-tuned} on driving data, a native-resolution reference, or a geometric lift (\supp{sec:supp_results}).

\noindent\textbf{The advantage is largest where the draft is blind.}
On frames where the draft misses the decisive object (\Cref{tab:regime}), the advantage over u512 is 13--17 points, about three times that where the draft contains it, and even u640, with 1.7\nobreakdash--2$\times$ \method's pixels, trails by 6--9 points: no detection marks the object, so verification reaches it through the field.
The advantage also grows with the range the planner must cover: the gap is within a point at low ego speed and grows with the range of the decisive object, from 2--4 points below 20\,m to 8--13 points at 40--60\,m. With the planner's ego speed doubled and its range limit lifted, the overall gap widens from 1--3 to 5--10 points; for the smallest decisive objects, the repair gap is 16--28 points (\supp{sec:supp_results}).

\subsection{Decision quality versus pixel cost}
\label{sec:exp_main}
\label{sec:exp_adaptive}

\noindent\textbf{\method matches the reference's F1 at higher precision with about a quarter of its pixels.}
Against the annotations, it has the highest F1 and precision on every dataset (\Cref{tab:main} and \Cref{fig:pareto}).
Its F1 exceeds the reference's by 4.0, 1.6, and 5.8 points (significantly on nuScenes) and u512's by 6.2\nobreakdash--6.9 points (all significant before correction; \supp{sec:supp_impl}).
Its mean F1 is significantly above that of every method except native-resolution inference.

\noindent\textbf{The gain is in precision, with a recall trade-off against the reference.}
Where \method departs from $\policy(\full)$, it mostly drops braking that the scene does not require: it brakes less in 77--81\% of these frames, and the annotations side with it in 61--72\% and with $\policy(\full)$ in only 21--33\%.
It is significantly more precise than u512 and the reference on every dataset. Restricted to the crops of \method, the reference becomes nearly as precise on nuScenes and Argoverse~2: there, the planner's choice of regions explains most of the gain.
In exchange, the reference has 2.0--4.2 points higher recall (significantly on Argoverse~2); u512 does not differ significantly. With thresholds selected per pipeline on development data, \method misses 38, 25, and 17 episodes and the reference 45, 24, and 20, with no significant difference in false braking.
On 65 held-out Argoverse-HD sequences, \method's F1 is significantly above that of u512 and the reference (\supp{sec:supp_results}).

\begin{figure*}[t]
\centering
\newcommand{\qimg}[3]{\begin{tikzpicture}\node[inner sep=0pt] (q) {\includegraphics[width=0.245\textwidth]{#1}};%
\if\relax\detokenize{#2}\relax\else\node[anchor=north west, fill=white, fill opacity=0.85, text opacity=1, inner sep=1.4pt, font=\scriptsize, text=#3] at (q.north west) {#2};\fi\end{tikzpicture}}
\newcommand{\qbev}[1]{\includegraphics[height=0.1225\textwidth]{#1}}
\newcommand{\qbevm}[3]{\begin{tikzpicture}[baseline=(b.south)]\node[inner sep=0pt, anchor=south west] (b) at (0,0) {\qbev{#1}};%
\begin{scope}[x={(b.south east)}, y={(b.north west)}]\draw[white, line width=1.4pt] (#2-0.05,#3-0.035) rectangle (#2+0.05,#3+0.035);%
\draw[qR, line width=0.7pt, dash pattern=on 1.1pt off 0.8pt] (#2-0.05,#3-0.035) rectangle (#2+0.05,#3+0.035);\end{scope}\end{tikzpicture}}
\definecolor{qB}{HTML}{1F77B4}\definecolor{qO}{HTML}{FF7F0E}\definecolor{qG}{HTML}{7F7F7F}\definecolor{qR}{HTML}{D62728}\definecolor{qN}{HTML}{2CA02C}
\newcommand{\qlg}[3]{\draw[#2, line width=1.1pt] (0,#1-0.09) rectangle ++(0.30,0.18); \node[anchor=west, align=left, inner sep=0.5pt, font=\scriptsize] at (0.38,#1) {#3};}
\setlength{\tabcolsep}{1.2pt}
\renewcommand{\arraystretch}{0.6}
\begin{tabular}{@{}ccccc@{}}
& {\footnotesize draft} & {\footnotesize field, support} & {\footnotesize verified crops} & {\footnotesize result} \\
\rotatebox{90}{\parbox{0.1225\textwidth}{\centering\footnotesize (a) unseen-object\\recovery}} &
\qimg{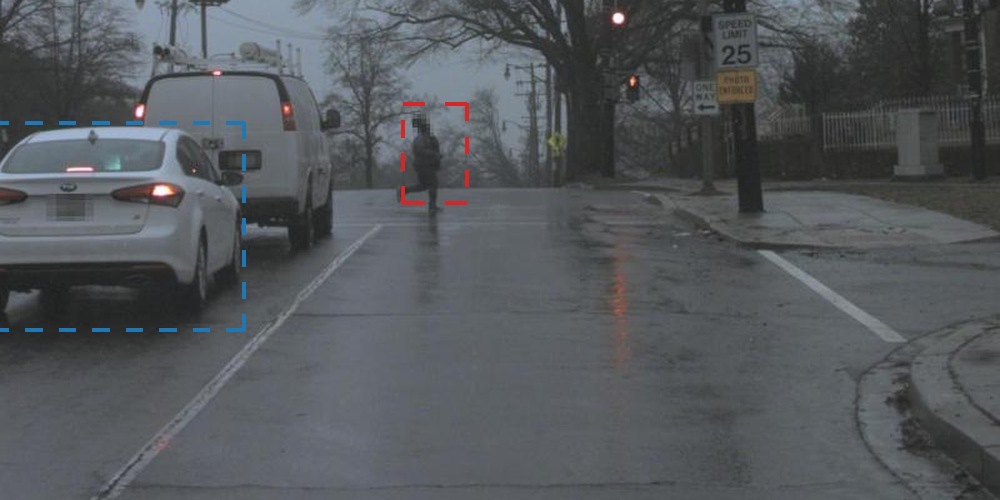}{draft: \textsc{keep}}{red!70!black} &
\qbevm{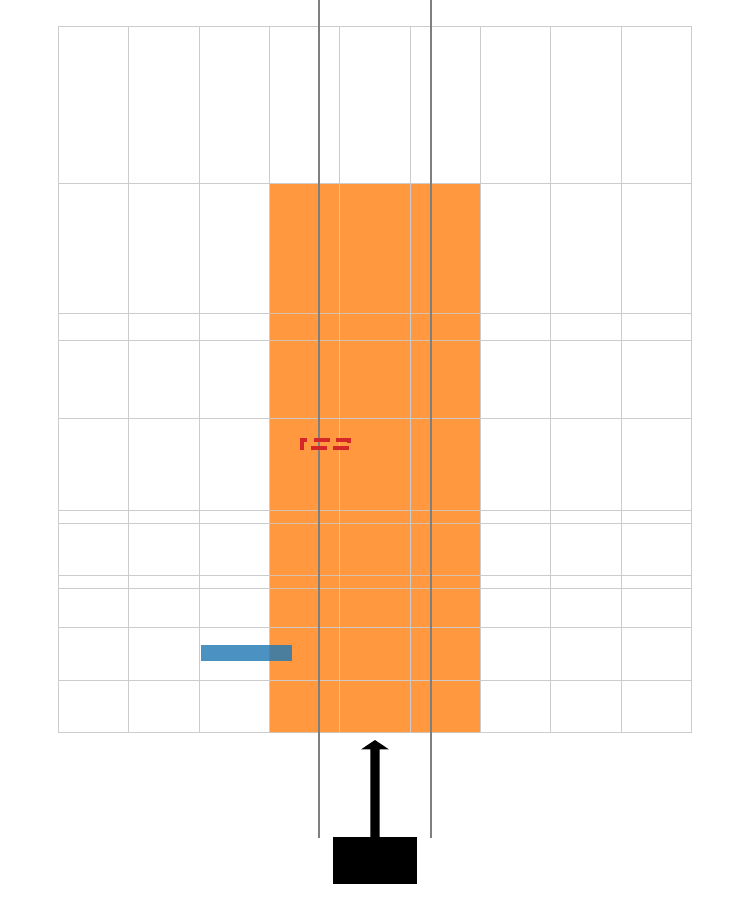}{0.4339}{0.5068} &
\qimg{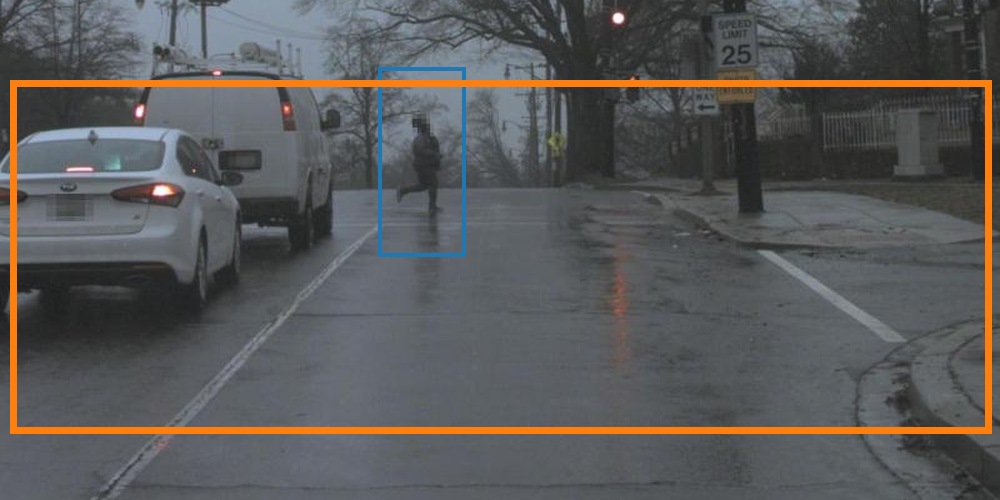}{}{black} &
\qimg{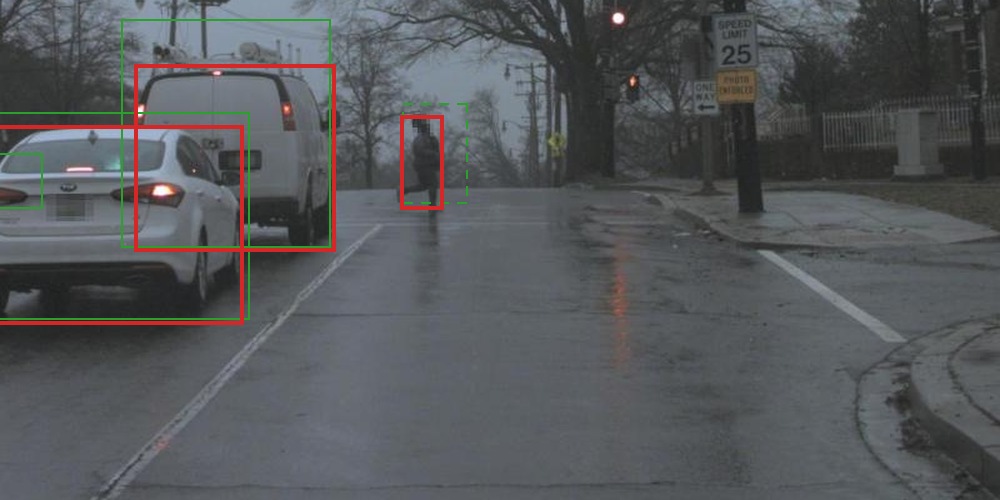}{\method: \textsc{decelerate} (required)}{green!40!black} \\[2pt]
\rotatebox{90}{\parbox{0.1225\textwidth}{\centering\footnotesize (b) support\\verification}} &
\qimg{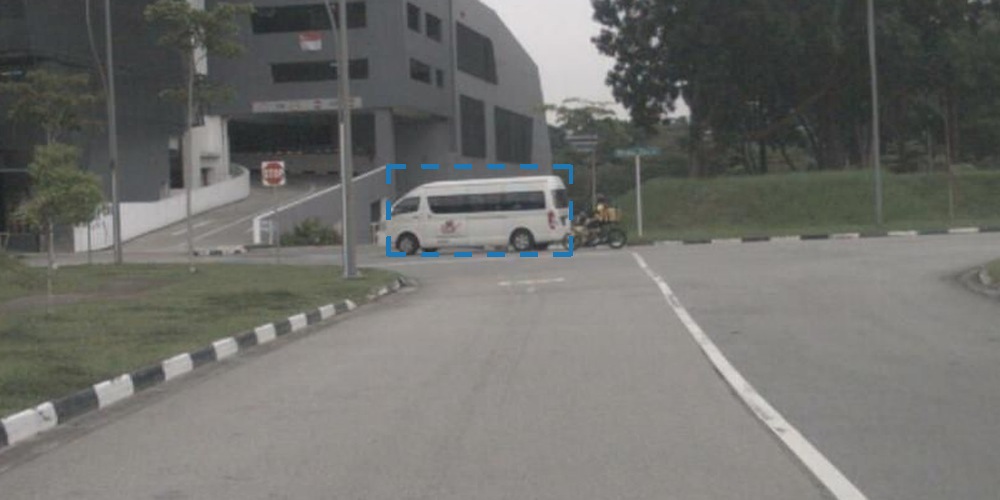}{draft: \textsc{decelerate}}{red!70!black} &
\qbev{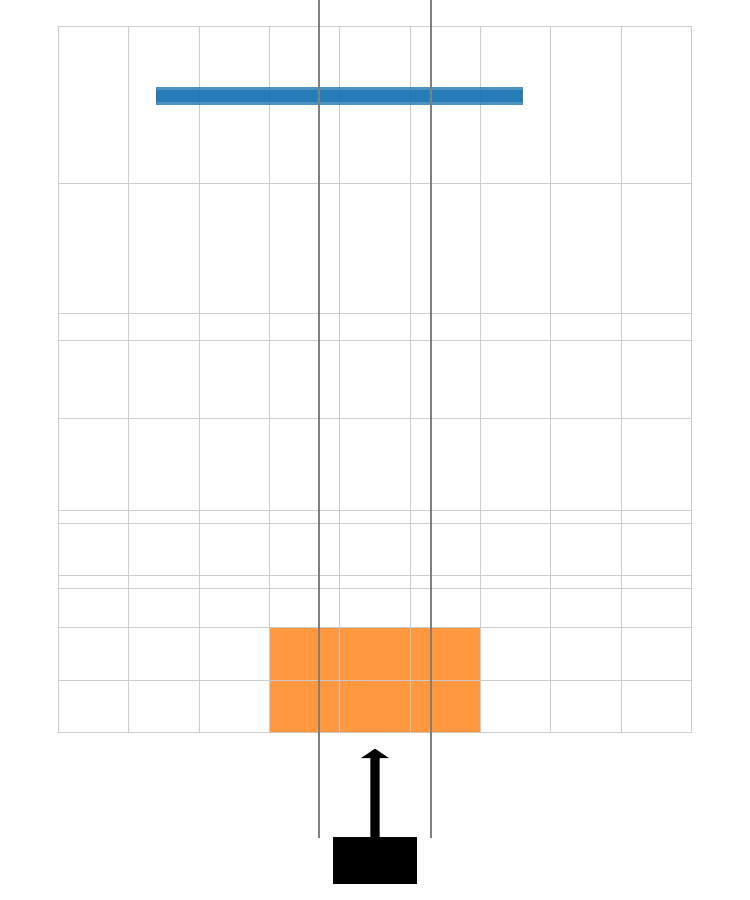} &
\qimg{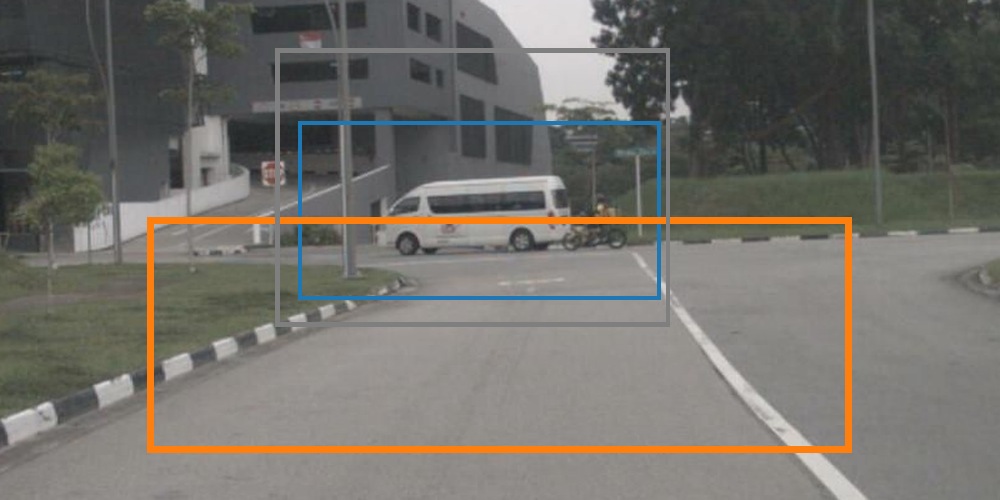}{}{black} &
\qimg{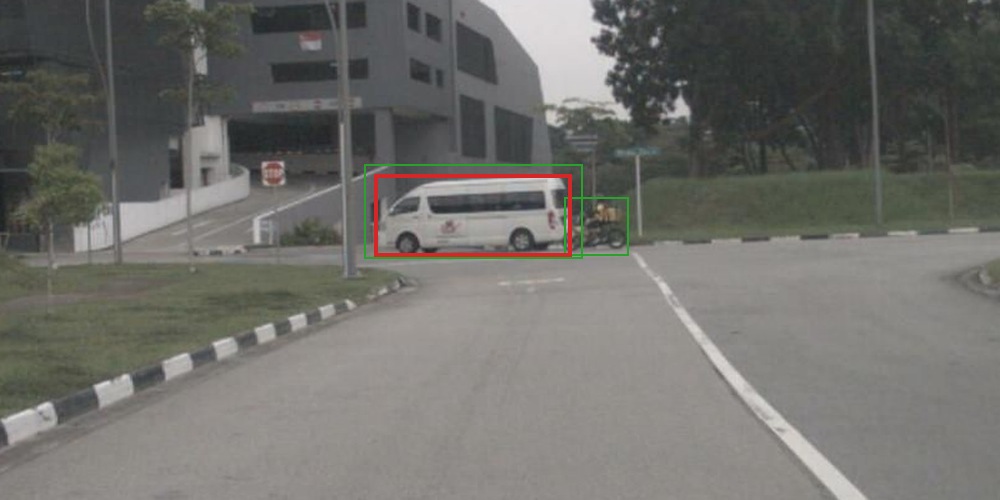}{\method: \textsc{keep} (required)}{green!40!black} \\
\end{tabular}\hspace{5pt}%
\begin{tikzpicture}[baseline=(current bounding box.center)]
\qlg{0}{qB}{draft box,\\[-1pt]support}
\qlg{-0.62}{qO}{field,\\[-1pt]footprint}
\qlg{-1.12}{qG}{guard crop}
\qlg{-1.62}{qR}{verified\\[-1pt]detection}
\qlg{-2.12}{qN}{annotation}
\qlg{-2.62}{qN, dash pattern=on 2pt off 1.3pt}{relevant\\[-1pt]annotation}
\qlg{-3.24}{qR, dash pattern=on 2pt off 1.3pt}{missed by\\[-1pt]the draft}
\end{tikzpicture}
\caption{\textbf{Complementary roles of the field and the support.} Test frames of Argoverse~2 (a) and nuScenes (b) (cropped views; selection: \supp{sec:supp_qual}). Dashed blue: replaced draft boxes. (a)~The footprint covers a pedestrian 37\,m ahead that the draft misses; once found, it joins the support (blue crop), and \textsc{keep} becomes the required \textsc{decelerate}. (b)~Verifying the support restores the required \textsc{keep}.}
\label{fig:qual}
\end{figure*}

\subsection{What field and support contribute}
\label{sec:exp_ablation}

\begin{table}[t]
\centering
\footnotesize
\setlength{\tabcolsep}{1.88pt}
\begin{tabular}{@{}lccccccc@{}}
\toprule
 & & \multicolumn{2}{c}{nuScenes} & \multicolumn{2}{c}{Argoverse 2} & \multicolumn{2}{c}{Argoverse-HD} \\
\cmidrule(lr){3-4}\cmidrule(lr){5-6}\cmidrule(l){7-8}
\raisebox{6.9pt}[0pt][0pt]{Variant} & \raisebox{6.9pt}[0pt][0pt]{px\dn} & Rep.\up & F1\up & Rep.\up & F1\up & Rep.\up & F1\up \\
\midrule
\oursrow\textbf{\method (full)} & 7.1 & \fst{86.9} & \fst{69.0} & \fst{91.7} & \fst{77.5} & \fst{92.0} & \ourslast \fst{46.7} \\
\midrule
Support only (w/o field) & 5.0 & 76.4\sgu{} & 61.9\sgu{} & 84.1\sgu{} & \thd{75.3} & 86.5\sgu{} & \thd{44.1} \\
Field only (w/o support) & 5.7 & 44.0\sgu{} & 48.6\sgu{} & 39.4\sgu{} & 44.6\sgu{} & 40.2\sgu{} & 19.0\sgu{} \\
Pivots instead of support & 6.9 & \thd{85.6\sgu{}} & \fst{69.0} & \thd{89.5\sgu{}} & 74.7\sgu{} & 88.4\sgu{} & 32.5\sgu{} \\
Single round & 6.8 & \scd{86.8} & \scd{67.4}\sgu{} & \scd{91.5} & \scd{77.0} & \scd{91.6} & \scd{44.2} \\
W/o truncation guard & 6.2 & 81.4\sgu{} & \thd{64.0\sgu{}} & 87.1\sgu{} & 72.3\sgu{} & \thd{90.7\sgu{}} & 40.0\sgu{} \\
W/o range refiner & 7.1 & 73.0\sgu{} & 63.0\sgu{} & 79.6\sgu{} & 72.9\sgu{} & 84.7\sgu{} & \scd{44.2} \\
\bottomrule
\end{tabular}
\caption{\textbf{Ablation of the mechanisms.} Removing any mechanism lowers the repair rate. px: mean over the datasets (\% of native); Rep.: repair rate (\%). \rankkey{} among the variants; \sgu{}: significantly below the full procedure.}
\label{tab:ablation}
\end{table}

\begin{figure}[t]
\centering
\includegraphics[width=\linewidth]{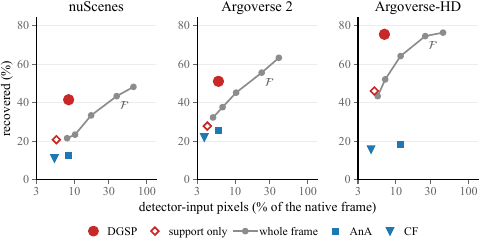}
\caption{\textbf{Recovering what the draft misses.} \method approaches the reference $\full$ with about a quarter of its pixels. Share of decision-relevant objects missed by the draft that are found.}
\label{fig:recovery}
\end{figure}

\begin{table}[t]
\centering
\footnotesize
\setlength{\tabcolsep}{3.56pt}
\begin{tabular}{@{}lccccccc@{}}
\toprule
 & \multicolumn{3}{c}{Curved path: Agr.\up} & \multicolumn{3}{c}{Curved path: F1\up} & Selector \\
\cmidrule(lr){2-4}\cmidrule(lr){5-7}\cmidrule(l){8-8}
\raisebox{6.9pt}[0pt][0pt]{Regions} & nuSc. & AV2 & A-HD & nuSc. & AV2 & A-HD & Agr.\up \\
\midrule
\rf{Whole frame} & \rf{\underline{97.6}\sgu{}} & \rf{\underline{96.2}\sgu{}} & \rf{\underline{94.7}\sgu{}} & \rf{\underline{55.4}} & \rf{\underline{69.2}\sgu{}} & \rf{\underline{46.3}} & \rf{94.1\sgu{}} \\
\midrule
\method, braking & 97.1\sgu{} & 94.6\sgu{} & \underline{94.7}\sgu{} & 54.6 & 65.4\sgu{} & 38.4\sgu{} & \underline{96.5}\sgu{} \\
\oursrow\textbf{\method}, own & \textbf{98.5} & \textbf{97.5} & \textbf{97.1} & \textbf{58.6} & \textbf{73.3} & \textbf{50.7} & \ourslast \textbf{97.8} \\
\bottomrule
\end{tabular}
\caption{\textbf{Other planners.} With its own regions, \method is the most faithful under both planners. Whole frame: u512; braking\,/\,own: regions of the straight-path\,/\,current planner. \rankkey; \sgu{}: last row significantly higher.}
\label{tab:planners}
\end{table}

\begin{table}[t]
\centering
\footnotesize
\setlength{\tabcolsep}{15.27pt}
\begin{tabular}{@{}lcccc@{}}
\toprule
Method & px\dn & Miss.\dn & Hard\dn & Prev.\dn \\
\midrule
\rf{u320 (draft)} & \rf{4.3} & \rf{0.44\sgu{}} & \rf{4.25\sgu{}} & \rf{1.5} \\
\rf{u512} & \rf{10.2} & \rf{0.35\sgu{}} & \rf{\underline{3.90}\sgu{}} & \rf{1.7\sgu{}} \\
\rf{u960 ($\full$)} & \rf{38.4} & \rf{\underline{0.33}} & \rf{3.97\sgu{}} & \rf{\underline{1.1}} \\
\midrule
\oursrow\textbf{\method} & 8.1 & \textbf{0.28} & \textbf{3.07} & \ourslast \textbf{0.4} \\
\bottomrule
\end{tabular}
\caption{\textbf{Controlled closed loop (155 routes${}\times{}$three seeds).} Miss.\,/\,Hard: missed required episodes\,/\,unnecessary hard-braking onsets per rollout; Prev.: rollouts with a preventable collision (\%). \rankkey; \sgu{}: \method significantly better.}
\label{tab:closedloop}
\end{table}

\noindent\textbf{Each mechanism addresses a distinct failure (\Cref{tab:ablation}).}
Without the field, which reaches the objects that the draft lacks, the repair rate falls by 5--11 points and the recall by 6\nobreakdash--14 points (all significant; \supp{sec:supp_field}).
Without the support, which verifies the detections on which the action rests, only 39--44\% of the disagreements are repaired, and pivots in its place significantly lower the repair rate on all three datasets and the F1 on two.
The truncation guard significantly improves both metrics, and a second round raises the F1 on nuScenes and the held-out sequences (\supp{sec:supp_results}).
The range refiner, outside the guarantee, raises the F1 by 2.5--6.0 points; its repair-rate gain is measured against a reference that also uses it.

\noindent\textbf{Probing reaches objects that the draft never detected (\Cref{fig:qual,fig:recovery}).}
Of the decision-relevant objects (annotated objects that alone require braking) that the draft misses, \method finds 41\%, 51\%, and 75\%, including 85\nobreakdash--97\% of those the reference detects: 1.6--2$\times$ as many as the support alone and significantly more than u512 (with more pixels), AnA, and CF.
With all relevant draft detections deleted, it still finds 93\nobreakdash--97\%, against 21--26\% for AnA and CF. The same crops placed at random repair only 20\nobreakdash--29\% and recover 8--18\%; crops around the annotated relevant objects recover about as many as \method but agree on only 88--95\%, as they leave spurious detections \mbox{unverified}. Uncertainty crops with \method's crop pixels repair less than u512. Of pedestrians and cyclists, \method finds 24\nobreakdash--45\%, against 0--21\% for u512 (\supp{sec:supp_field}).

\noindent\textbf{Field and support follow from the planner (\Cref{tab:planners}).}
We apply \method to a braking controller on a curved path and to a bottleneck selector over 15 fixed trajectories on NAVSIM navtest~\cite{navsim}, each with its own field and certifying support.
On the curved path, the planner's own regions reproduce its reference action significantly more often than u512 (at 0.75--0.89$\times$ its pixels) and than the braking regions, whose F1 they raise by 4--12 points to the reference's level (n.s.).
Under a time-to-collision warning, they are no more faithful than u512, and less so on Argoverse-HD (\supp{sec:supp_field}).
On the selector (\Cref{cor:bottleneck}), \method agrees with $\policy(\full)$ on 97.8\% of the 12,146 scenes, against 94.0\% for the draft and 95.9\% at native resolution.
PDM-Closed~\cite{pdm}, outside the guarantee, is probed as a black box: \method selects a trajectory within 0.5\,m of the reference's on 97.2\% of these scenes (91.2\% for the draft). On both planners, it uses two thirds of the reference's pixels (\supp{sec:consumers_outside}).

\subsection{Closed-loop braking}
\label{sec:exp_planning}

We run a controlled closed loop in CARLA~\cite{carla} on the 155 braking routes of Bench2Drive~\cite{bench2drive}: a route follower steers, and the braking controller acts on each pipeline's estimated scene, without the learned parts of the lift (\supp{sec:supp_embedded}).
\method has significantly fewer unnecessary hard-braking onsets than u320, u512, and the reference (\Cref{tab:closedloop}) and misses the fewest episodes, significantly fewer than u320 and u512; its preventable collisions are the fewest.

\section{Conclusion}
\label{sec:conclusion}

An image cannot show what it failed to resolve, so \method asks the planner where an unseen object would change the action and verifies these places together with the support of the action; under explicit conditions, this provably returns the action of whole-frame high-resolution perception.
On three driving datasets, it reproduces this action on 97.5\nobreakdash--98.7\% of frames with 22--27\% of its detector-input pixels, has the highest braking F1, and leads uniform downscaling most where the draft misses the decisive object, by 13--17 points over u512.
Because verification runs on crops, it meets these conditions only approximately, and its temporal estimate needs a history at one scale, which crops provide for fewer objects than whole frames do, most visibly under a time-to-collision warning; \method also trades some recall for precision.
Because the regions follow the planner, the saving grows with how concentrated the planner's sensitivity is: the braking field is a short corridor ahead of the vehicle.
The guarantee covers max-composable and bottleneck planners; general planners over object lists can be probed but lie outside it, and our single-camera results characterize decision quality and controlled braking, not driving safety.
Within this scope, a planner that can be queried about unseen objects turns its decision rule into a specification of what perception must verify.

{
  \small
  \bibliographystyle{ieeenat_fullname}
  \bibliography{main}
}

\clearpage
\onecolumn
\appendix
\renewcommand{\theHsection}{appendix.\Alph{section}}
\renewcommand{\ci}[2]{[#1,\allowbreak\,#2]}
\renewcommand{\rf}[1]{#1}
\setlength{\parindent}{1pc}\setlength{\emergencystretch}{1.5em}
\setlength{\abovecaptionskip}{4pt}\setlength{\belowcaptionskip}{0pt}
\setlength{\textfloatsep}{14pt plus 4pt minus 2pt}\setlength{\floatsep}{12pt plus 4pt minus 2pt}\setlength{\intextsep}{12pt plus 4pt minus 2pt}
\renewcommand{\arraystretch}{1.08}
\renewcommand{\floatpagefraction}{0.6}\renewcommand{\topfraction}{0.85}\renewcommand{\bottomfraction}{0.6}\renewcommand{\textfraction}{0.12}
\makeatletter\setlength{\@fptop}{0pt}\setlength{\@fpsep}{16pt plus 2fil}\setlength{\@fpbot}{0pt plus 1fil}\makeatother
\setlist{itemsep=2pt, topsep=3pt, parsep=0pt}
\raggedbottom
\begin{center}{\Large\bfseries Appendix\par}\end{center}
\vspace{6pt}

This appendix gives the complete proofs (\cref{S-sec:supp_theory}), the implementation and the evaluation protocol needed to reproduce the results (\cref{S-sec:supp_impl}), the full tables behind \cref{sec:exp_exact,sec:exp_main} (\cref{S-sec:supp_results}), analyses of the field and the support (\cref{S-sec:supp_field}), the controlled closed loop (\cref{S-sec:supp_embedded}), and further qualitative examples (\cref{S-sec:supp_qual}).
Notation and terms are those of the main paper. Unless stated otherwise, results are on the test splits, intervals are $95\%$ paired bootstrap intervals ($2{,}000$ resamples over scenes, logs, or sequences), and $^{\dagger}$ / $^{\ast}$ mark a value of \method that is significantly higher / lower than that of the row.

\section{Formal guarantees}
\label{S-sec:supp_theory}

\subsection{Setting and notation}
\label{S-sec:setting_supp}

A \emph{scene} is a finite set of objects; each object $o$ has a box $b(o)$ in image coordinates and a state $\state(o)$ in the ego frame (range, lateral extent, and time to collision).
Objects with equal states are distinct, and a \emph{probe} $o_{\state}$ is a hypothetical object with state $\state$.
A \emph{planner} $\policy$ maps a scene to an action through the states of its objects; the ego state is fixed within a frame and omitted.

\begin{itemize}
\item \emph{Field.} The field of a scene $\scene$ contains the states of a single inserted object that changes the action,
\begin{equation}
\field(\scene)=\{\,\state \,:\, \policy(\scene\cup\{o_{\state}\})\neq\policy(\scene)\,\}.
\label{S-eq:s_field}
\end{equation}
\item \emph{Insertion stability.} $\policy$ is insertion-stable if, for every scene $\scene$ and every finite set $\mathcal{X}$ of objects whose states lie outside $\field(\scene)$,
\begin{equation}
\policy(\scene\cup \mathcal{X})=\policy(\scene).
\label{S-eq:s_insertion}
\end{equation}
\item \emph{Certifying support.} Let $\mathcal{Y}(\scene)$ be the family that contains the empty set and every singleton $\{o\}$ with $\state(o)\notin\field(\scene)$. A support $\Sup(\scene)\subseteq \scene$ is certifying if
\begin{equation}
\policy\big((\scene\setminus \mathcal{D})\cup \mathcal{O}\big)=\policy(\scene)\qquad\text{for all } \mathcal{D}\subseteq \scene\setminus\Sup(\scene)\ \text{and}\ \mathcal{O}\in\mathcal{Y}(\scene).
\label{S-eq:s_support}
\end{equation}
\item \emph{Verified region and merged scene.} A crop is an axis-aligned image rectangle, and a verified region $\reg$ is a finite set of crops; $\scene|_{\reg}$ denotes the objects of a scene $\scene$ whose box center lies in a crop of $\reg$.
$\ver(\reg)=\bigcup_{w\in \reg}\detH(w)$ denotes what verification returns: the reference detector runs on each crop of $\reg$, in the order in which Algorithm~\ref{alg:loop} adds them, the boxes are mapped to image coordinates, and duplicates of overlapping crops are merged.
$\ver$ is a function of the crops, not only of their union, since a detection depends on the crop that contains it. The merged scene is
\begin{equation}
\merged(\reg)=(\draft\setminus\draft|_{\reg})\cup \ver(\reg).
\label{S-eq:s_merged}
\end{equation}
\item \emph{Requested crops.} The crops requested for a scene $\scene$ are its field footprint $\foot(\scene)$, a single crop for the braking controllers (the planners of \cref{S-sec:consumers_outside} use one crop per cluster of sensitive cells; the proof uses only the union), and the set $\supcrops(\scene)$, one crop per support object, which contains the box center of that object. Algorithm~\ref{alg:loop} stops when every crop requested for the current merged scene lies inside the union of the crops of $\reg$.
\item \emph{Locality.} $\ver(\reg)=\full|_{\reg}$, with identical boxes and states. This is a condition on verification, not a definition of $\ver$: $\full|_{\reg}$ restricts the hypothetical whole-frame reference scene, whereas $\ver(\reg)$ is computed on crops.
\item \emph{Coverage.} Every object of $\full$ with a state in $\field(\merged)$ has its box center in $\reg$.
\end{itemize}

\begin{restateddef}{Definition}{\ref{def:maxcomp}}{Max-composable planner}
Let the actions $a\in\{0,\dots,K\}$ be ordered by intensity, with thresholds $0=\tau_0<\tau_1<\dots<\tau_K<\tau_{K+1}=\infty$.
A planner $\policy$ is max-composable if, for a per-object demand $\dem\ge0$,
\begin{equation}
\policy(\scene)=\mathsf{q}\Big(\max_{o\in \scene}\dem\big(\state(o)\big)\Big),\qquad \mathsf{q}(u)=\max\{a:\,u\ge\tau_a\},\qquad \max\varnothing=0.
\label{S-eq:s_maxcomp}
\end{equation}
\end{restateddef}

The braking controller of the experiments is max-composable with $K=2$, the actions \textsc{keep}~($0$), \textsc{decelerate}~($1$), and \textsc{hard-brake}~($2$), and $\tau_1=1.0$ and $\tau_2=3.5$\,m/s$^2$ (\cref{S-sec:supp_metric}).

\subsection{Decision exactness (Theorem~\protect\ref{thm:exact})}
\label{S-sec:proof_thm}

\begin{restated}{Theorem}{\ref{thm:exact}}{Decision exactness}
Let Algorithm~\ref{alg:loop} stop with verified region $\reg$ and merged scene $\merged$.
It returns $\policy(\full)$ if (i)~$\policy$ is insertion-stable; (ii)~$\Sup$ is certifying; (iii)~\emph{locality}: $\ver(\reg)=\full|_{\reg}$, with identical boxes and states; and (iv)~\emph{coverage}: every object of $\full$ whose state lies in $\field(\merged)$ has its box center in $\reg$.
\end{restated}

\begin{strategy}
The proof first removes the unverified draft objects using the certification by the support (Step~3) and then inserts the remaining reference objects using insertion stability (Step~4).
\end{strategy}

\begin{proof}
\emph{Step 0: the state at termination.}
Let $\reg$ be the verified region at termination and $\merged=\merged(\reg)$.
Since the loop stopped, every crop requested for $\merged$ lies inside the union of the crops of $\reg$, which the inclusion below abbreviates:
\begin{equation}
\reg\;\supseteq\;\foot(\merged)\,\cup\,\supcrops(\merged).
\label{S-eq:s_term}
\end{equation}

\emph{Step 1: the support has been verified.}
By locality, $\ver(\reg)=\full|_{\reg}$, so that \cref{S-eq:s_merged} gives $\merged=(\draft\setminus\draft|_{\reg})\cup\full|_{\reg}$, and the objects of $\merged$ whose box center lies in $\reg$ are exactly those of $\full|_{\reg}$.
Every support object has its box center in its crop and hence, by~\cref{S-eq:s_term}, in $\reg$, so that $\Sup(\merged)\subseteq\full|_{\reg}$.
The unverified part of the merged scene therefore lies outside the support:
\begin{equation}
\mathcal{D}\;\coloneqq\;\draft\setminus\draft|_{\reg}\;=\;\merged\setminus\full|_{\reg}\;\subseteq\;\merged\setminus\Sup(\merged).
\label{S-eq:s_D}
\end{equation}

\emph{Step 2: the unverified reference objects lie outside the field of $\merged$.}
Let $\bar\full=\full\setminus\full|_{\reg}$.
Every $o\in\bar\full$ has its box center outside $\reg$, and by coverage,
\begin{equation}
\state(o)\notin\field(\merged)\qquad\text{for all } o\in\bar\full .
\label{S-eq:s_outside}
\end{equation}

\emph{Step 3: certification transfers the action to $\full|_{\reg}$.}
We apply~\cref{S-eq:s_support} to $\scene=\merged$ with the set $\mathcal{D}$ of~\cref{S-eq:s_D}, for which $\merged\setminus \mathcal{D}=\full|_{\reg}$.
With $\mathcal{O}=\varnothing$, and with $\mathcal{O}=\{o\}$ for $o\in\bar\full$, which belongs to $\mathcal{Y}(\merged)$ by~\cref{S-eq:s_outside},
\begin{align}
\policy(\full|_{\reg})&=\policy(\merged),\label{S-eq:s_step3a}\\
\policy\big(\full|_{\reg}\cup\{o\}\big)&=\policy(\merged)=\policy(\full|_{\reg})\qquad\text{for all } o\in\bar\full .\label{S-eq:s_step3b}
\end{align}
Since a planner sees objects only through their states, \cref{S-eq:s_field,S-eq:s_step3b} give $\state(o)\notin\field(\full|_{\reg})$ for every $o\in\bar\full$.

\emph{Step 4: insertion stability transfers the action to $\full$.}
Since $\bar\full$ is finite and all its states lie outside $\field(\full|_{\reg})$, insertion stability~\cref{S-eq:s_insertion} with $\scene=\full|_{\reg}$ and $\mathcal{X}=\bar\full$ gives, together with~\cref{S-eq:s_step3a},
\begin{equation}
\policy(\full)=\policy(\full|_{\reg}\cup\bar\full)=\policy(\full|_{\reg})=\policy(\merged).
\end{equation}
The algorithm returns $\policy(\merged)$, which proves the claim.
If $\bar\full=\varnothing$, Step~4 is vacuous and~\cref{S-eq:s_step3a} suffices.
\end{proof}

\begin{remark}[Where the conditions enter]
Locality is used in Step~1 to identify the verified objects with $\full|_{\reg}$, coverage in Step~2, certification in Step~3, and insertion stability in Step~4.
Certification is needed because the merged scene contains unverified draft objects, which may be spurious or mislocalized; insertion stability is needed because $\bar\full$ may contain arbitrarily many objects that the procedure never observes.
Coverage is a condition on the reference scene, not on the probe grid: a finite grid can still satisfy it, as long as every reference object whose state lies in the field has its box center in the verified region; \cref{S-sec:supp_coverage} measures how often this holds for the decisive object. The proof uses the footprint only through coverage: the footprint is the means of reaching it.
\end{remark}

\subsection{Max-composable planners (Corollary~\protect\ref{cor:maxcomp})}
\label{S-sec:proof_maxcomp}

\begin{restated}{Corollary}{\ref{cor:maxcomp}}{Max-composable planners}
A max-composable planner is insertion-stable with $\field(\scene)=\{\state:\dem(\state)\ge\tau_{a+1}\}$, where $a=\policy(\scene)$, and $\Sup(\scene)=\{o\in \scene:\dem(\state(o))\ge\tau_a\}$ for $a\ge1$ and $\Sup(\scene)=\varnothing$ for $a=0$ is certifying.
\end{restated}

\begin{strategy}
The action depends on the scene only through the maximum demand $m$.
An inserted object can only raise $m$, and it changes the action exactly if it raises $m$ to $\tau_{a+1}$ or above; removing objects can only lower $m$, and the action is retained as long as one object with demand at least $\tau_a$ remains.
For $a=0$, the lower bound $\tau_0=0$ holds for every scene, so that no object needs to remain.
\end{strategy}

\begin{proof}
Let $a=\policy(\scene)$ and $m=\max_{o\in \scene}\dem(\state(o))$, with $m=0$ if $\scene=\varnothing$.
By~\cref{S-eq:s_maxcomp},
\begin{equation}
\tau_a\;\le\; m\;<\;\tau_{a+1}.
\label{S-eq:s_bracket}
\end{equation}

\emph{(i) Field.}
For a state $\state$, $\policy(\scene\cup\{o_{\state}\})=\mathsf{q}\big(\max(m,\dem(\state))\big)$.
Since $\max(m,\dem(\state))\ge m\ge\tau_a$, the action cannot decrease, and since $m<\tau_{a+1}$,
\begin{equation}
\policy(\scene\cup\{o_{\state}\})\neq a\iff\max(m,\dem(\state))\ge\tau_{a+1}\iff\dem(\state)\ge\tau_{a+1}.
\end{equation}
Hence $\field(\scene)=\{\state:\dem(\state)\ge\tau_{a+1}\}$; for $a=K$, $\tau_{K+1}=\infty$ and the field is empty.

\emph{(ii) Insertion stability.}
If $\dem(\state)<\tau_{a+1}$ for every state $\state$ of a finite set $\mathcal{X}$, then
\begin{equation}
\tau_a\;\le\; m\;\le\;\max\Big(m,\max_{\state\in \mathcal{X}}\dem(\state)\Big)\;<\;\tau_{a+1},
\end{equation}
so that $\policy(\scene\cup \mathcal{X})=a$.

\emph{(iii) Certification.}
Let $\mathcal{T}=\{o\in \scene:\dem(\state(o))\ge\tau_a\}$ if $a\ge1$ and $\mathcal{T}=\varnothing$ if $a=0$.
Let $\mathcal{D}\subseteq \scene\setminus \mathcal{T}$, $\mathcal{O}\in\mathcal{Y}(\scene)$, and $\mathcal{W}=(\scene\setminus \mathcal{D})\cup \mathcal{O}$; we show $\tau_a\le\max_{o\in \mathcal{W}}\dem(\state(o))<\tau_{a+1}$.
\begin{itemize}
\item \emph{Upper bound.} Every object of $\scene$ has a demand of at most $m<\tau_{a+1}$, and the object of $\mathcal{O}$, if any, has a state outside $\field(\scene)$ and hence, by~(i), a demand below $\tau_{a+1}$.
\item \emph{Lower bound, $a\ge1$.} Since $m\ge\tau_a>0$, $\scene$ is nonempty, and since it is finite, an object $o^\ast$ with $\dem(\state(o^\ast))=m$ exists. It belongs to $\mathcal{T}$ and, since $\mathcal{D}\cap \mathcal{T}=\varnothing$, to $\mathcal{W}$.
\item \emph{Lower bound, $a=0$.} The bound $\tau_0=0\le\max_{o\in \mathcal{W}}\dem(\state(o))$ holds for every scene, including $\mathcal{W}=\varnothing$.
\end{itemize}
Hence $\policy(\mathcal{W})=a$ for all such $\mathcal{D}$ and $\mathcal{O}$, \ie, $\Sup(\scene)=\mathcal{T}$ satisfies~\cref{S-eq:s_support}.
\end{proof}

\begin{remark}[The case \textsc{keep}]
For $a=0$, the formula $\{o\in \scene:\dem(\state(o))\ge\tau_0\}$ would contain every object of the scene, since $\tau_0=0$.
This set is also certifying but needlessly large: removing objects cannot raise the action, and an inserted object outside the field does not raise it either, so that the empty support suffices.
At \textsc{keep}, Algorithm~\ref{alg:loop} therefore requests no support crops, and its verified region consists of the footprint of the field only.
\end{remark}

\begin{remark}[Smaller supports and the deployed choice]
For $a\ge1$, the certification uses a single object of $\mathcal{T}$ with demand at least $\tau_a$, so that every subset of $\scene$ that contains such an object is certifying (this also follows from \cref{S-lem:witness} below).
The deployed system verifies all of $\mathcal{T}$, which also provides the objects to which the range refiner is applied.
\end{remark}

\subsection{Bottleneck selectors (Corollary~\protect\ref{cor:bottleneck})}
\label{S-sec:proof_bottleneck}

Let $\mathcal{K}$ be a fixed finite set of candidates with a total order $\prec$, let $\beta:\mathcal{K}\to\bar{\mathbb{R}}$ be a preference and $\mathsf{h}$ a per-object score, and let
\begin{equation}
\mathsf{s}(c,\scene)=\min\Big(\beta(c),\,\min_{o\in \scene}\mathsf{h}\big(c,\state(o)\big)\Big),\qquad \min\varnothing=+\infty .
\label{S-eq:s_score}
\end{equation}
The planner selects the first maximizer with respect to $\prec$, \ie, $\policy(\scene)=c^\ast$ if and only if
\begin{equation}
\mathsf{s}(c^\ast,\scene)\ge \mathsf{s}(c,\scene)\ \ \text{for all }c\in\mathcal{K},\qquad \mathsf{s}(c^\ast,\scene)>\mathsf{s}(c,\scene)\ \ \text{for all }c\prec c^\ast.
\label{S-eq:s_first}
\end{equation}
Two properties of~\cref{S-eq:s_score} are used throughout: scores compose by the minimum, and removing objects cannot decrease a score,
\begin{equation}
\mathsf{s}(c,\scene\cup \scene')=\min\big(\mathsf{s}(c,\scene),\mathsf{s}(c,\scene')\big),\qquad \scene'\subseteq \scene\;\Rightarrow\;\mathsf{s}(c,\scene')\ge \mathsf{s}(c,\scene).
\label{S-eq:s_props}
\end{equation}

\begin{restated}{Corollary}{\ref{cor:bottleneck}}{Bottleneck selection}
A bottleneck selector is insertion-stable, and $\Sup(\scene)=\{o\in \scene:\mathsf{h}(c,\state(o))\le \mathsf{s}(\policy(\scene),\scene)\text{ for some }c\neq\policy(\scene)\}$ is certifying.
\end{restated}

The corollary follows from Theorem~\ref{thm:exact} and the following two lemmas.

\begin{lemma}[Insertion stability]
\label{S-lem:insertion}
If $\policy(\scene\cup\{o\})=\policy(\scene)$ for every object $o$ of a finite nonempty set $\mathcal{X}$, then $\policy(\scene\cup \mathcal{X})=\policy(\scene)$.
\end{lemma}

\begin{strategy}
By~\cref{S-eq:s_props}, the scores of $\scene\cup \mathcal{X}$ are the minima of the scores of the single insertions. The single insertion that lowers the score of $c^\ast$ most bounds every other candidate from above.
\end{strategy}

\begin{proof}
Let $c^\ast=\policy(\scene)$. By~\cref{S-eq:s_props}, $\mathsf{s}(c,\scene\cup \mathcal{X})=\min_{o\in \mathcal{X}}\mathsf{s}(c,\scene\cup\{o\})$ for every $c$.
Let $o_0\in \mathcal{X}$ minimize $\mathsf{s}(c^\ast,\scene\cup\{o\})$, so that $\mathsf{s}(c^\ast,\scene\cup \mathcal{X})=\mathsf{s}(c^\ast,\scene\cup\{o_0\})$.
For every $c$,
\begin{equation}
\mathsf{s}(c,\scene\cup \mathcal{X})\;\le\; \mathsf{s}(c,\scene\cup\{o_0\})\;\le\; \mathsf{s}(c^\ast,\scene\cup\{o_0\})\;=\;\mathsf{s}(c^\ast,\scene\cup \mathcal{X}),
\end{equation}
where the second inequality holds since $c^\ast=\policy(\scene\cup\{o_0\})$.
If equality holds throughout for some $c\neq c^\ast$, then $c$ ties with $c^\ast$ in $\scene\cup\{o_0\}$, and by~\cref{S-eq:s_first} for $\scene\cup\{o_0\}$, $c^\ast\prec c$.
Hence $c^\ast$ satisfies~\cref{S-eq:s_first} for $\scene\cup \mathcal{X}$.
\end{proof}

\begin{lemma}[Certification by the binding objects]
\label{S-lem:binding}
Let $c^\ast=\policy(\scene)$, $\theta=\mathsf{s}(c^\ast,\scene)$, and $\Sup(\scene)=\{o\in \scene:\mathsf{h}(c,\state(o))\le\theta\text{ for some }c\neq c^\ast\}$.
Then $\policy\big((\scene\setminus \mathcal{D})\cup \mathcal{O}\big)=c^\ast$ for every $\mathcal{D}\subseteq \scene\setminus\Sup(\scene)$ and $\mathcal{O}\in\mathcal{Y}(\scene)$.
\end{lemma}

\begin{strategy}
An object outside the support has a score above $\theta$ for every competing candidate, so that removing it cannot raise any competing score above the score of $c^\ast$; removing objects can only raise the score of $c^\ast$.
\end{strategy}

\begin{proof}
Let $\scene^+=\scene\cup \mathcal{O}$ and $\mathcal{W}=(\scene\setminus \mathcal{D})\cup \mathcal{O}=\scene^+\setminus(\mathcal{D}\setminus\mathcal{O})$; $\policy(\scene^+)=c^\ast$ since $\mathcal{O}\in\mathcal{Y}(\scene)$.
For $c\neq c^\ast$, $\mathsf{s}(c,\scene^+)\le \mathsf{s}(c^\ast,\scene^+)\le\theta$, whereas every object $o\in \mathcal{D}$ satisfies $\mathsf{h}(c,\state(o))>\theta$; removing $\mathcal{D}\setminus\mathcal{O}$ therefore leaves the minimum unchanged,
\begin{equation}
\mathsf{s}(c,\mathcal{W})=\mathsf{s}(c,\scene^+)\qquad\text{for all }c\neq c^\ast .
\end{equation}
Moreover, $\mathsf{s}(c^\ast,\mathcal{W})\ge \mathsf{s}(c^\ast,\scene^+)$ by~\cref{S-eq:s_props}.
Hence $\mathsf{s}(c,\mathcal{W})\le \mathsf{s}(c^\ast,\mathcal{W})$ for every $c\neq c^\ast$, and equality implies $\mathsf{s}(c,\scene^+)=\mathsf{s}(c^\ast,\scene^+)$, a tie in $\scene^+$ that~\cref{S-eq:s_first} resolves in favor of $c^\ast$, \ie, $c^\ast\prec c$.
\end{proof}

\textbf{Computing the field.}
The field requires only the per-candidate scores of the probe.
With $c^\ast=\policy(\scene)$ and $\theta=\mathsf{s}(c^\ast,\scene)$,
\begin{equation}
\state\in\field(\scene)\iff\exists\,c\neq c^\ast:\ \min\big(\mathsf{s}(c,\scene),\mathsf{h}(c,\state)\big)>\mathsf{h}(c^\ast,\state),\ \text{or equality holds and}\ c\prec c^\ast.
\label{S-eq:s_bnfield}
\end{equation}
If $\mathsf{h}(c^\ast,\state)<\theta$, the score of $c^\ast$ after the insertion is $\mathsf{h}(c^\ast,\state)$, and~\cref{S-eq:s_bnfield} is the condition that another candidate overtakes it.
If $\mathsf{h}(c^\ast,\state)\ge\theta$, the insertion leaves the score of $c^\ast$ at $\theta$, every competitor stays at most $\theta$, and strictly below it if $c\prec c^\ast$; neither branch of~\cref{S-eq:s_bnfield} can then hold, in agreement with $\state\notin\field(\scene)$.
Each cell thus costs $|\mathcal{K}|$ score evaluations and no re-planning, and single-object probes suffice by \cref{S-lem:insertion}.
The support, in contrast, requires a planner that exposes the scores $\mathsf{h}$ of all candidates.

\textbf{Max-composable planners as bottleneck selectors.}
With $\mathcal{K}=\{0,\dots,K\}$ ordered by $0\prec1\prec\dots\prec K$, $\beta(a)=-a$, and
\begin{equation}
\mathsf{h}(a,\state)=\begin{cases}-(K{+}1) & \text{if }\dem(\state)\ge\tau_{a+1},\\ +\infty & \text{otherwise,}\end{cases}
\end{equation}
so that
\begin{equation}
\mathsf{s}(a,\scene)=\begin{cases}-a & \text{if }\max_{o\in \scene}\dem(\state(o))<\tau_{a+1},\\ -(K{+}1) & \text{otherwise,}\end{cases}
\end{equation}
and
the unique maximizer is $\mathsf{q}(\max_{o\in \scene}\dem(\state(o)))$, independently of the order.
\Cref{S-lem:insertion} therefore contains the insertion stability of Corollary~\ref{cor:maxcomp}.
The binding set of \cref{S-lem:binding} is $\{o:\dem(\state(o))\ge\tau_1\}$ for $a\ge1$ and empty for $a=0$; it contains the support of Corollary~\ref{cor:maxcomp} and, for $a\ge2$, can be strictly larger.
The smaller support suffices because of the following property.

\begin{lemma}[Witness supports]
\label{S-lem:witness}
Let $c^\ast=\policy(\scene)$ and $\theta=\mathsf{s}(c^\ast,\scene)$, and suppose that $\mathsf{s}(c^\ast,\scene\cup \mathcal{O})=\theta$ for every $\mathcal{O}\in\mathcal{Y}(\scene)$.
Then every $\mathcal{T}\subseteq \scene$ with $\mathsf{s}(c,\mathcal{T})\le\theta$ for all $c\neq c^\ast$, with strict inequality if $c\prec c^\ast$, is a certifying support.
\end{lemma}

\begin{proof}
Let $\mathcal{D}\subseteq \scene\setminus \mathcal{T}$, $\mathcal{O}\in\mathcal{Y}(\scene)$, and $\mathcal{W}=(\scene\setminus \mathcal{D})\cup \mathcal{O}$, so that $\mathcal{T}\subseteq \mathcal{W}\subseteq \scene\cup \mathcal{O}$.
By~\cref{S-eq:s_props} and the assumption, $\mathsf{s}(c^\ast,\mathcal{W})\ge \mathsf{s}(c^\ast,\scene\cup \mathcal{O})=\theta$, and $\mathsf{s}(c,\mathcal{W})\le \mathsf{s}(c,\mathcal{T})\le\theta$ for $c\neq c^\ast$, strictly if $c\prec c^\ast$.
Hence $c^\ast$ satisfies~\cref{S-eq:s_first}.
\end{proof}

For a max-composable planner, an object outside the field has $\mathsf{h}(a,\state)=+\infty$ at the selected action $a$, so that the selected score is never lowered and the assumption of \cref{S-lem:witness} holds.
Every object with demand at least $\tau_a$ bounds all lower actions at $-(K{+}1)<-a$, whereas every higher action $a'>a$ is bounded by its preference $-a'<-a$; the support of Corollary~\ref{cor:maxcomp} is therefore a witness, and for $a=0$, the empty set is one, since no action precedes $0$ and every higher action is bounded by its preference.
For a general bottleneck selector, an insertion outside the field can lower the selected score, and a witness need not be certifying (construction~(f) in \cref{S-sec:constructions}).

\subsection{Termination}
\label{S-sec:termination}

Each round of Algorithm~\ref{alg:loop} that does not stop adds at least one crop that does not lie inside the union of the previous crops. The union therefore grows strictly, by at least one image pixel, and the loop stops after finitely many rounds; this argument needs neither locality nor a bound on the number of rounds.
Under locality, the requested crops are moreover drawn from a small finite set, namely the footprints of subsets of the finite ground grid and the crops of the finitely many objects of $\draft$ and $\full$.
The experiments cap the loop at three rounds, and a frame that reaches the cap returns the action of its last merged scene; Theorem~\ref{thm:exact} covers only frames on which the loop reaches its stopping condition.
The implemented stopping test approximates containment: a requested crop counts as verified once at least $95\%$ of its area lies inside one earlier crop, and for a support crop, the guard crops also count. A support crop not clipped by the image boundary is centered on its box before the outward alignment to the stride grid, which moves each edge by at most $31$ pixels at the verification resolution; with $95\%$ of its area inside an earlier crop, it therefore has the box center of its object in that crop whenever the box is wider and taller than $4$ pixels; a footprint, however, can leave up to $5\%$ of its area unverified, which can affect coverage (\cref{S-tab:practice}).

\subsection{Counterexamples for omitted or weakened conditions}
\label{S-sec:constructions}

The following constructions show failures when the coverage of the field, the certification by the support, insertion stability, or the iteration to a fixed point is omitted or weakened; \cref{S-tab:constructions} summarizes them.
In (a)--(d), $\policy$ is max-composable, and all constructions retain locality, so they do not test whether locality can be dropped.

\begin{table}[!htbp]
\centering

\small
\begin{tabular}{@{}llll@{}}
\toprule
 & Procedure or planner & Condition violated & Returned \vs $\policy(\full)$ \\
\midrule
(a) & field only & support (empty support not certifying) & $a\ge1$ \vs $0$ \\
(b) & support only & coverage of the field & $0$ \vs $\ge1$ \\
(c) & pivots instead of the support & certification (two objects justify the action) & $a$ \vs $0$ \\
(d) & a single round & fixed point & $0$ \vs $1$ \\
(e) & additive demands & insertion stability & $0$ \vs $1$ \\
(f) & witness instead of the binding set & certification (bottleneck selector) & $c_\mathrm{A}$ \vs $c_\mathrm{B}$ \\
\bottomrule
\end{tabular}
\caption{\textbf{Counterexamples for omitted or weakened conditions.} For each construction, the procedure that drops or weakens the condition returns an action that differs from $\policy(\full)$; \method returns $\policy(\full)$ in (a)--(d) and (f).}
\label{S-tab:constructions}
\end{table}

\textbf{(a) Field only.}
Let, for some $a\ge1$,
\begin{equation*}
\draft=\{p\}\ \text{with}\ \tau_a\le\dem(p)<\tau_{a+1},\qquad \full=\varnothing,
\end{equation*}
so that $\policy(\draft)=a$ and $\policy(\full)=0$.
The field $\field(\draft)=\{\state:\dem(\state)\ge\tau_{a+1}\}$ does not contain the state of $p$, so that a procedure that verifies only the footprint of the field returns $a\neq\policy(\full)$ whenever the image region of $p$ lies outside this footprint; the empty support is not certifying.
Verifying the support removes $p$.

\textbf{(b) Support only.}
Let
\begin{equation*}
\draft=\varnothing,\qquad \full=\{o\}\ \text{with}\ \dem(o)\ge\tau_1 .
\end{equation*}
Then $\policy(\draft)=0$, and the support is empty; a support-only procedure verifies no region and returns $0\neq\policy(\full)$.
The field $\field(\varnothing)=\{\state:\dem(\state)\ge\tau_1\}$ contains the state of $o$, so that its footprint is verified and $o$ is detected if its box center lies in the footprint.

\textbf{(c) Pivots do not replace the support.}
Let
\begin{equation*}
\draft=\{p_1,p_2\}\ \text{with}\ \tau_a\le\dem(p_j)<\tau_{a+1}\ \ (j=1,2),\qquad \full=\varnothing .
\end{equation*}
Removing either object leaves the action at $a$; there is no pivot, the empty set of pivots is not certifying, and if the image regions of $p_1$ and $p_2$ lie outside the footprint of $\field(\draft)$, a pivot-based procedure returns $a$, whereas $\policy(\full)=0$.
The support contains both objects.

\textbf{(d) A single round is insufficient.}
Let $K=2$ and
\begin{equation*}
\draft=\{p\}\ \text{with}\ \dem(p)\ge\tau_2,\qquad \full=\{o\}\ \text{with}\ \tau_1\le\dem(o)<\tau_2,
\end{equation*}
where $o$ lies outside the crop of $p$ and inside the footprint of $\field(\varnothing)$.
Since $\policy(\draft)=K$, the field of the first round is empty.
The first round removes $p$, the merged scene is empty, and the action becomes $0$.
The state of $o$ is contained only in the field for action $0$, which is computed in the second round. The single-round procedure covers every field that it computes; it fails only because it stops before the field of its merged scene.

\textbf{(e) Joint effects.}
Let the demands add up, $\policy(\scene)=\mathsf{q}\big(\sum_{o\in \scene}\dem(\state(o))\big)$, and let
\begin{equation*}
\draft=\varnothing,\qquad \full=\{o_1,o_2\}\ \text{with}\ \dem(\state(o_1))=\dem(\state(o_2))=0.6\,\tau_1,\quad 1.2\,\tau_1<\tau_2 .
\end{equation*}
The field $\field(\varnothing)=\{\state:\dem(\state)\ge\tau_1\}$ contains neither state, and the empty support of $\varnothing$ is certifying.
If the image regions of $o_1$ and $o_2$ lie outside the footprint, the procedure terminates after one round and returns $0$, whereas $\policy(\full)=1$; only insertion stability is violated.

\textbf{(f) Witnesses are not certifying for bottleneck selectors.}
Let $\mathcal{K}=\{c_\mathrm{A},c_\mathrm{B}\}$ with $c_\mathrm{A}\prec c_\mathrm{B}$ and $\beta\equiv5$.
The draft contains $o_1$ and $o_2$, which affect only $c_\mathrm{B}$; the reference scene contains $o_1$ and an object $o_{\state}$ with state $\state$ that the draft misses:
\begin{equation*}
\mathsf{h}(c_\mathrm{B},o_1)=4,\quad \mathsf{h}(c_\mathrm{B},o_2)=2,\qquad \mathsf{h}(c_\mathrm{A},\state)=3,\quad \mathsf{h}(c_\mathrm{B},\state)=+\infty .
\end{equation*}
The scores $(\mathsf{s}(c_\mathrm{A},\cdot),\mathsf{s}(c_\mathrm{B},\cdot))$ are
\begin{equation*}
\draft:\ (5,2)\ \Rightarrow\ \policy(\draft)=c_\mathrm{A},\qquad
\full:\ (3,4)\ \Rightarrow\ \policy(\full)=c_\mathrm{B},\qquad
\draft\cup\{o_{\state}\}:\ (3,2),
\end{equation*}
so that the state of $\state$ lies outside $\field(\draft)$.
The set $\{o_1\}$ is a witness, since $\mathsf{s}(c_\mathrm{B},\{o_1\})=4\le5$, but the assumption of \cref{S-lem:witness} fails, since inserting $o_{\state}$ lowers the selected score from $5$ to $3$.
Verifying only $o_1$, whose crop contains neither $o_2$ nor $o_{\state}$, leaves the merged scene equal to the draft, and the procedure returns $c_\mathrm{A}$.
The binding set $\{o_1,o_2\}$ removes the spurious $o_2$; the field of $\{o_1\}$ then contains $\state$ (scores $(3,4)$), and the procedure returns $c_\mathrm{B}$.
In driving terms, $o_2$ is a spurious obstacle close to an alternative trajectory, $o_1$ a real obstacle farther from it, and $o_{\state}$ an obstacle on the current trajectory that the draft misses.

\noindent\textbf{Executable checks.}
The released code includes property tests (\nolinkurl{tests/test_properties.py}) that check, on $100{,}200$ random scenes with fixed seeds, the closed-form fields of Corollaries~\ref{cor:maxcomp} and~\ref{cor:bottleneck} against \cref{S-eq:s_field}, insertion stability, the certification of the support over every subset of the other objects with and without one inserted object, and the conclusion of Theorem~\ref{thm:exact} for the loop under locality and coverage; none of the checks fails.

\subsection{Planners outside the class}
\label{S-sec:consumers_outside}

\noindent\textbf{Certification in PDM-Closed.}
PDM-Closed~\cite{pdm} is not a bottleneck selector: its collision and time-to-collision terms are minima over objects, but they are combined multiplicatively and additively with other terms, and its candidate set depends on the objects.
Its field is obtained by probing the planner on the grid described below (a cell is sensitive if the planned $4$\,s trajectory moves by more than $0.5$\,m), and its support by removal queries, \ie, as the set of pivots, which is not guaranteed to be certifying.
On $1{,}000$ navtest scenes, violations of the certification condition~\cref{eq:support} are rare: of $9{,}449$ pairs of objects outside the support, $6$ change the plan only when removed together ($\mathcal{D}$ of size two, $\mathcal{O}=\varnothing$), and of $139{,}675$ combinations of a probe outside the field and the removal of an object outside the support, $11$ do ($|\mathcal{D}|=|\mathcal{O}|=1$); adding the corresponding objects to the support changes the score of none of the $1{,}000$ scenes.

\noindent\textbf{PDM-Closed end to end.}
On all $12{,}146$ navtest scenes, \method runs with this field and support, one crop per 8-connected cluster of sensitive cells, real crop detections, and at most three rounds.
It selects a trajectory within $0.5$\,m of that of $\policy(\full)$ on $97.2\%$ $\ci{96.7}{97.7}$ of the scenes and repairs $77.0\%$ of the $1{,}070$ disagreements of the draft, which agrees on $91.2\%$; it uses $17.8\%$ of the native pixels on average (median $13.6\%$), against $26.7\%$ for the reference.
On its final scenes, $198$ of $214{,}532$ pairs of objects outside the support and $28$ of $521{,}519$ combinations of a probe outside the field with a removal outside the support change the plan.

\noindent\textbf{Empirical check of Corollary~\ref{cor:bottleneck}.}
We instantiate a bottleneck selector on NAVSIM navtest with $15$ candidates, namely the route centerline offset by $0$ and $\pm1.5$\,m combined with constant accelerations of $+2$, $+1$, $0$, $-1.5$, and $-3$\,m/s$^2$; the preference $\beta$ is $5$\,s times the normalized progress, and $\mathsf{h}(c,o)$ is the time of the first overlap of the ego footprint along $c$ with object $o$ under constant velocity, capped at $5$\,s.
The field is probed with a static vehicle on a $7\times7$ ground grid ($8$--$55$\,m ahead, $\pm10.5$\,m laterally), and the support is the binding set of \cref{S-lem:binding}.
With the whole-frame detections inside the crops, \ie, with locality enforced, the procedure agrees with $\policy(\full)$ on $99.71\%$ $\ci{99.58}{99.83}$ of the $12{,}146$ scenes and on $95.2\%$ of the $724$ scenes in which the cheap draft does not.
In each of the $35$ remaining scenes, an object outside the crops changes the selection when it is inserted alone, \ie, the finite grid of static probes misses a sensitive state; no disagreement is due to a joint effect.
On $1{,}000$ scenes, none of the $769{,}561$ pairs of probes whose single insertions leave the selection unchanged changes it jointly, consistent with \cref{S-lem:insertion}; this checks the implementation on the sampled pairs, whereas the guarantee itself follows from the lemma.

\noindent\textbf{The selector with real crops.}
With real crop detections, \method agrees with $\policy(\full)$ on $97.83\%$ $\ci{97.38}{98.22}$ of the scenes and repairs $74.5\%$ of the $724$ disagreements of the draft, with $17.7\%$ of the native pixels on average (median $13.4\%$), two thirds of the $26.7\%$ of the reference.
The draft agrees on $94.04\%$, whole-frame inference at $512$ pixels on $94.08\%$ with $7.1\%$ of the pixels (\cref{tab:planners}), and whole-frame inference at the native resolution on $95.92\%$; \method agrees $3.8$ $\ci{3.2}{4.5}$, $3.8$ $\ci{3.1}{4.5}$, and $1.9$ $\ci{1.5}{2.4}$ points more often (paired bootstrap over the $136$ logs).
The field of this selector spans all $15$ candidates; its footprint is one crop per 8-connected cluster of sensitive cells ($1.0$ per scene on average), and each support crop is the object box enlarged $1.25$ times about its center ($3.1$ per scene); the field crops take $6.1\%$ and the support crops $2.8\%$ of the native pixels, besides the draft and the re-detections of the truncation guard. Both choices keep the field and the certifying support of Corollary~\ref{cor:bottleneck}, since every sampled sensitive probe box and every support box lies inside a requested crop; a scene takes $1.5$ rounds on average, and $0.45\%$ of the scenes reach the round cap.
For the braking controllers, one crop per cluster returns the single crop of \method on every frame of the three datasets, because their sensitive cells, a range interval within the ego corridor, always form one cluster.

\noindent\textbf{Regions of another planner.}
With the field and support of the straight-path braking controller instead of its own, \method agrees on $96.53\%$ $\ci{95.96}{97.00}$ of the scenes with $10.7\%$ of the pixels, $1.3$ points $\ci{1.0}{1.7}$ below its own regions and $0.6$ points $\ci{0.1}{1.1}$ above native whole-frame inference (\cref{tab:planners}).

\section{Implementation and evaluation protocol}
\label{S-sec:supp_impl}

\subsection{Planner and lift}
\label{S-sec:supp_metric}

\noindent\textbf{Braking controller.}
The planner of the experiments is the braking controller of~\cite{deep}.
For an object within the ego corridor (half-width $1.2$\,m) and within $80$\,m, at range $r$ and with ego speed $v$, the demand is
\begin{equation}
\dem = \min\!\Big(\max\Big(\tfrac{v^2}{2\,(r-r_0-v\,t_r)},\ \tfrac{v_{\mathrm{rel}}^2}{2\,(r-r_0)}\Big),\ 9\Big)\ \mathrm{m/s^2},
\label{S-eq:s_demand}
\end{equation}
with stand-off distance $r_0=2$\,m, reaction time $t_r=0.4$\,s, and closing speed $v_{\mathrm{rel}}=r/\mathrm{ttc}$, where the denominators are bounded from below by $0.1$\,m; objects outside the corridor have $\dem=0$.
The action is \textsc{hard-brake} if the largest demand is at least $\tau_2=3.5\,\mathrm{m/s^2}$, \textsc{decelerate} if it is at least $\tau_1=1.0\,\mathrm{m/s^2}$, and \textsc{keep} otherwise.
The controller is max-composable (Definition~\ref{def:maxcomp}) with $K=2$.

\noindent\textbf{Geometric lift.}
Each detection is lifted to an object state in the ego frame with a pinhole camera model, as in~\cite{deep}.
The range combines a ground-plane cue, computed from the bottom edge of the box, and a height cue, computed from the box height and a class-specific height prior; the weight of the ground-plane cue increases with the distance of the contact point from the horizon.
The lateral extent follows from the horizontal box edges at the estimated range.
The draft (u320) uses this lift with a time to collision from the change of the box between two consecutive frames.

\noindent\textbf{Looming estimate.}
For every pipeline other than the draft, the time to collision comes from a learned looming estimate: a ResNet-18~\cite{resnet} with shared weights is applied to the $64\times64$ crops of an object in the last three frames, upsampled to $128\times128$ pixels, and its features are combined with the box features, the ratios of consecutive box heights, and the frame interval to regress the relative closing rate, \ie, the inverse time to collision.
All three observations must stem from detections at the same scale; an object with fewer observations carries no time to collision, and an object is assigned one only if the estimated rate exceeds $10^{-3}$ per frame interval.
The estimate is trained on detections from Argoverse~2 training logs and nuScenes training scenes, with model selection on held-out logs and scenes of both.

\noindent\textbf{Detector settings.}
Of the COCO classes, person, bicycle, motorcycle, car, bus, and truck are kept and mapped to person, cyclist (bicycle, motorcycle), and vehicle (car, bus, truck); a box takes its best class among the six, and all other classes are dropped before non-maximum suppression.
Every detector pass, \ie, the draft, whole frames, crops, and the re-detections of the truncation guard and of the same-scale history, keeps candidates with a class score above $0.10$, applies class-aware non-maximum suppression at an IoU of $0.65$ with at most $100$ boxes, and retains the detections with a confidence of at least $0.25$.
Detections are matched to annotations within compatible classes by Hungarian matching at an IoU of at least $0.3$.

\noindent\textbf{Association across frames.}
The time to collision of the draft associates each box greedily with a draft box of the previous frame at an IoU of at least $0.3$ and equals $\Delta t/(h_t/h_{t-1}-1)$ for a box of height $h$ that grows, and infinity otherwise.
For the looming estimate, each pipeline keeps its own history of detections at one scale in the two previous frames (crop-verified detections for \method and the crop rules, whole-frame detections otherwise) and chains a detection to it by greedy association at an IoU of at least $0.3$; an object without a chain keeps the geometric range and lateral extent and carries no time to collision.
The thresholds are the same for every pipeline.

\noindent\textbf{Range refiner.}
The refiner is a ResNet-18 that receives a $64\times64$ crop, upsampled likewise, and box features and regresses a correction of the log-range.
It is applied once, after the loop has stopped, to the verified support objects of the final scene, and, for whole-frame pipelines including $\full$, to the support of their scene.
It is trained on detections from KITTI~\cite{kitti} train ($14{,}605$), nuScenes training scenes ($62{,}527$), and $340$ Argoverse~2 training logs ($96{,}918$), with model selection on KITTI val and $60$ further Argoverse~2 training logs.
No scene, log, or sequence of a development or test split, and no Argoverse-HD data, is used to train the looming estimate or the refiner.

\subsection{Field, support, and verification loop}
\label{S-sec:supp_loop}

\noindent\textbf{Field.}
The field is evaluated on a ground grid with the ranges $\{12.5, 17.5, 22.5, 27.5, 35, 45, 55, 70\}$\,m and lateral offsets from $-6$ to $6$\,m in steps of $1.5$\,m, with a vehicle-sized probe, $1.8$\,m wide and $1.6$\,m tall, that closes at $\vclose=3$\,m/s, \ie, with the time to collision $r/\vclose$ in \cref{S-eq:s_demand}; for the braking controller on this grid, it selects the same cells as a static probe (\cref{S-tab:grid}).
For the braking controller, the closed form of Corollary~\ref{cor:maxcomp} applies: a cell is sensitive if the demand of the probe reaches the threshold of the next stronger action.
We use a margin and count a cell as sensitive from $60\%$ of that threshold; the current action is always that of the true thresholds.
The margin enlarges the field, and verifying a superset of the field preserves coverage.
The footprint is the rectangle that bounds the image boxes of the sensitive probes, aligned to the stride grid of the detector.

\noindent\textbf{Support.}
The support is that of Corollary~\ref{cor:maxcomp}: every object whose demand reaches the threshold of the current action, and no object at \textsc{keep}.
Each support object contributes a crop of twice its box size around its center.

\noindent\textbf{Grid-aligned crops.}
The reference detector processes the frame resized to a long side of $960$ pixels and letterboxed.
Each crop window is cut from this input on the stride-$32$ grid of the detector, without resizing, so that the detector sees in a crop the pixels that it sees in the whole frame, with image context instead of padding up to the window border.

\noindent\textbf{Truncation guard.}
A crop detection whose box is cut by a crop border inside the image is re-detected in an enlarged crop, which extends the box by its own size on each cut side and takes twice the extended box (at most two enlargements, after which it is merged as it is); it is dropped if no detection of the enlarged crop covers half of it.
The guard skips a truncated box whose demand is zero for every completion of the box, since it cannot change the action.
Guard crops count as pixels.

\noindent\textbf{Same-scale history.}
The looming estimate needs three observations of an object at one scale, which crops provide only if the object is verified in consecutive frames.
Two mechanisms keep this history.
The first round also requests the support crops that were verified in the previous frame.
An object that is newly verified in the ego corridor within $40$\,m and has no same-scale history is, in addition, detected in the same window (twice its box) of the stored reference input of the previous frame; these passes count as pixels. This supplies an observation from the previous frame next to the one from the current frame; the next frame supplies the third, so that the looming estimate starts one frame earlier than without the re-detection; the estimate is never computed from fewer than three observations.

\noindent\textbf{Merging and stopping.}
A draft detection is replaced if its box center lies in a verified crop; of two detections from overlapping crops with an IoU of at least $0.5$, the more confident one is kept, and a draft detection with such an overlap to a kept detection is removed once guard crops were used.
The loop runs at most three rounds.
A requested crop counts as verified once at least $95\%$ of its area lies inside one earlier crop.
\method-fast is \method with a single round.

\begin{table}[!ht]
\centering

\footnotesize
\renewcommand{\arraystretch}{1.15}
\begin{tabular}{@{}p{2.3cm}p{7.4cm}p{1.1cm}p{4.2cm}@{}}
\toprule
Condition & Implementation & Status & Where its effect is measured \\
\midrule
Insertion stability & The braking controllers and the time-to-collision warning are max-composable; the selector on NAVSIM is a bottleneck selector & exact & Corollaries~\ref{cor:maxcomp}, \ref{cor:bottleneck}; joint effects in a general planner: \cref{S-sec:consumers_outside} \\
Certifying support & Closed forms of the two corollaries & exact & pivots in its place: \cref{tab:ablation} \\
Coverage & Field on a finite ground grid with one probe size and one closing speed; the margin enlarges it; the footprint bounds the sensitive probes; the stopping test accepts $95\%$ overlap & approx. & $35$--$42\%$ of the disagreements: \cref{S-tab:conditions}; oracle probes: \cref{S-tab:attribution}; decisive object: \cref{S-tab:coverage}; other probes: \cref{S-tab:grid} \\
Locality & Crop detections differ from whole-frame ones through context, truncation, and temporal cues; grid-aligned crops, the truncation guard, and the same-scale history narrow the gap & approx. & $34$--$47\%$ of the disagreements: \cref{S-tab:conditions}; oracle detections and states: \cref{S-tab:attribution}; without the guard: \cref{tab:ablation} \\
Fixed point & At most three rounds & bounded & cap reached on no test frame: \cref{S-sec:supp_attribution}; single round: \cref{tab:ablation} \\
\midrule
(none) & Range refiner, applied after the loop & outside & without it: \cref{tab:ablation} \\
\bottomrule
\end{tabular}
\caption{\textbf{Conditions of Theorem~\ref{thm:exact} in the implementation.} ``Exact'': the condition holds by construction; ``approx.'': the implementation approximates it; ``bounded'': the iteration is capped.}
\label{S-tab:practice}
\end{table}

\Cref{S-tab:practice} relates each condition of Theorem~\ref{thm:exact} to its implementation.
Conditions (i) and (ii) hold exactly for the braking controllers, the time-to-collision warning, and the bottleneck selector (\cref{S-sec:consumers_outside} tests the pivot support of PDM-Closed separately), so every disagreement between \method and $\policy(\full)$ on a frame whose loop stops within the cap stems from locality, from coverage, or from the range refiner, which acts after the loop and which the theorem does not cover.

\subsection{Datasets and metrics}
\label{S-sec:supp_data}

\begin{table}[!ht]
\centering

\footnotesize
\setlength{\tabcolsep}{5pt}
\renewcommand{\arraystretch}{1.1}
\begin{tabular}{@{}llllll@{}}
\toprule
Dataset & Front camera & Development & Test & Frames & Episodes \\
\midrule
nuScenes~\cite{nuscenes} & $1600\times900$, keyframes at $2$\,Hz & $152$ scenes & $228$ scenes & $9{,}181$ & $147$ \\
Argoverse~2~\cite{av2} & $1550\times2048$, annotated frames & $300$ training logs & val, $150$ logs & $11{,}818$ & $190$ \\
Argoverse-HD~\cite{streaming} & $1920\times1200$, $30$\,Hz & \na & val, $24$ sequences & $15{,}006$ & $71$ \\
\ \ held-out & & \na & $65$ training sequences & $39{,}265$ & $154$ \\
NAVSIM~\cite{navsim} & $1920\times1080$ & navmini & navtest & $12{,}146$ scenes & \na \\
\bottomrule
\end{tabular}
\caption{\textbf{Datasets and splits.} Episodes: braking episodes required by the annotations. The nuScenes development and test splits are disjoint sets of scenes of its trainval set (the latter is not the official test set) and are disjoint from the scenes used to train the learned parts of the lift.}
\label{S-tab:data}
\end{table}

\noindent\textbf{Datasets.}
\Cref{S-tab:data} lists the splits.
The $300$ Argoverse~2 development logs are training logs that are disjoint from the logs used to train the refiner and the looming estimate.
Argoverse-HD has no development split: it is evaluated with the configuration shared by all datasets.
The front camera of Argoverse~2 images the hood of the ego vehicle, which the COCO-pretrained detector reports as a vehicle in the ego corridor at low resolution; a fixed image band (rows $\ge1830$ of $2048$) is therefore masked for every pipeline by discarding the detections whose box lies in the band by at least one half.
On the two Argoverse datasets, the geometric lift uses the camera height above the road ($1.75$ and $1.70$\,m).

\noindent\textbf{Selection of the configuration.}
The braking configuration of \method for the three camera datasets is selected only on the $152$ nuScenes development scenes and the $300$ Argoverse~2 development logs of \cref{S-tab:data}, which are disjoint from the units used to train the learned parts of the lift and from the test splits.
A component or parameter is adopted if, tested against the incumbent configuration, the confidence interval of its effect on episode F1 is above zero on one development set and its point estimate is non-negative on the other; shared values are preferred to dataset-specific ones.
The baselines are likewise selected by episode F1 on the same development sets (\cref{S-sec:supp_baselines}).
The selected configuration, its code, the weights of the looming estimate and the range refiner, and the decision criterion are frozen before evaluation on the three test splits.
Argoverse-HD is used neither for configuration selection nor for tuning: it receives the same configuration.
\Cref{S-sec:supp_loop} specifies the resulting field, support, crops, guard, history, and stopping rule; the baseline settings are selected separately as described in \cref{S-sec:supp_baselines}.

\noindent\textbf{Evaluation protocol.}
Unless stated otherwise, results are reported on the test splits of \cref{S-tab:data} with the frozen configuration.
The test-split analyses use the final configuration; no configuration is changed on the basis of these analyses.

\noindent\textbf{Fidelity.}
The agreement is the share of frames on which a pipeline takes the action $\policy(\full)$, and the repair rate is the agreement on the frames on which the draft takes another action.
We also report the agreement on the frames on which $\policy(\full)$ changes from the preceding frame (Trans.).
All three are computed from the actions alone, and each pipeline keeps its own temporal history.

\noindent\textbf{Decision quality.}
The \emph{required} action of a frame is the action of the controller on the annotated objects with their annotated states.
These states come from the 3D annotations of each dataset, not from image boxes or from the lift: the footprint of an annotated cuboid in the ego frame gives its range and lateral extent, and the change of its range along the annotated track gives its closing rate and time to collision.
Argoverse-HD itself provides image boxes only; its states are taken from the 3D track labels of the underlying Argoverse tracking logs ($10$\,Hz), assigned to a camera frame from the LiDAR sweep closest in time (within $50$\,ms) and motion-compensated to the camera timestamp; the $56$ of its $15{,}062$ test frames without such a sweep are processed but not scored.
A required braking episode is a maximal run of frames whose required action is at least \textsc{decelerate}, with gaps of one frame merged; it is missed (FN) if the pipeline brakes in none of its frames, and it is a true positive (TP) otherwise.
A false episode (FP) is a maximal run of braking frames of the pipeline, with gaps of one frame merged, that contains no frame of a required episode.
Precision is TP$/($TP$+$FP$)$, recall is TP$/($TP$+$FN$)$, and F1 is $2$TP$/(2$TP$+$FP$+$FN$)$.
The \emph{decisive object} of a frame that requires braking is the annotated object with the largest demand, and an annotated object is \emph{decision-relevant} if its own demand reaches $\tau_1$.

\noindent\textbf{Pixels.}
The detector-input pixels of a frame are measured on the input tensors of the detector, padded to multiples of $32$ pixels: the draft pass, every crop, and every re-detection of the truncation guard and of the same-scale history; native-resolution inference therefore exceeds $100\%$.
For a baseline, they include the cheap pass if the method uses it, the mosaics of UFP, and the canvases of FOVEA.
The $64\times64$ inputs of the looming estimate and of the refiner are not counted as detector-input pixels in any pipeline.

\noindent\textbf{Statistical comparisons.}
We use $2{,}000$ paired bootstrap resamples of scenes, logs, or sequences, retaining the same sampled units for the pipelines being compared, and report $95\%$ percentile intervals.
The Mean columns of \cref{tab:main} weight the three datasets equally, and their intervals average the bootstrap replicates of the three datasets, each resampled over its own units.
A difference is significant when its interval excludes zero; $^{\dagger}$ and $^{\ast}$ indicate that the value of \method is significantly higher and lower, respectively, whatever the direction of improvement of the metric; only the closed-loop tables mark it as significantly better and worse.
For the Holm correction, the two-sided $p$-value of a difference is twice the share of its bootstrap replicates on the far side of zero, and at least $1/2{,}000$.
Unless stated otherwise, comparisons are not adjusted for multiplicity.
The resampling for the controlled closed loop is specified separately in \cref{S-sec:supp_embedded}.
\Cref{S-tab:contrasts} gives the intervals of the primary comparisons with u512 in agreement and repair rate; all remain significant after a Holm correction.

\begin{table}[!ht]
\centering

\footnotesize
\setlength{\tabcolsep}{8pt}
\begin{tabular}{@{}lccc@{}}
\toprule
 & nuScenes & Argoverse 2 & Argoverse-HD \\
\midrule
\method $-$ u512, Agr. & $+1.1$ \ci{$0.7$}{$1.6$}$^{\circ}$ & $+1.8$ \ci{$1.4$}{$2.3$}$^{\circ}$ & $+3.0$ \ci{$1.7$}{$4.9$}$^{\circ}$ \\
\method $-$ u512, Rep. & $+7.3$ \ci{$2.4$}{$11.8$}$^{\circ}$ & $+7.7$ \ci{$5.6$}{$10.0$}$^{\circ}$ & $+9.4$ \ci{$4.7$}{$15.3$}$^{\circ}$ \\
\bottomrule
\end{tabular}
\caption{\textbf{Primary comparisons with intervals} (points; test splits). Difference with its $95\%$ paired bootstrap interval; $^{\circ}$: significant at $0.05$ after a Holm correction.}
\label{S-tab:contrasts}
\end{table}

\subsection{Baselines}
\label{S-sec:supp_baselines}

\begin{table}[!ht]
\centering

\footnotesize
\renewcommand{\arraystretch}{1.1}
\begin{tabular}{@{}>{\raggedright\arraybackslash}p{3.6cm}>{\raggedright\arraybackslash}p{6.2cm}>{\raggedright\arraybackslash}p{5.6cm}@{}}
\toprule
Baseline & Implementation & Setting selected on development data \\
\midrule
AdaScale~\cite{adascale} & Re-implementation (original: Caffe, R-FCN); ResNet-18 scale classifier over the whole-frame resolutions & Predicted scale raised by one step, resolutions up to u1280 \\
SAHI~\cite{sahi} & Official library & Slices of $640$ pixels, overlap $0.1$ \\
FOVEA~\cite{fovea} & Official grid generator, saliency from the draft detections & Canvas of u1280 \\
UFP~\cite{ufpmp} & Official foreground packing module, YOLOv8s on the mosaics & Merge scale $2.0$, mosaics of $512$ pixels \\
Budget-Adaptive Routing (BAR)~\cite{bgtada} & Official estimators, offloader, and arbiter & u1280 as the strong mode, half of the frames routed \\
Chanakya~\cite{chanakya} & Objective of~\cite{chanakya} implemented directly, since the official controller adapts detector settings that do not map to our actions & Cost-sensitive learner; pixel budget matched to \method \\
AnA~\cite{ana} & Re-detects the draft detections in the ego corridor within $80$\,m & Iterated up to three rounds \\
Counterfactual importance (CF)~\cite{cfimportance} & Re-detects the draft objects whose removal or perturbation changes the action & Perturbations of $50\%$ in range and $2$\,s in time to collision \\
\bottomrule
\end{tabular}
\caption{\textbf{Baselines and their implementations.} ``Official'': the released code; ``re-implementation'': our implementation of the published method; AnA, Chanakya: the selection rule of the cited work within our pipeline and with our planner, not a reproduction of the original system.}
\label{S-tab:baselines}
\end{table}

\noindent\textbf{Shared components.}
Every baseline runs the draft pass shared by all pipelines (YOLOv8s at a long side of $320$ pixels, confidence threshold $0.25$) if it uses one, and YOLOv8s at the scale or on the regions that the method selects.
In \method, a draft object that is not verified keeps the geometric lift of the draft (u320); the looming estimate and the refiner act on verified objects only.
Every baseline uses the lift of the reference: the geometric lift, the looming estimate on its own same-scale history, and the range refiner on the support of the braking controller in its own scene.
Methods that crop use the grid-aligned crops, the truncation guard, and the merge rule of \method.
All pipelines thus share the detector, the draft, the lift, and the planner, so that the comparison is controlled: they differ from \method and from one another in where the detector runs and in what follows from it, such as which objects have the same-scale history of the looming estimate.

\noindent\textbf{Region rules of the object-level baselines.}
AnA and counterfactual importance select regions from the detections of the draft with their own rule and use no region of \method.
The former re-detects every draft detection in the ego corridor within $80$\,m, as in~\cite{ana}.
The latter re-detects the draft objects whose removal, or whose perturbation in range or time to collision, changes the action, as in~\cite{cfimportance}.
Neither searches where the draft detects nothing.

\noindent\textbf{Selection.}
For each baseline, a predefined grid of settings, which includes its official or previously reported form, is scored on the development data, and the setting with the highest mean F1 over nuScenes and Argoverse~2 is evaluated once on the test splits; \cref{S-tab:baselines} lists the selected settings.
The two frame-level routers (Budget-Adaptive Routing and Chanakya) are computed from whole-frame rows, with settings selected on the same development data.
Argoverse-HD uses the baseline settings selected on Argoverse~2, without tuning on its sequences.
No learned component of any baseline uses a test unit.
On Argoverse~2, the scale classifier of AdaScale and the router estimators were fit on training logs that include $242$ of the $300$ development logs, which can make their selection optimistic.

\subsection{Reproducibility}
\label{S-sec:supp_repro}

The planner, the lift, and their constants are given in \cref{S-sec:supp_metric}; the probe grid, the crops, the guard, the history, and the loop in \cref{S-sec:supp_loop}; the splits, the metrics, and the pixel accounting in \cref{S-sec:supp_data}; and the baselines and their selected settings in \cref{S-sec:supp_baselines}.
All intervals are paired bootstrap intervals over the units (scenes, logs, sequences, or routes) with $2{,}000$ resamples, a fixed seed, and $95\%$ percentiles.
The qualitative frames follow the rule of \cref{S-sec:supp_qualitative}.

\section{Fidelity and decision quality}
\label{S-sec:supp_results}

\subsection{All pipelines}
\label{S-sec:supp_all}

\begin{table}[!ht]
\centering

\scriptsize
\setlength{\tabcolsep}{2.6pt}
\begin{tabular}{@{}lcccccccccccc@{}}
\toprule
 & \multicolumn{4}{c}{nuScenes} & \multicolumn{4}{c}{Argoverse 2} & \multicolumn{4}{c}{Argoverse-HD} \\
\cmidrule(lr){2-5}\cmidrule(lr){6-9}\cmidrule(l){10-13}
Pipeline & Agr. & Rep. & Trans. & px & Agr. & Rep. & Trans. & px & Agr. & Rep. & Trans. & px \\
\midrule
\rf{u320 (draft)} & \rf{92.7\sgu{}} & \rf{0\sgu{}} & \rf{65.3\sgu{}} & \rf{4.3} & \rf{85.3\sgu{}} & \rf{0\sgu{}} & \rf{57.6\sgu{}} & \rf{2.6} & \rf{78.8\sgu{}} & \rf{0\sgu{}} & \rf{45.5\sgu{}} & \rf{3.1} \\
\rf{u320$^{+}$} & \rf{96.5\sgu{}} & \rf{59.9\sgu{}} & \rf{71.8\sgu{}} & \rf{4.3} & \rf{94.7\sgu{}} & \rf{70.2\sgu{}} & \rf{66.0\sgu{}} & \rf{2.6} & \rf{91.4\sgu{}} & \rf{63.8\sgu{}} & \rf{54.0\sgu{}} & \rf{3.1} \\
\rf{u448} & \rf{97.4\sgu{}} & \rf{76.7\sgu{}} & \rf{76.4\sgu{}} & \rf{8.0} & \rf{96.0\sgu{}} & \rf{81.8\sgu{}} & \rf{72.2\sgu{}} & \rf{5.0} & \rf{93.8\sgu{}} & \rf{79.5\sgu{}} & \rf{57.5\sgu{}} & \rf{5.6} \\
\rf{u512} & \rf{97.7\sgu{}} & \rf{79.6\sgu{}} & \rf{76.7\sgu{}} & \rf{10.2} & \rf{96.4\sgu{}} & \rf{84.0\sgu{}} & \rf{74.9\sgu{}} & \rf{6.7} & \rf{94.5\sgu{}} & \rf{82.6\sgu{}} & \rf{54.0\sgu{}} & \rf{7.1} \\
\rf{u640} & \rf{98.2\sgu{}} & \rf{84.7} & \rf{81.6} & \rf{17.1} & \rf{97.3\sgu{}} & \rf{88.3\sgu{}} & \rf{77.0\sgu{}} & \rf{10.3} & \rf{95.8\sgu{}} & \rf{87.4\sgu{}} & \rf{58.4\sgu{}} & \rf{11.6} \\
\rf{u768} & \rf{98.6} & \rf{88.6} & \rf{84.7} & \rf{23.9} & \rf{97.8\sgu{}} & \rf{90.4} & \rf{80.9} & \rf{14.7} & \rf{96.1\sgu{}} & \rf{88.5\sgu{}} & \rf{56.8\sgu{}} & \rf{16.0} \\
\rf{u960 ($\full$)} & \rf{100} & \rf{100} & \rf{100} & \rf{38.4} & \rf{100} & \rf{100} & \rf{100} & \rf{23.2} & \rf{100} & \rf{100} & \rf{100} & \rf{25.3} \\
\rf{u1280} & \rf{98.3\sgu{}} & \rf{87.8} & \rf{84.2} & \rf{65.4} & \rf{97.6\sgu{}} & \rf{91.8} & \rf{82.1} & \rf{40.0} & \rf{96.2\sgu{}} & \rf{90.0} & \rf{60.5\sgu{}} & \rf{44.4} \\
\rf{native} & \rf{97.7\sgu{}} & \rf{81.4\sgu{}} & \rf{78.2\sgu{}} & \rf{103.1} & \rf{96.0\sgu{}} & \rf{87.9\sgu{}} & \rf{77.0\sgu{}} & \rf{101.2} & \rf{93.7\sgu{}} & \rf{85.4\sgu{}} & \rf{55.0\sgu{}} & \rf{101.3} \\
\midrule
Chanakya~\cite{chanakya} & 96.3\sgu{} & 54.2\sgu{} & 77.1\sgu{} & 7.9 & 92.7\sgu{} & 53.0\sgu{} & 69.4\sgu{} & 4.9 & 91.5\sgu{} & 63.1\sgu{} & 55.7\sgu{} & 6.3 \\
BAR~\cite{bgtada} & 96.1\sgu{} & 52.2\sgu{} & 77.3\sgu{} & 36.6 & 91.3\sgu{} & 44.6\sgu{} & 68.1\sgu{} & 21.4 & 88.8\sgu{} & 50.2\sgu{} & 53.8\sgu{} & 23.1 \\
SAHI~\cite{sahi} & 97.4\sgu{} & 85.9 & 81.8 & 187.7 & 93.6\sgu{} & 83.5\sgu{} & 76.1\sgu{} & 165.2 & 91.4\sgu{} & 82.7\sgu{} & 54.7\sgu{} & 153.8 \\
CF~\cite{cfimportance} & 97.5\sgu{} & 71.3\sgu{} & 76.7\sgu{} & 5.3 & 96.1\sgu{} & 78.7\sgu{} & 75.2\sgu{} & 3.8 & 91.5\sgu{} & 64.4\sgu{} & 60.5\sgu{} & 4.5 \\
FOVEA~\cite{fovea} & 97.9\sgu{} & 85.0 & 80.0\sgu{} & 69.7 & 96.5\sgu{} & 88.0\sgu{} & 78.7\sgu{} & 42.6 & 95.6\sgu{} & 88.1\sgu{} & 58.4\sgu{} & 47.6 \\
AdaScale~\cite{adascale} & 98.4 & 86.9 & 84.0 & 30.6 & 99.1\sgd{} & 96.8\sgd{} & 93.4\sgd{} & 26.8 & 98.2 & 95.2 & 79.9\sgd{} & 29.9 \\
AnA~\cite{ana} & 98.1\sgu{} & 78.5\sgu{} & 81.1 & 8.3 & 97.4\sgu{} & 85.6\sgu{} & 80.3 & 5.8 & 94.7\sgu{} & 77.5\sgu{} & 71.4 & 11.7 \\
UFP~\cite{ufpmp} & 98.0\sgu{} & 83.2 & 80.0 & 18.1 & 97.2\sgu{} & 87.6\sgu{} & 77.6\sgu{} & 9.0 & 95.3\sgu{} & 84.5\sgu{} & 59.6\sgu{} & 12.8 \\
\midrule
\method-fast & 98.7 & 86.8 & 84.4 & 8.1 & 98.2 & 91.5 & 83.4 & 5.6 & 97.4\sgu{} & 91.6 & 67.7 & 6.5 \\
\rowcolor{oursbg}\textbf{\method} & 98.7 & 86.9 & 84.4 & 8.4 & 98.2 & 91.7 & 83.4 & 5.9 & 97.5 & 92.0 & 68.1 & 6.9 \\
\bottomrule
\end{tabular}
\caption{\textbf{Decision fidelity of all pipelines} (test splits; complements \cref{tab:fid}). Agr.: frames with the action $\policy(\full)$ (\%); Rep.: the same on the frames where the draft differs from $\policy(\full)$; Trans.: the same on the frames where $\policy(\full)$ changes from the preceding frame; px: detector-input pixels (\% of native). Top block: whole-frame inference. \sgu{} / \sgd{}: \method significantly higher / lower.}
\label{S-tab:fid_full}
\end{table}

\noindent\textbf{Fidelity.}
\Cref{S-tab:fid_full} extends \cref{tab:fid} to all pipelines and adds the agreement at changes of the reference action.
Only AdaScale, with $3.6$--$4.5\times$ the pixels of \method, agrees with $\policy(\full)$ more often, on Argoverse~2 (significantly) and Argoverse-HD; no other baseline does so on any dataset.
u448, which uses fewer pixels than \method on every dataset ($5.0$--$8.0\%$), agrees $1.3$--$3.7$ points less often and repairs $9.9$--$12.5$ points fewer disagreements, all significantly.
The frame-level routers repair about half of the disagreements of the draft, the object-level methods, which verify only what the draft detects, $64$--$86\%$, and \method $87$--$92\%$.
The frames on which $\policy(\full)$ changes are the hardest for every pipeline; there, \method agrees on $68$--$84\%$, above every whole-frame pipeline other than the reference on the two Argoverse datasets (significantly except u768 and u1280 on Argoverse~2) and level with u768 and u1280 on nuScenes, and only AdaScale is significantly higher (on Argoverse~2 and Argoverse-HD).

\begin{table}[!ht]
\centering

\footnotesize
\setlength{\tabcolsep}{4pt}
\begin{tabular}{@{}lccccccccc@{}}
\toprule
 & \multicolumn{3}{c}{nuScenes} & \multicolumn{3}{c}{Argoverse 2} & \multicolumn{3}{c}{Argoverse-HD} \\
\cmidrule(lr){2-4}\cmidrule(lr){5-7}\cmidrule(l){8-10}
Method & Missed & False & False\,/\,1k & Missed & False & False\,/\,1k & Missed & False & False\,/\,1k \\
\midrule
\rf{u320 (draft)} & \rf{51} & \rf{210} & \rf{22.9\sgd{}} & \rf{34} & \rf{376} & \rf{31.8\sgd{}} & \rf{21\sgu{}} & \rf{408} & \rf{27.2\sgd{}} \\
\rf{u320$^{+}$} & \rf{63\sgd{}} & \rf{75} & \rf{8.2\sgd{}} & \rf{44} & \rf{86} & \rf{7.3\sgd{}} & \rf{35} & \rf{105} & \rf{7.0\sgd{}} \\
\rf{u448} & \rf{57\sgd{}} & \rf{73} & \rf{8.0\sgd{}} & \rf{34} & \rf{90} & \rf{7.6\sgd{}} & \rf{34} & \rf{73} & \rf{4.9\sgd{}} \\
\rf{u512} & \rf{50} & \rf{68} & \rf{7.4\sgd{}} & \rf{37} & \rf{86} & \rf{7.3\sgd{}} & \rf{37} & \rf{66} & \rf{4.4\sgd{}} \\
\rf{u640} & \rf{50} & \rf{64} & \rf{7.0\sgd{}} & \rf{35\sgu{}} & \rf{61} & \rf{5.2\sgd{}} & \rf{32} & \rf{64} & \rf{4.3\sgd{}} \\
\rf{u768} & \rf{49} & \rf{59} & \rf{6.4\sgd{}} & \rf{33\sgu{}} & \rf{62} & \rf{5.2\sgd{}} & \rf{37} & \rf{68} & \rf{4.5\sgd{}} \\
\rf{u960 ($\full$)} & \rf{45} & \rf{65} & \rf{7.1\sgd{}} & \rf{34\sgu{}} & \rf{65} & \rf{5.5\sgd{}} & \rf{34} & \rf{73} & \rf{4.9\sgd{}} \\
\rf{u1280} & \rf{47} & \rf{60} & \rf{6.5\sgd{}} & \rf{34\sgu{}} & \rf{70} & \rf{5.9\sgd{}} & \rf{35} & \rf{73} & \rf{4.9\sgd{}} \\
\rf{native} & \rf{52} & \rf{47} & \rf{5.1} & \rf{33\sgu{}} & \rf{71} & \rf{6.0\sgd{}} & \rf{32} & \rf{68} & \rf{4.5\sgd{}} \\
\midrule
Chanakya~\cite{chanakya} & 58\sgd{} & 124 & 13.5\sgd{} & 37 & 270 & 22.8\sgd{} & 32 & 201 & 13.4\sgd{} \\
BAR~\cite{bgtada} & 53 & 122 & 13.3\sgd{} & 33\sgu{} & 260 & 22.0\sgd{} & 31 & 223 & 14.9\sgd{} \\
SAHI~\cite{sahi} & 47 & 114 & 12.4\sgd{} & 30\sgu{} & 143 & 12.1\sgd{} & 26\sgu{} & 149 & 9.9\sgd{} \\
CF~\cite{cfimportance} & 66\sgd{} & 48 & 5.2 & 47 & 61 & 5.2\sgd{} & 40 & 137 & 9.1\sgd{} \\
FOVEA~\cite{fovea} & 49 & 64 & 7.0\sgd{} & 35 & 88 & 7.4\sgd{} & 36 & 77 & 5.1\sgd{} \\
AdaScale~\cite{adascale} & 50 & 67 & 7.3\sgd{} & 37 & 67 & 5.7\sgd{} & 34 & 76 & 5.1\sgd{} \\
AnA~\cite{ana} & 61\sgd{} & 41 & 4.5 & 43 & 47 & 4.0 & 39 & 62 & 4.1 \\
UFP~\cite{ufpmp} & 55 & 55 & 6.0\sgd{} & 35 & 60 & 5.1\sgd{} & 33 & 63 & 4.2\sgd{} \\
\midrule
\method-fast & 49 & 46 & 5.0\sgd{} & 41 & 48 & 4.1 & 35 & 56 & 3.7\sgd{} \\
\rowcolor{oursbg}\textbf{\method} & 48 & 41 & 4.5 & 42 & 44 & 3.7 & 36 & 44 & 2.9 \\
\bottomrule
\end{tabular}
\caption{\textbf{Episode counts behind \cref{tab:main}} (test splits). Missed: required braking episodes in which the pipeline never brakes, of $147$, $190$, and $71$; False: braking episodes without a requirement; False\,/\,1k: the same per $1{,}000$ frames. Lower is better. Top block: whole-frame inference. \sgu{} / \sgd{}: the value of \method is significantly higher / lower.}
\label{S-tab:episodes_full}
\end{table}

\noindent\textbf{Episode counts.}
\Cref{S-tab:episodes_full} gives the counts behind the precision and recall of \cref{tab:main}.
\method has the fewest false braking episodes on every dataset ($41$--$44$, level with AnA on nuScenes), against $65$--$73$ for the reference and $66$--$86$ for u512.

\subsection{Fidelity by regime}
\label{S-sec:supp_strata}

\begin{table}[!ht]
\centering

\footnotesize
\setlength{\tabcolsep}{4pt}
\begin{tabular}{@{}llcccccccc@{}}
\toprule
 & & \multicolumn{3}{c}{Ego speed (tercile)} & \multicolumn{5}{c}{Range of the decisive object (m)} \\
\cmidrule(lr){3-5}\cmidrule(l){6-10}
 & Pipeline & Low & Mid & High & $<$20 & 20--40 & 40--60 & $>$60 & none \\
\midrule
\multirow{5}{*}{\rotatebox{90}{nuSc.}} & u512 & 99.3\sgd{} & 97.6\sgu{} & 95.8\sgu{} & 92.5 & 88.1\sgu{} & 82.7\sgu{} & 58.8 & 98.6\sgu{} \\
 & u640 & 99.5\sgd{} & 98.0\sgu{} & 97.0\sgu{} & 93.2 & 92.2 & 89.0\sgu{} & 70.6 & 98.9 \\
 & \textbf{\method} & 98.8 & 98.8 & 98.7 & 94.8 & 95.1 & 95.3 & 82.4 & 99.2 \\
 & \emph{frames} & 3{,}330 & 2{,}992 & 2{,}859 & 439 & 345 & 127 & 17 & 8{,}253 \\
\midrule
\multirow{5}{*}{\rotatebox{90}{AV2}} & u512 & 98.3 & 95.0\sgu{} & 95.9\sgu{} & 92.8\sgu{} & 92.5\sgu{} & 88.8\sgu{} & 80.5\sgu{} & 98.1\sgu{} \\
 & u640 & 98.7 & 96.2\sgu{} & 96.8\sgu{} & 95.0 & 93.4\sgu{} & 92.6\sgu{} & 84.6\sgu{} & 98.5 \\
 & \textbf{\method} & 98.2 & 97.8 & 98.8 & 96.5 & 96.8 & 96.7 & 92.3 & 98.8 \\
 & \emph{frames} & 4{,}102 & 3{,}902 & 3{,}814 & 1{,}108 & 1{,}066 & 511 & 169 & 8{,}964 \\
\midrule
\multirow{5}{*}{\rotatebox{90}{A-HD}} & u512 & 96.4\sgu{} & 96.4\sgu{} & 90.2\sgu{} & 95.5\sgu{} & 85.5\sgu{} & 87.4\sgu{} & 81.0\sgu{} & 96.7\sgu{} \\
 & u640 & 96.9 & 96.9\sgu{} & 93.2\sgu{} & 95.3\sgu{} & 89.6\sgu{} & 92.6\sgu{} & 83.6\sgu{} & 97.3\sgu{} \\
 & \textbf{\method} & 97.2 & 98.6 & 96.5 & 99.2 & 95.2 & 96.7 & 93.4 & 97.9 \\
 & \emph{frames} & 4{,}486 & 5{,}833 & 4{,}687 & 716 & 1{,}789 & 783 & 305 & 11{,}413 \\
\bottomrule
\end{tabular}
\caption{\textbf{Agreement with $\policy(\full)$ by ego speed and by the range of the decisive object} (\%; test splits; complements \cref{tab:regime}). Terciles of the ego speed with cut points from the development data; none: frames without a decisive object. \sgu{} / \sgd{}: \method significantly higher / lower.}
\label{S-tab:strata}
\end{table}

\Cref{S-tab:strata} gives the ego-speed and range stratifications of \cref{tab:regime} in full, with their frame counts.
The Seen and Missed columns of \cref{tab:regime} contain $787$ and $141$, $2{,}574$ and $280$, and $2{,}531$ and $1{,}062$ frames on nuScenes, Argoverse~2, and Argoverse-HD.
Together, these are the frames with a decisive object, which are the frames that require braking ($10$--$24\%$ of all frames, as in \cref{S-tab:coverage}); on them, \method agrees with $\policy(\full)$ on $94.7$--$96.4\%$, against $87.5$--$91.2\%$ for u512 and $90.9$--$93.4\%$ for u640, whereas on the frames without a decisive object all three agree on $96.7$--$99.2\%$.
The overall agreement of \cref{tab:fid} is thus dominated by frames that require no braking, and the lead of \method over u512 is three to five times larger on the frames that require braking ($5.2$--$8.6$ points against $1.1$--$3.0$).
The strata are subsets of the evaluated test splits, not independent evaluations on new data; the ego-speed cut points are fixed from development data.
By ego speed, \method and u512 lie within one point of each other in the lowest tercile, where u512 is ahead on nuScenes, and \method leads by $2.9$ to $6.3$ points in the highest tercile.
By the range of the decisive object, the lead over u512 grows with the range, from $2$--$4$ points below $20$\,m to $8$--$13$ points from $40$ to $60$\,m and $12$--$24$ points beyond $60$\,m: the farther the object that decides the action, the less a uniformly downscaled frame resolves it.

\begin{table}[!ht]
\centering

\footnotesize
\setlength{\tabcolsep}{8pt}
\begin{tabular}{@{}llccc@{}}
\toprule
 & Pipeline & $\times1.5$ & $\times2$ & $\times2$, $\infty$ \\
\midrule
\multirow{3}{*}{\rotatebox{90}{nuSc.}} & u512 & 95.6\sgu{} & 94.5\sgu{} & 92.5\sgu{} \\
 & u640 & \underline{96.7}\sgu{} & \underline{95.7}\sgu{} & \underline{94.2}\sgu{} \\
 & \textbf{\method} & \textbf{98.2} & \textbf{97.9} & \textbf{97.3} \\
\midrule
\multirow{3}{*}{\rotatebox{90}{AV2}} & u512 & 96.2\sgu{} & 96.1\sgu{} & 91.8\sgu{} \\
 & u640 & \underline{97.1}\sgu{} & \underline{96.9}\sgu{} & \underline{93.9}\sgu{} \\
 & \textbf{\method} & \textbf{98.4} & \textbf{98.6} & \textbf{97.5} \\
\midrule
\multirow{3}{*}{\rotatebox{90}{A-HD}} & u512 & 92.3\sgu{} & 91.1\sgu{} & 85.8\sgu{} \\
 & u640 & \underline{95.0}\sgu{} & \underline{93.9}\sgu{} & \underline{89.9}\sgu{} \\
 & \textbf{\method} & \textbf{97.5} & \textbf{96.8} & \textbf{96.0} \\
\bottomrule
\end{tabular}
\caption{\textbf{Agreement with $\policy(\full)$ under a scaled ego speed} (\%; test splits). The ego speed given to the planner is multiplied by $s$ ($\times s$) and the $80$\,m range limit of the controller is removed ($\infty$); the images are unchanged, and so are the detections and closing rates of whole-frame inference, so that this is a counterfactual stress test of the range that the planner must cover, not a measurement at higher driving speeds. \textbf{Best}, \underline{second}; \sgu{} / \sgd{}: \method significantly higher / lower.}
\label{S-tab:stress}
\end{table}

\noindent\textbf{Scaled ego speed.}
\Cref{S-tab:stress} scales the ego speed that the planner receives, which lengthens the range it must cover.
The lead of \method over u512 grows from $1$--$3$ points at the recorded speed to $3$--$6$ points at twice the speed and to $5$--$10$ points without the range limit, all significant; the images are unchanged, so the result concerns the planner's range, not driving at higher speeds.

\begin{table}[!ht]
\centering

\scriptsize
\setlength{\tabcolsep}{3pt}
\begin{tabular}{@{}lcccccccccccc@{}}
\toprule
 & \multicolumn{4}{c}{nuScenes} & \multicolumn{4}{c}{Argoverse 2} & \multicolumn{4}{c}{Argoverse-HD} \\
\cmidrule(lr){2-5}\cmidrule(lr){6-9}\cmidrule(l){10-13}
Condition & Agr. & u512 & Rep. & u512 & Agr. & u512 & Rep. & u512 & Agr. & u512 & Rep. & u512 \\
\midrule
All frames & 98.7 & \rf{97.7\sgu{}} & 86.9 & \rf{79.6\sgu{}} & 98.2 & \rf{96.4\sgu{}} & 91.7 & \rf{84.0\sgu{}} & 97.5 & \rf{94.5\sgu{}} & 92.0 & \rf{82.6\sgu{}} \\
\midrule
Day & 98.9 & \rf{98.0\sgu{}} & 87.3 & \rf{81.7\sgu{}} & \na & \na & \na & \na & \na & \na & \na & \na \\
Night & 97.8 & \rf{95.4\sgu{}} & 84.8 & \rf{68.6\sgu{}} & \na & \na & \na & \na & \na & \na & \na & \na \\
Dry & 99.0 & \rf{98.0\sgu{}} & 88.8 & \rf{81.4\sgu{}} & \na & \na & \na & \na & \na & \na & \na & \na \\
Rain & 97.9 & \rf{96.4\sgu{}} & 81.5 & \rf{74.6} & \na & \na & \na & \na & \na & \na & \na & \na \\
\midrule
Decisive box: small & 95.5 & \rf{84.8\sgu{}} & 89.5 & \rf{61.6\sgu{}} & 96.5 & \rf{89.6\sgu{}} & 93.2 & \rf{77.4\sgu{}} & 95.6 & \rf{85.3\sgu{}} & 95.4 & \rf{75.4\sgu{}} \\
\quad medium & 94.8 & \rf{88.7\sgu{}} & 88.3 & \rf{80.6} & 96.8 & \rf{92.4\sgu{}} & 89.7 & \rf{76.9\sgu{}} & 94.7 & \rf{89.9} & 92.5 & \rf{85.7} \\
\quad large & 93.9 & \rf{93.2} & 80.0 & \rf{71.7} & 95.8 & \rf{91.7\sgu{}} & 91.1 & \rf{77.8\sgu{}} & 98.2 & \rf{87.4\sgu{}} & 97.8 & \rf{79.0\sgu{}} \\
\midrule
Objects in corridor: 0 & 99.7 & \rf{99.2\sgu{}} & 89.1 & \rf{82.6} & 99.5 & \rf{98.6\sgu{}} & 90.8 & \rf{80.8\sgu{}} & 98.4 & \rf{97.4\sgu{}} & 74.6 & \rf{63.8\sgu{}} \\
\quad 1 & 97.9 & \rf{96.0\sgu{}} & 87.6 & \rf{77.9\sgu{}} & 98.1 & \rf{96.3\sgu{}} & 92.2 & \rf{85.5\sgu{}} & 97.7 & \rf{93.1\sgu{}} & 95.2 & \rf{83.7\sgu{}} \\
\quad 2 & 97.7 & \rf{95.8\sgu{}} & 86.8 & \rf{78.5\sgu{}} & 97.8 & \rf{95.7\sgu{}} & 91.4 & \rf{83.6\sgu{}} & 98.3 & \rf{94.3\sgu{}} & 96.8 & \rf{86.1\sgu{}} \\
\quad 3+ & 96.8 & \rf{96.1} & 78.8 & \rf{84.8} & 97.2 & \rf{94.5\sgu{}} & 91.7 & \rf{83.5\sgu{}} & 95.4 & \rf{92.2\sgu{}} & 88.4 & \rf{83.1\sgu{}} \\
\bottomrule
\end{tabular}
\caption{\textbf{Agreement and repair rate by condition} (\%; test splits; each metric for \method, then for u512). Day\,/\,night and dry\,/\,rain from the scene descriptions of nuScenes (the Argoverse datasets carry no such labels); decisive box: image area of the annotated box of the decisive object, terciles of the frames that require braking; objects in corridor: annotated objects within $\pm1.2$\,m and $80$\,m. Every cell has at least $309$ frames. \sgu{}: \method significantly higher; u512 is significantly higher in no cell on either metric.}
\label{S-tab:cond}
\end{table}

\noindent\textbf{By condition.}
\Cref{S-tab:cond} splits the test frames by lighting and weather (nuScenes), by the image size of the decisive object, and by the number of objects in the ego corridor.
\method agrees with $\policy(\full)$ at least as often as u512 in every cell, significantly in $25$ of the $28$ cells across the three datasets, and u512 is significantly ahead in no cell on either metric.
The lead is largest where low resolution struggles most: for the smallest decisive objects, \method agrees $6.9$--$10.7$ points more often and repairs $15.8$--$27.9$ points more of the draft's disagreements, and at night it repairs $16.2$ points more.

\subsection{Direction of the departures and the restricted reference}
\label{S-sec:supp_direction}

\begin{table}[!ht]
\centering

\footnotesize
\setlength{\tabcolsep}{3pt}
\begin{tabular}{@{}lccccccccccccccc@{}}
\toprule
 & \multicolumn{5}{c}{nuScenes} & \multicolumn{5}{c}{Argoverse 2} & \multicolumn{5}{c}{Argoverse-HD} \\
\cmidrule(lr){2-6}\cmidrule(lr){7-11}\cmidrule(l){12-16}
Pipeline & Dep. & Less & Less$\wedge$Row & Row & $\full$ & Dep. & Less & Less$\wedge$Row & Row & $\full$ & Dep. & Less & Less$\wedge$Row & Row & $\full$ \\
\midrule
u512 & 2.3 & 54 & 27 & 40 & 56 & 3.6 & 55 & 23 & 38 & 59 & 5.5 & 53 & 17 & 32 & 58 \\
u640 & 1.8 & 57 & 27 & 38 & 59 & 2.7 & 55 & 26 & 43 & 51 & 4.2 & 47 & 19 & 35 & 55 \\
u1280 & 1.7 & 54 & 33 & 56 & 40 & 2.4 & 47 & 28 & 52 & 43 & 3.8 & 61 & 41 & 53 & 37 \\
native & 2.3 & 70 & 42 & 57 & 38 & 4.0 & 57 & 34 & 49 & 45 & 6.3 & 65 & 46 & 56 & 35 \\
\rowcolor{oursbg}\textbf{\method} & 1.3 & 80 & 50 & 61 & 33 & 1.8 & 81 & 53 & 62 & 32 & 2.5 & 77 & 59 & 72 & 21 \\
\bottomrule
\end{tabular}
\caption{\textbf{Departures from the reference action} (test splits). Dep.: frames on which the pipeline takes another action than $\policy(\full)$ (\%); Less: share of these frames on which it brakes less than $\policy(\full)$; Less$\wedge$Row: share on which it brakes less and the annotated scene requires its action; Row / $\full$: share on which the annotated scene requires the action of the pipeline / of $\policy(\full)$ (\%; the annotation can require neither).}
\label{S-tab:direction}
\end{table}

\Cref{S-tab:direction} characterizes the frames on which a pipeline departs from $\policy(\full)$.
\method departs least often, on $1.3$--$2.5\%$ of the frames.
On $77$--$81\%$ of these frames it brakes less than the reference, and the annotated scene requires the action of \method on $61$--$72\%$ of them and that of the reference on $21$--$33\%$.
Uniform downscaling departs differently: u512 and u640 brake less and more about equally often, and the annotation sides with the reference on $51$--$59\%$ of their departures.
Taken jointly, on $50$--$59\%$ of its departures \method brakes less than the reference and the annotated scene requires its action, against $17$--$27\%$ for u512 and at most $46\%$ for any whole-frame resolution: most departures of \method drop braking that the scene does not require, which is the source of its higher precision in \cref{tab:main}.
The detections of the reference itself show where this comes from. Restricting $\full$ to the crops that \method requests on each frame, \ie, keeping the $31$--$42\%$ of its detections whose box center lies in a crop and the draft elsewhere, raises its precision from $61.1\%$ and $70.6\%$ to $69.4\%$ and $75.5\%$ on nuScenes and Argoverse~2, not significantly different from that of \method ($70.7\%$ and $77.1\%$): there, the gain comes from not acting on what the reference detects outside the regions that the planner marks as relevant. On Argoverse-HD the restricted reference stays at $35.4\%$, significantly below \method ($44.3\%$), so that part of the gain there comes from the crop detections and their states. The restricted reference agrees with $\policy(\full)$ on $98.5$--$99.4\%$ of the frames, which bounds what the regions of \method lose by themselves.

\subsection{Held-out sequences}
\label{S-sec:supp_holdout}

\begin{table}[!ht]
\centering

\footnotesize
\setlength{\tabcolsep}{5pt}
\begin{tabular}{@{}lccccccc@{}}
\toprule
Pipeline & px & Agr. & Rep. & P & R & F1 & False\,/\,1k \\
\midrule
\rf{u320 (draft)} & \rf{3.1} & \rf{76.7\sgu{}} & \rf{0\sgu{}} & \rf{7.5\sgu{}} & \rf{81.8\sgd{}} & \rf{13.8\sgu{}} & \rf{39.5\sgd{}} \\
\rf{u512} & \rf{7.1} & \rf{95.8\sgu{}} & \rf{87.1\sgu{}} & \rf{32.0\sgu{}} & \rf{62.3} & \rf{42.3\sgu{}} & \rf{5.2\sgd{}} \\
\rf{u960 ($\full$)} & \rf{25.3} & \rf{100} & \rf{100} & \rf{37.6\sgu{}} & \rf{63.0} & \rf{47.1\sgu{}} & \rf{4.1\sgd{}} \\
\midrule
\method-fast & 6.3 & 98.3\sgu{} & 94.5\sgu{} & 36.0\sgu{} & 62.3 & 45.6\sgu{} & 4.4\sgd{} \\
\rowcolor{oursbg}\textbf{\method} & 6.7 & 98.5 & 95.3 & 43.0 & 61.7 & 50.7 & 3.2 \\
\bottomrule
\end{tabular}
\caption{\textbf{Held-out Argoverse-HD sequences} ($65$ training sequences, $154$ required episodes; \%). Columns as in \cref{tab:fid,tab:main}; False\,/\,1k: false braking episodes per $1{,}000$ frames. \sgu{} / \sgd{}: the value of \method is significantly higher / lower.}
\label{S-tab:holdout}
\end{table}

\noindent\textbf{Protocol and scope.}
\Cref{S-tab:holdout} repeats the comparison on $65$ training sequences of Argoverse-HD, which share the sensor, cities, and annotation process of its test split and were used neither for training nor for configuration selection.

\noindent\textbf{Results.}
\method agrees with $\policy(\full)$ on $98.5\%$ of the frames with $6.7\%$ of the native pixels and repairs $95.3\%$ of the disagreements of the draft, both significantly more than u512.
Its F1 is $50.7$, significantly above that of u512 ($42.3$) and of the reference ($47.1$); it brakes falsely significantly less often than both ($3.2$ against $5.2$ and $4.1$ episodes per $1{,}000$ frames), and its recall does not differ significantly from theirs.
The second round raises the agreement, the repair rate, and the F1 significantly, from $98.3\%$, $94.5\%$, and $45.6$ for \method-fast.

\subsection{Where the remaining disagreement comes from}
\label{S-sec:supp_attribution}

\begin{table}[!ht]
\centering

\footnotesize
\setlength{\tabcolsep}{6pt}
\begin{tabular}{@{}lcccccc@{}}
\toprule
 & \multicolumn{2}{c}{nuScenes} & \multicolumn{2}{c}{Argoverse 2} & \multicolumn{2}{c}{Argoverse-HD} \\
\cmidrule(lr){2-3}\cmidrule(lr){4-5}\cmidrule(l){6-7}
Intervention (cumulative) & Agr. & Diff. & Agr. & Diff. & Agr. & Diff. \\
\midrule
\multicolumn{7}{@{}l}{\emph{With the range refiner (in \method and in $\full$)}} \\
\method & 98.75 & 115 & 98.25 & 207 & 97.52 & 372 \\
+ crop detections of $\full$ & 99.18\sgu{} & 75 & 98.63\sgu{} & 162 & 98.73\sgu{} & 191 \\
+ states of $\full$ for these objects & 99.38\sgu{} & 57 & 99.16\sgu{} & 99 & 98.99\sgu{} & 151 \\
+ probe closing at $10$\,m/s & 99.54\sgu{} & 42 & 99.37\sgu{} & 75 & 99.05\sgu{} & 143 \\
+ extended grid & 99.67\sgu{} & 30 & 99.65\sgu{} & 41 & 99.19\sgu{} & 121 \\
+ three probe sizes & 99.93\sgu{} & 6 & 99.79\sgu{} & 25 & 99.70\sgu{} & 45 \\
+ cap of $10$ rounds & 99.93\sgu{} & 6 & 99.79\sgu{} & 25 & 99.70\sgu{} & 45 \\
\midrule
\multicolumn{7}{@{}l}{\emph{Without the range refiner (in \method and in $\full$; the loop that Theorem~\ref{thm:exact} covers)}} \\
\method & 98.42 & 145 & 97.97 & 240 & 97.46 & 381 \\
+ crop detections of $\full$ & 99.07\sgu{} & 85 & 98.47\sgu{} & 181 & 98.45\sgu{} & 233 \\
+ states of $\full$ for these objects & 99.37\sgu{} & 58 & 99.04\sgu{} & 113 & 98.74\sgu{} & 189 \\
+ probe closing at $10$\,m/s & 99.53\sgu{} & 43 & 99.31\sgu{} & 82 & 98.87\sgu{} & 170 \\
+ extended grid & 99.66\sgu{} & 31 & 99.63\sgu{} & 44 & 99.03\sgu{} & 145 \\
+ three probe sizes & 99.93\sgu{} & 6 & 99.80\sgu{} & 24 & 99.75\sgu{} & 38 \\
+ cap of $10$ rounds & 99.93\sgu{} & 6 & 99.80\sgu{} & 24 & 99.75\sgu{} & 38 \\
\bottomrule
\end{tabular}
\caption{\textbf{Where the remaining disagreement comes from} (test splits). Oracle interventions, applied cumulatively from top to bottom, that replace one approximation of the implementation by its ideal: the detections of $\full$ inside the verified crops (locality of the boxes), the states of $\full$ for these objects (locality of the temporal estimate), and denser or larger probes and more rounds (coverage and termination). Extended grid: ranges from $8$ to $80$\,m and lateral offsets up to $\pm9$\,m; three probe sizes: pedestrian, vehicle, and truck, with the union of their footprints. The first two interventions are not deployable; the others are, at a cost in pixels (\cref{S-tab:grid}). Agr.: agreement with $\policy(\full)$ (\%); Diff.: frames that still disagree. \sgu{}: significantly above the \method row of the same block (unlike the other tables).}
\label{S-tab:attribution}
\end{table}

Theorem~\ref{thm:exact} returns $\policy(\full)$ when locality and coverage hold; the implementation approximates both.
\Cref{S-tab:attribution} replaces the approximations one at a time by their ideal, cumulatively.
Giving the crops the detections of $\full$ (locality of the boxes) and then the states of $\full$ for these objects (locality of the temporal estimate) resolves $50$--$59\%$ of the disagreeing frames; a faster probe, an extended grid, and three probe sizes (coverage) resolve most of the rest, and together the interventions raise the agreement from $97.52$--$98.75\%$ to $99.70$--$99.93\%$.
The coverage interventions can be deployed but enlarge the footprint: alone, a probe closing at $10$\,m/s or a truck-sized probe raises the agreement by $0.1$--$0.3$ points at $1.15$--$1.30\times$ the pixels (\cref{S-tab:grid}).
Raising the cap from three to ten rounds changes no frame: no frame stops at the cap.
Without the range refiner, which lies outside the theorem, the picture is the same.

\begin{table}[!ht]
\centering

\footnotesize
\setlength{\tabcolsep}{6pt}
\begin{tabular}{@{}lcccccc@{}}
\toprule
 & \multicolumn{2}{c}{nuScenes} & \multicolumn{2}{c}{Argoverse 2} & \multicolumn{2}{c}{Argoverse-HD} \\
\cmidrule(lr){2-3}\cmidrule(lr){4-5}\cmidrule(l){6-7}
Range refiner & with & without & with & without & with & without \\
\midrule
Frames with an action $\neq\policy(\full)$ & 115 & 145 & 207 & 240 & 372 & 381 \\
\quad refiner only & 13 & \na & 45 & \na & 97 & \na \\
\quad coverage & 48 & 50 & 73 & 88 & 150 & 209 \\
\quad locality & 36 & 60 & 70 & 106 & 83 & 103 \\
\quad no object of $\full$ raises the action & 18 & 35 & 19 & 46 & 42 & 69 \\
\quad stopped at the round cap & 0 & 0 & 0 & 0 & 0 & 0 \\
\midrule
\multicolumn{7}{@{}l}{\emph{Rounds of the loop (\% of frames)}} \\
0 (draft only) & \multicolumn{2}{c}{35.5} & \multicolumn{2}{c}{28.7} & \multicolumn{2}{c}{24.3} \\
1 & \multicolumn{2}{c}{60.3} & \multicolumn{2}{c}{61.3} & \multicolumn{2}{c}{64.0} \\
2 & \multicolumn{2}{c}{4.1} & \multicolumn{2}{c}{10.0} & \multicolumn{2}{c}{11.6} \\
3 & \multicolumn{2}{c}{0.1} & \multicolumn{2}{c}{0.1} & \multicolumn{2}{c}{0.1} \\
Crop detections matching $\full$ (IoU $\ge0.9$, \%) & \multicolumn{2}{c}{69.0} & \multicolumn{2}{c}{63.1} & \multicolumn{2}{c}{72.8} \\
\bottomrule
\end{tabular}
\caption{\textbf{The disagreeing frames by the condition of Theorem~\ref{thm:exact} that fails} (test splits). Coverage: an object of $\full$ that raises the action of the loop has its box center outside every crop; locality: all such objects lie inside crops but are missed or lifted differently; refiner only: the loop agrees with $\policy(\full)$ without the range refiner. Bottom: rounds per frame.}
\label{S-tab:conditions}
\end{table}

\Cref{S-tab:conditions} classifies the disagreeing frames directly by the condition that fails; unlike the cumulative interventions of \cref{S-tab:attribution}, it assigns each frame to one condition, so the shares of the two tables differ.
With the refiner, $42\%$, $35\%$, and $40\%$ of them violate coverage (an object of $\full$ that would raise the action lies outside every crop), $31\%$, $34\%$, and $22\%$ violate locality, and $11\%$, $22\%$, and $26\%$ are due to the refiner alone; on the remaining $16\%$, $9\%$, and $11\%$, \method brakes more than $\policy(\full)$ without any object of $\full$ that would raise the action, \ie, on a detection that $\full$ lacks or lifts to a lower demand; since conditions (i), (ii), and coverage hold on these frames, this too is a violation of locality, and \cref{sec:exp_exact} counts the two together as locality ($34$--$47\%$).
Most frames need one round; a second round occurs on $4$--$12\%$ of the frames and a third on fewer than $0.2\%$.
About two thirds of the crop detections ($63$--$73\%$) match a detection of $\full$ with an IoU of at least $0.9$.

\begin{table}[!ht]
\centering

\footnotesize
\setlength{\tabcolsep}{4.5pt}
\begin{tabular}{@{}lccccccccc@{}}
\toprule
 & \multicolumn{3}{c}{nuScenes} & \multicolumn{3}{c}{Argoverse 2} & \multicolumn{3}{c}{Argoverse-HD} \\
\cmidrule(lr){2-4}\cmidrule(lr){5-7}\cmidrule(l){8-10}
Pipeline & Agr. & Rep. & Trans. & Agr. & Rep. & Trans. & Agr. & Rep. & Trans. \\
\midrule
\rf{u960, frame shifted by $1$ pixel} & \rf{99.2\sgd{}} & \rf{93.6\sgd{}} & \rf{90.0\sgd{}} & \rf{98.8\sgd{}} & \rf{95.2\sgd{}} & \rf{90.4\sgd{}} & \rf{98.7\sgd{}} & \rf{96.4\sgd{}} & \rf{77.1\sgd{}} \\
\rf{u960, half-pixel phase} & \rf{99.5\sgd{}} & \rf{96.0\sgd{}} & \rf{94.2\sgd{}} & \rf{99.1\sgd{}} & \rf{96.3\sgd{}} & \rf{91.4\sgd{}} & \rf{99.0\sgd{}} & \rf{96.8\sgd{}} & \rf{80.6\sgd{}} \\
\rf{u512} & \rf{97.7\sgu{}} & \rf{79.6\sgu{}} & \rf{76.7\sgu{}} & \rf{96.4\sgu{}} & \rf{84.0\sgu{}} & \rf{74.9\sgu{}} & \rf{94.5\sgu{}} & \rf{82.6\sgu{}} & \rf{54.0\sgu{}} \\
\rf{u768} & \rf{98.6} & \rf{88.6} & \rf{84.7} & \rf{97.8\sgu{}} & \rf{90.4} & \rf{80.9} & \rf{96.1\sgu{}} & \rf{88.5\sgu{}} & \rf{56.8\sgu{}} \\
\rf{u1280} & \rf{98.3\sgu{}} & \rf{87.8} & \rf{84.2} & \rf{97.6\sgu{}} & \rf{91.8} & \rf{82.1} & \rf{96.2\sgu{}} & \rf{90.0} & \rf{60.5\sgu{}} \\
\midrule
\method-fast & 98.7 & 86.8 & 84.4 & 98.2 & 91.5 & 83.4 & 97.4\sgu{} & 91.6 & 67.7 \\
\rowcolor{oursbg}\textbf{\method} & 98.7 & 86.9 & 84.4 & 98.2 & 91.7 & 83.4 & 97.5 & 92.0 & 68.1 \\
\bottomrule
\end{tabular}
\caption{\textbf{Noise floor of the agreement} (\%; test splits). The reference recomputed on the same frame shifted by one pixel, or resampled with a half-pixel phase, with the same lift and its own history, measures how much $\policy(\full)$ changes under perturbations that carry no information. Columns as in \cref{S-tab:fid_full}. \sgu{} / \sgd{}: \method significantly higher / lower.}
\label{S-tab:ceiling}
\end{table}

\noindent\textbf{Noise floor of the agreement.}
The reference itself is not stable under perturbations that carry no information (\cref{S-tab:ceiling}): recomputed on the frame shifted by one pixel, it agrees with $\policy(\full)$ on $99.15\%$, $98.79\%$, and $98.72\%$ of the frames, and with a half-pixel resampling phase on $99.0$--$99.5\%$.
The disagreement of \method ($1.3$--$2.5\%$) is thus about one and a half to two times that of the shifted copy of the reference, and on the frames where $\policy(\full)$ changes, the shifted copy agrees on only $90\%$, $90\%$, and $77\%$, against $84\%$, $83\%$, and $68\%$ for \method.

\subsection{A second detector and a native-resolution reference}
\label{S-sec:supp_yolo}

\begin{table}[!ht]
\centering

\footnotesize
\setlength{\tabcolsep}{3.4pt}
\begin{tabular}{@{}lcccccccccccc@{}}
\toprule
 & \multicolumn{4}{c}{nuScenes} & \multicolumn{4}{c}{Argoverse 2} & \multicolumn{4}{c}{Argoverse-HD} \\
\cmidrule(lr){2-5}\cmidrule(lr){6-9}\cmidrule(l){10-13}
Pipeline & px & Agr. & Rep. & F1 & px & Agr. & Rep. & F1 & px & Agr. & Rep. & F1 \\
\midrule
\rf{u320 (draft)} & \rf{4.3} & \rf{93.1\sgu{}} & \rf{0\sgu{}} & \rf{47.3\sgu{}} & \rf{2.6} & \rf{85.8\sgu{}} & \rf{0\sgu{}} & \rf{41.4\sgu{}} & \rf{3.1} & \rf{77.8\sgu{}} & \rf{0\sgu{}} & \rf{20.2\sgu{}} \\
\rf{u320$^{+}$} & \rf{4.3} & \rf{96.7\sgu{}} & \rf{59.9\sgu{}} & \rf{60.3\sgu{}} & \rf{2.6} & \rf{94.4\sgu{}} & \rf{67.5\sgu{}} & \rf{68.2\sgu{}} & \rf{3.1} & \rf{92.0\sgu{}} & \rf{66.6\sgu{}} & \rf{36.2\sgu{}} \\
\rf{u512} & \rf{10.2} & \rf{97.7\sgu{}} & \rf{77.3\sgu{}} & \rf{59.7\sgu{}} & \rf{6.7} & \rf{96.7\sgu{}} & \rf{85.2\sgu{}} & \rf{72.3\sgu{}} & \rf{7.1} & \rf{95.6\sgu{}} & \rf{86.8} & \rf{35.2\sgu{}} \\
\rf{u640} & \rf{17.1} & \rf{98.2\sgu{}} & \rf{83.8} & \rf{62.0\sgu{}} & \rf{10.3} & \rf{97.3\sgu{}} & \rf{89.1\sgu{}} & \rf{74.8} & \rf{11.6} & \rf{97.3} & \rf{92.1} & \rf{38.6\sgu{}} \\
\rf{u960 ($\full$)} & \rf{38.4} & \rf{100} & \rf{100} & \rf{65.0\sgu{}} & \rf{23.2} & \rf{100} & \rf{100} & \rf{77.3} & \rf{25.3} & \rf{100} & \rf{100} & \rf{41.9} \\
\rf{u1280} & \rf{65.4} & \rf{98.5} & \rf{88.2} & \rf{64.5\sgu{}} & \rf{40.0} & \rf{97.7\sgu{}} & \rf{91.5} & \rf{74.5} & \rf{44.4} & \rf{97.1} & \rf{91.8} & \rf{42.0} \\
\midrule
Support only (w/o field) & 5.6 & 97.7\sgu{} & 71.2\sgu{} & 63.8\sgu{} & 4.2 & 97.3\sgu{} & 85.1\sgu{} & 75.3 & 4.9 & 96.0\sgu{} & 84.3\sgu{} & 45.6 \\
\midrule
\method-fast & 8.1 & 98.7 & 86.0 & 69.0 & 5.7 & 98.1 & 91.9 & 75.9 & 6.4 & 97.2\sgu{} & 89.9\sgu{} & 41.7\sgu{} \\
\rowcolor{oursbg}\textbf{\method} & 8.3 & 98.7 & 86.3 & 69.2 & 5.9 & 98.2 & 92.1 & 76.5 & 6.8 & 97.5 & 91.0 & 46.6 \\
\bottomrule
\end{tabular}
\caption{\textbf{A second detector: YOLO11s} as the cheap and the reference detector (\%; test splits), with the configuration of \method, the lift, and the planner unchanged; $\full$ is YOLO11s at the verification resolution. Columns as in \cref{tab:fid,tab:main}. Top block: whole-frame inference. \sgu{} / \sgd{}: \method significantly higher / lower.}
\label{S-tab:yolo}
\end{table}

\Cref{S-tab:yolo} repeats the main comparison with YOLO11s~\cite{yolo11} as the cheap and the reference detector, keeping the configuration of \method, the lift, and the planner.
\method agrees with $\policy(\full)$ on $97.5$--$98.7\%$ of the frames with $5.9$--$8.3\%$ of the native pixels, significantly more than u512 on all three datasets, and its F1 ($69.2$, $76.5$, and $46.6$) is significantly above that of u512 on all three and above that of the reference on nuScenes (significantly) and Argoverse-HD.
The behavior of \method thus carries over to a second detector without retuning.

\noindent\textbf{A native-resolution reference.}
The result does not depend on the choice of u960 as the reference either. Detected at the native resolution instead of the verification resolution, the regions of \method agree with native-resolution inference on $98.4\%$, $97.1\%$, and $96.8\%$ of the frames with $14$--$16\%$ of the native pixels, significantly more often than u960 does ($97.7\%$, $96.0\%$, and $93.7\%$); \cref{S-tab:unc} lists this variant (``crops at native scale'') with its agreement against $\policy(\full)$.

\subsection{A detector fine-tuned on driving data}
\label{S-sec:supp_ft}

\begin{table}[!ht]
\centering

\scriptsize
\setlength{\tabcolsep}{2.1pt}
\begin{tabular}{@{}lcccccccccccccccccc@{}}
\toprule
 & \multicolumn{6}{c}{nuScenes} & \multicolumn{6}{c}{Argoverse 2} & \multicolumn{6}{c}{Argoverse-HD} \\
\cmidrule(lr){2-7}\cmidrule(lr){8-13}\cmidrule(l){14-19}
Pipeline & px & Agr. & Rep. & P & R & F1 & px & Agr. & Rep. & P & R & F1 & px & Agr. & Rep. & P & R & F1 \\
\midrule
\rf{u320 (draft)} & \rf{4.3} & \rf{93.1\sgu{}} & \rf{0\sgu{}} & \rf{35.6\sgu{}} & \rf{75.5} & \rf{48.4\sgu{}} & \rf{2.6} & \rf{78.3\sgu{}} & \rf{0\sgu{}} & \rf{25.7\sgu{}} & \rf{91.1\sgd{}} & \rf{40.1\sgu{}} & \rf{3.1} & \rf{78.9\sgu{}} & \rf{0\sgu{}} & \rf{12.0\sgu{}} & \rf{77.5\sgd{}} & \rf{20.8\sgu{}} \\
\rf{u320$^{+}$} & \rf{4.3} & \rf{96.2\sgu{}} & \rf{56.7\sgu{}} & \rf{50.0\sgu{}} & \rf{66.7\sgu{}} & \rf{57.1\sgu{}} & \rf{2.6} & \rf{91.9\sgu{}} & \rf{69.5\sgu{}} & \rf{45.2\sgu{}} & \rf{86.3} & \rf{59.3\sgu{}} & \rf{3.1} & \rf{93.4\sgu{}} & \rf{73.5\sgu{}} & \rf{26.4\sgu{}} & \rf{54.9} & \rf{35.6\sgu{}} \\
\rf{u512} & \rf{10.2} & \rf{96.8\sgu{}} & \rf{71.2\sgu{}} & \rf{51.5} & \rf{70.7} & \rf{59.6\sgu{}} & \rf{6.7} & \rf{92.9\sgu{}} & \rf{78.8\sgu{}} & \rf{46.3\sgu{}} & \rf{86.8} & \rf{60.4\sgu{}} & \rf{7.1} & \rf{94.6\sgu{}} & \rf{85.0} & \rf{26.6\sgu{}} & \rf{57.7} & \rf{36.4} \\
\rf{u640} & \rf{17.1} & \rf{97.4} & \rf{77.6} & \rf{48.0\sgu{}} & \rf{72.1} & \rf{57.6\sgu{}} & \rf{10.3} & \rf{93.5\sgu{}} & \rf{81.2\sgu{}} & \rf{44.7\sgu{}} & \rf{86.8} & \rf{59.0\sgu{}} & \rf{11.6} & \rf{95.9} & \rf{87.3} & \rf{28.6\sgu{}} & \rf{56.3} & \rf{37.9\sgu{}} \\
\rf{u768} & \rf{23.9} & \rf{97.7} & \rf{80.2} & \rf{49.3\sgu{}} & \rf{72.8} & \rf{58.8\sgu{}} & \rf{14.7} & \rf{94.0\sgu{}} & \rf{83.2} & \rf{46.7\sgu{}} & \rf{86.3} & \rf{60.6\sgu{}} & \rf{16.0} & \rf{96.1} & \rf{89.1} & \rf{31.4} & \rf{60.6} & \rf{41.3} \\
\rf{u960 ($\full$)} & \rf{38.4} & \rf{\na} & \rf{\na} & \rf{48.4\sgu{}} & \rf{72.8} & \rf{58.2\sgu{}} & \rf{23.2} & \rf{\na} & \rf{\na} & \rf{51.4\sgu{}} & \rf{87.9\sgd{}} & \rf{64.9\sgu{}} & \rf{25.3} & \rf{\na} & \rf{\na} & \rf{31.3} & \rf{59.2} & \rf{41.0} \\
\rf{u1280} & \rf{65.4} & \rf{96.4\sgu{}} & \rf{75.0\sgu{}} & \rf{49.1\sgu{}} & \rf{74.1} & \rf{59.1\sgu{}} & \rf{40.0} & \rf{93.0\sgu{}} & \rf{84.1} & \rf{55.5} & \rf{85.3} & \rf{67.2} & \rf{44.4} & \rf{94.6\sgu{}} & \rf{86.2} & \rf{28.8\sgu{}} & \rf{56.3} & \rf{38.1} \\
\rf{native} & \rf{103.1} & \rf{95.2\sgu{}} & \rf{70.9\sgu{}} & \rf{50.5} & \rf{67.3\sgu{}} & \rf{57.7\sgu{}} & \rf{101.2} & \rf{87.0\sgu{}} & \rf{78.6\sgu{}} & \rf{61.4\sgd{}} & \rf{81.1} & \rf{69.8} & \rf{101.3} & \rf{85.3\sgu{}} & \rf{75.2\sgu{}} & \rf{25.9\sgu{}} & \rf{50.7\sgu{}} & \rf{34.3\sgu{}} \\
\midrule
Chanakya~\cite{chanakya} & 7.9 & 94.7\sgu{} & 29.6\sgu{} & 37.4\sgu{} & 74.8 & 49.9\sgu{} & 4.9 & 82.7\sgu{} & 23.1\sgu{} & 25.2\sgu{} & 91.1\sgd{} & 39.5\sgu{} & 6.3 & 87.4\sgu{} & 42.8\sgu{} & 11.9\sgu{} & 74.6\sgd{} & 20.5\sgu{} \\
BAR~\cite{bgtada} & 36.6 & 95.1\sgu{} & 44.4\sgu{} & 38.7\sgu{} & 73.5 & 50.7\sgu{} & 21.4 & 86.0\sgu{} & 41.8\sgu{} & 32.1\sgu{} & 89.5\sgd{} & 47.2\sgu{} & 23.1 & 87.9\sgu{} & 47.0\sgu{} & 15.3\sgu{} & 66.2\sgd{} & 24.8\sgu{} \\
SAHI~\cite{sahi} & 187.7 & 92.1\sgu{} & 71.7\sgu{} & 35.5\sgu{} & 78.2 & 48.8\sgu{} & 165.2 & 87.7\sgu{} & 72.5\sgu{} & 39.4\sgu{} & 92.6\sgd{} & 55.3\sgu{} & 153.8 & 85.3\sgu{} & 76.0\sgu{} & 22.2\sgu{} & 71.8\sgd{} & 33.9\sgu{} \\
CF~\cite{cfimportance} & 6.0 & 97.1\sgu{} & 68.5\sgu{} & 54.9 & 65.3\sgu{} & 59.6 & 4.5 & 92.2\sgu{} & 71.7\sgu{} & 42.8\sgu{} & 83.2 & 56.5\sgu{} & 5.6 & 92.9\sgu{} & 70.3\sgu{} & 21.2\sgu{} & 59.2 & 31.2\sgu{} \\
FOVEA~\cite{fovea} & 69.7 & 95.4\sgu{} & 71.7\sgu{} & 46.4\sgu{} & 74.8 & 57.3\sgu{} & 42.6 & 87.1\sgu{} & 73.0\sgu{} & 39.3\sgu{} & 86.3 & 54.0\sgu{} & 47.6 & 91.6\sgu{} & 79.1\sgu{} & 25.0\sgu{} & 53.5\sgu{} & 34.1\sgu{} \\
AdaScale~\cite{adascale} & 30.6 & 97.5 & 80.3 & 49.5\sgu{} & 72.1 & 58.7\sgu{} & 26.8 & 96.8\sgd{} & 91.3\sgd{} & 48.4\sgu{} & 88.4 & 62.6\sgu{} & 29.9 & 97.9\sgd{} & 94.0\sgd{} & 28.7\sgu{} & 60.6 & 38.9\sgu{} \\
AnA~\cite{ana} & 11.0 & 97.5 & 73.2\sgu{} & 53.3 & 66.7\sgu{} & 59.2\sgu{} & 9.7 & 94.9 & 83.7 & 53.0\sgu{} & 83.2 & 64.8\sgu{} & 26.0 & 95.8 & 82.2\sgu{} & 31.5 & 56.3 & 40.4 \\
UFP~\cite{ufpmp} & 17.6 & 96.4\sgu{} & 73.2\sgu{} & 47.2\sgu{} & 70.1 & 56.4\sgu{} & 9.5 & 92.1\sgu{} & 79.3\sgu{} & 44.0\sgu{} & 88.4\sgd{} & 58.7\sgu{} & 12.6 & 92.8\sgu{} & 81.0\sgu{} & 26.0\sgu{} & 63.4 & 36.9\sgu{} \\
\midrule
\method-fast & 9.6 & 97.8 & 79.5 & 55.9 & 74.1 & 63.7 & 7.3 & 95.1 & 84.6 & 55.7 & 84.7 & 67.2 & 8.5 & 96.6 & 88.8 & 30.3\sgu{} & 60.6 & 40.4\sgu{} \\
\rowcolor{oursbg}\textbf{\method} & 10.0 & 97.8 & 79.8 & 55.6 & 74.1 & 63.6 & 8.1 & 95.3 & 85.2 & 56.4 & 85.3 & 67.9 & 9.1 & 96.7 & 89.1 & 34.1 & 60.6 & 43.7 \\
\bottomrule
\end{tabular}
\caption{\textbf{A detector fine-tuned on driving data} (\%; test splits): YOLOv8s fine-tuned on the training images of nuScenes and Argoverse~2 as the cheap and the reference detector, with the looming estimate and the range refiner retrained on its crops; the configuration of \method, the confidence threshold, and the planner are unchanged. px: detector-input pixels (\% of native); Agr., Rep.: as in \cref{tab:fid}; P, R, F1: braking episodes as in \cref{tab:main}; \na: $100$ by definition. Top block: whole-frame inference. \sgu{} / \sgd{}: \method significantly higher / lower.}
\label{S-tab:ft}
\end{table}

The detector of the main paper is pretrained on COCO only.
\Cref{S-tab:ft} repeats the main comparison with YOLOv8s fine-tuned on the front-camera images of the nuScenes and Argoverse~2 training sets, excluding every development and test scene or log ($15$ epochs at $640$ pixels, checkpoint selected on held-out training scenes), as the cheap and the reference detector; the looming estimate and the range refiner are retrained on its crops, and the configuration of \method, the confidence threshold ($0.25$), and the planner are unchanged.
\method agrees with $\policy(\full)$ on $95.3$--$97.8\%$ of the frames, significantly more often than u512 on all three datasets and, on Argoverse~2, than every other whole-frame resolution.
Its F1 is the highest of all pipelines on nuScenes (with \method-fast) and on Argoverse-HD, and the second highest on Argoverse~2, below native-resolution inference (not significantly); it is significantly above those of u512 and the reference on nuScenes and Argoverse~2.
It uses $2.8$--$3.8\times$ fewer pixels than the reference.

\subsection{A geometric lift in every pipeline}
\label{S-sec:supp_geo}

The learned parts of the lift, the looming estimate and the range refiner, are shared by every pipeline of the main paper except the draft, including the reference.
With both removed from every pipeline, ranges come from the ground plane and class heights and the time to collision from the growth of the box, and $\policy(\full)$ is recomputed with this lift; the crops, and hence the pixels, stay close to those of \cref{tab:fid} (\method: $8.2\%$, $5.8\%$, and $6.9\%$).
Fidelity carries over: \method agrees with $\policy(\full)$ on $96.7\%$, $91.2\%$, and $88.0\%$ of the frames, $3.3$, $7.3$, and $10.7$ points more than u512 (significantly on all three datasets).
The fidelity of \method thus does not depend on the learned parts of the lift.

\subsection{Pixels and FLOPs}
\label{S-sec:supp_px}

\begin{table}[!ht]
\centering

\footnotesize
\setlength{\tabcolsep}{8pt}
\begin{tabular}{@{}lccc@{}}
\toprule
 & nuScenes & Argoverse 2 & Argoverse-HD \\
\midrule
\multicolumn{4}{@{}l}{\emph{Frames}} \\
Frames with the draft only (\%) & 35.5 & 28.7 & 24.3 \\
\midrule
\multicolumn{4}{@{}l}{\emph{Detector-input pixels per frame (\% of native)}} \\
Mean & 8.4 & 5.9 & 6.9 \\
Median & 7.4 & 4.7 & 5.3 \\
\midrule
\multicolumn{4}{@{}l}{\emph{Mean pixels by source (\% of native)}} \\
Draft & 4.27 & 2.58 & 3.11 \\
Field footprint & 1.96 & 1.25 & 1.33 \\
Support crops & 0.52 & 0.49 & 0.39 \\
Crops of objects verified in the previous frame & 0.50 & 0.57 & 0.92 \\
Truncation guard & 0.88 & 0.89 & 1.10 \\
Previous-frame re-detection & 0.24 & 0.12 & 0.08 \\
\bottomrule
\end{tabular}
\caption{\textbf{Where the pixels of \method go} (test splits). Frames on which the draft alone decides, the mean and median detector-input pixels per frame, and their mean by source; the sources sum to the mean.}
\label{S-tab:pxstats}
\end{table}

\Cref{S-tab:pxstats} breaks down the cost of \method.
On $35\%$, $29\%$, and $24\%$ of the frames, the draft alone decides: the field and the support request nothing.
The draft accounts for $44$--$51\%$ of the pixels, the field footprint for $19$--$23\%$, and the truncation guard for $11$--$16\%$; the support crops, the crops of objects verified in the previous frame, and the previous-frame re-detections take the rest.

\begin{table}[!ht]
\centering

\scriptsize
\setlength{\tabcolsep}{3.4pt}
\begin{tabular}{@{}lccccccccc@{}}
\toprule
 & \multicolumn{3}{c}{nuScenes} & \multicolumn{3}{c}{Argoverse 2} & \multicolumn{3}{c}{Argoverse-HD} \\
\cmidrule(lr){2-4}\cmidrule(lr){5-7}\cmidrule(l){8-10}
Pipeline & Det. & Lift & Total & Det. & Lift & Total & Det. & Lift & Total \\
\midrule
\rf{u320 (draft)} & \rf{4.3} & \rf{0.0} & \rf{4.3} & \rf{5.8} & \rf{0.0} & \rf{5.8} & \rf{5.0} & \rf{0.0} & \rf{5.0} \\
\rf{u320$^{+}$} & \rf{4.3} & \rf{5.3} & \rf{9.6} & \rf{5.8} & \rf{13.2} & \rf{19.0} & \rf{5.0} & \rf{15.6} & \rf{20.6} \\
\rf{u512} & \rf{10.4} & \rf{7.4} & \rf{17.8} & \rf{15.0} & \rf{18.1} & \rf{33.1} & \rf{11.5} & \rf{21.1} & \rf{32.6} \\
\rf{u640} & \rf{17.3} & \rf{8.8} & \rf{26.1} & \rf{23.0} & \rf{19.8} & \rf{42.9} & \rf{18.7} & \rf{24.9} & \rf{43.6} \\
\rf{u768} & \rf{24.2} & \rf{9.4} & \rf{33.6} & \rf{32.8} & \rf{21.1} & \rf{53.9} & \rf{25.9} & \rf{26.6} & \rf{52.5} \\
\rf{u960 ($\full$)} & \rf{38.9} & \rf{10.0} & \rf{48.9} & \rf{51.9} & \rf{23.0} & \rf{74.9} & \rf{41.1} & \rf{29.7} & \rf{70.8} \\
\rf{u1280} & \rf{66.3} & \rf{10.5} & \rf{76.8} & \rf{89.3} & \rf{25.6} & \rf{114.9} & \rf{72.0} & \rf{33.6} & \rf{105.6} \\
\rf{native} & \rf{104.4} & \rf{10.6} & \rf{115.0} & \rf{225.9} & \rf{28.0} & \rf{253.9} & \rf{164.2} & \rf{36.8} & \rf{201.0} \\
\midrule
Chanakya~\cite{chanakya} & 8.0 & 1.2 & 9.2 & 10.9 & 3.9 & 14.8 & 10.3 & 6.9 & 17.2 \\
BAR~\cite{bgtada} & 37.0 & 6.6 & 43.6 & 47.8 & 14.0 & 61.8 & 37.5 & 14.8 & 52.3 \\
SAHI~\cite{sahi} & 190.2 & 11.5 & 201.7 & 368.8 & 29.0 & 397.8 & 249.2 & 39.0 & 288.2 \\
CF~\cite{cfimportance} & 5.4 & 0.7 & 6.2 & 8.4 & 3.5 & 11.9 & 7.4 & 3.6 & 10.9 \\
FOVEA~\cite{fovea} & 70.6 & 9.5 & 80.1 & 95.1 & 21.5 & 116.6 & 77.1 & 30.6 & 107.7 \\
AdaScale~\cite{adascale} & 31.0 & 9.0 & 40.0 & 59.8 & 22.6 & 82.4 & 48.4 & 29.1 & 77.5 \\
AnA~\cite{ana} & 8.4 & 2.5 & 10.9 & 13.0 & 10.5 & 23.6 & 18.9 & 15.1 & 34.0 \\
UFP~\cite{ufpmp} & 18.3 & 8.4 & 26.7 & 20.0 & 19.3 & 39.4 & 20.7 & 26.7 & 47.5 \\
\midrule
\method-fast & 8.2 & 2.6 & 10.8 & 12.5 & 6.5 & 19.0 & 10.6 & 10.9 & 21.5 \\
\rowcolor{oursbg}\textbf{\method} & 8.5 & 2.9 & 11.4 & 13.2 & 8.1 & 21.3 & 11.2 & 13.9 & 25.1 \\
\bottomrule
\end{tabular}
\caption{\textbf{FLOPs per frame} (GFLOPs, mean over the test frames). Det.: the detector, proportional to the detector-input pixels; Lift: the looming estimate (once per round for the loops) and the range refiner (once, after the loop) on the objects that each pipeline passes through them; Total: their sum. Not counted: the planner queries of the field and the support (closed form here) and the selection networks of AdaScale and the routers. Top block: whole-frame inference.}
\label{S-tab:flops}
\end{table}

\noindent\textbf{FLOPs.}
YOLOv8s performs $70.3$\,kFLOPs per input pixel at every input size of the experiments ($36$ canvases from $32\times32$ to $2048\times1568$, counted with thop), so the detector FLOPs of a frame are proportional to its detector-input pixels.
All FLOPs follow the convention of this counter for the detector, two FLOPs per multiply-accumulate.
The looming estimate and the refiner are standard ResNet-18 networks applied to $64\times64$ crops that are upsampled to $128\times128$ pixels ($1.19$\,GFLOPs per crop), so the lift adds $3.57$\,GFLOPs per object for the looming estimate (three crops) and $1.19$\,GFLOPs per object for the refiner.
\Cref{S-tab:flops} counts them for the objects that each pipeline passes through them: whole-frame pipelines pass every detection with a same-scale chain over three frames through the looming estimate, \method only its verified objects.
The detector FLOPs of \method are $22\%$, $25\%$, and $27\%$ of those of the reference, the ratios of the pixels.
In total, \method uses $11.4$, $21.3$, and $25.1$\,GFLOPs per frame on average, $23\%$, $28\%$, and $36\%$ of the reference, and less than u512 ($17.8$, $33.1$, and $32.6$) on every dataset.

\section{Field and support}
\label{S-sec:supp_field}

\subsection{Ablation}
\label{S-sec:supp_ablation}

\begin{table}[!ht]
\centering

\footnotesize
\setlength{\tabcolsep}{3.2pt}
\begin{tabular}{@{}lcccccccccccc@{}}
\toprule
 & \multicolumn{4}{c}{nuScenes} & \multicolumn{4}{c}{Argoverse 2} & \multicolumn{4}{c}{Argoverse-HD} \\
\cmidrule(lr){2-5}\cmidrule(lr){6-9}\cmidrule(l){10-13}
Variant & px & Agr. & P & R & px & Agr. & P & R & px & Agr. & P & R \\
\midrule
\textbf{\method (full)} & 8.4 & 98.7 & 70.7 & 67.3 & 5.9 & 98.2 & 77.1 & 77.9 & 6.9 & 97.5 & 44.3 & 49.3 \\
\midrule
Support only (w/o field) & 5.7 & 97.9\sgu{} & 74.3 & 53.1\sgu{} & 4.1 & 97.2\sgu{} & 79.5\sgd{} & 71.6\sgu{} & 5.0 & 96.4\sgu{} & 46.2 & 42.3\sgu{} \\
Field only (w/o support) & 7.1 & 95.7\sgu{} & 36.7\sgu{} & 72.1 & 4.6 & 90.8\sgu{} & 30.4\sgu{} & 83.7\sgd{} & 5.3 & 86.9\sgu{} & 11.0\sgu{} & 70.4\sgd{} \\
Pivots instead of support & 8.3 & 98.6\sgu{} & 69.9 & 68.0 & 5.7 & 97.9\sgu{} & 71.8\sgu{} & 77.9 & 6.8 & 96.8\sgu{} & 23.6\sgu{} & 52.1 \\
Single round & 8.1 & 98.7 & 68.1\sgu{} & 66.7 & 5.6 & 98.2 & 75.6 & 78.4 & 6.5 & 97.4\sgu{} & 39.1\sgu{} & 50.7 \\
W/o truncation guard & 7.5 & 98.0\sgu{} & 63.3\sgu{} & 64.6\sgu{} & 5.1 & 97.0\sgu{} & 66.7\sgu{} & 78.9 & 5.9 & 97.1\sgu{} & 33.0\sgu{} & 50.7 \\
W/o range refiner & 8.4 & 96.9\sgu{} & 57.6\sgu{} & 69.4 & 5.9 & 95.2\sgu{} & 66.5\sgu{} & 80.5\sgd{} & 6.9 & 94.8\sgu{} & 39.1\sgu{} & 50.7 \\
\bottomrule
\end{tabular}
\caption{\textbf{Ablation: pixels, agreement, precision, and recall} (\%; test splits; complements the repair rate and F1 of \cref{tab:ablation}, whose px column averages the unrounded values of the three datasets). \sgu{} / \sgd{}: the full procedure is significantly higher / lower.}
\label{S-tab:ablation_full}
\end{table}

\Cref{S-tab:ablation_full} adds the pixels, the agreement, and the precision and recall behind \cref{tab:ablation}.
The field and the support act on opposite errors.
Without the field, the recall falls by $14$, $6$, and $7$ points (all significant) at $1.8$--$2.7$ points fewer pixels: the support reaches what the draft does not show only incidentally.
Without the support, the recall is about that of the draft and the precision falls to $11$--$37\%$, since spurious draft detections outside the field footprint are never verified.
Removal pivots in place of the support keep the recall and lose precision, most on Argoverse-HD ($23.6\%$ against $44.3\%$), consistent with construction~(c) of \cref{S-sec:constructions}: draft detections that each justify the action leave no pivot to verify.
The truncation guard costs $0.8$--$1.0$ points of pixels and buys $7$--$11$ points of precision, and the range refiner buys $5$--$13$ points of precision without changing the pixels.

\subsection{Recovery of objects that the draft lacks}
\label{S-sec:supp_missed}

\begin{table}[!ht]
\centering

\footnotesize
\setlength{\tabcolsep}{4.5pt}
\begin{tabular}{@{}lccccccccc@{}}
\toprule
 & \multicolumn{3}{c}{nuScenes} & \multicolumn{3}{c}{Argoverse 2} & \multicolumn{3}{c}{Argoverse-HD} \\
\cmidrule(lr){2-4}\cmidrule(lr){5-7}\cmidrule(l){8-10}
Pipeline & px & Natural & Deleted & px & Natural & Deleted & px & Natural & Deleted \\
\midrule
\rf{u448} & \rf{8.0} & \rf{21.5\sgu{}} & \rf{98.5\sgd{}} & \rf{5.0} & \rf{32.3\sgu{}} & \rf{99.0\sgd{}} & \rf{5.6} & \rf{43.3\sgu{}} & \rf{96.0} \\
\rf{u512} & \rf{10.2} & \rf{23.3\sgu{}} & \rf{98.3\sgd{}} & \rf{6.7} & \rf{37.6\sgu{}} & \rf{99.3\sgd{}} & \rf{7.1} & \rf{52.0\sgu{}} & \rf{96.3} \\
\rf{u640} & \rf{17.1} & \rf{33.3} & \rf{97.8\sgd{}} & \rf{10.3} & \rf{45.2\sgu{}} & \rf{99.6\sgd{}} & \rf{11.6} & \rf{64.2\sgu{}} & \rf{98.0} \\
\rf{u960 ($\full$)} & \rf{38.4} & \rf{43.3} & \rf{98.7\sgd{}} & \rf{23.2} & \rf{55.5\sgd{}} & \rf{99.5\sgd{}} & \rf{25.3} & \rf{74.5} & \rf{99.2\sgd{}} \\
\rf{u1280} & \rf{65.4} & \rf{48.1\sgd{}} & \rf{99.0\sgd{}} & \rf{40.0} & \rf{63.3\sgd{}} & \rf{99.6\sgd{}} & \rf{44.4} & \rf{76.3} & \rf{99.6\sgd{}} \\
\midrule
CF~\cite{cfimportance} & 5.3 & 11.1\sgu{} & 22.4\sgu{} & 3.8 & 22.0\sgu{} & 22.1\sgu{} & 4.5 & 15.7\sgu{} & 21.5\sgu{} \\
AnA~\cite{ana} & 8.3 & 12.6\sgu{} & 25.5\sgu{} & 5.8 & 25.7\sgu{} & 26.0\sgu{} & 11.7 & 18.3\sgu{} & 26.1\sgu{} \\
\midrule
Support only (w/o field) & 5.7 & 20.7\sgu{} & 40.7\sgu{} & 4.1 & 27.8\sgu{} & 58.4\sgu{} & 5.0 & 46.0\sgu{} & 57.5\sgu{} \\
\rowcolor{oursbg}\textbf{\method} & 8.4 & 41.5 & 93.3 & 5.9 & 51.1 & 95.9 & 6.9 & 75.5 & 97.4 \\
\bottomrule
\end{tabular}
\caption{\textbf{Recovery of decision-relevant objects that the draft lacks} (\%; test splits; complements \cref{fig:recovery}). Natural: annotated objects that alone require braking and that the draft misses ($270$, $713$, and $1{,}378$ objects); Deleted: the decision-relevant objects that the draft detects ($823$, $2{,}923$, and $2{,}723$ objects), with their draft detections deleted before the loop; px: detector-input pixels (\% of native). Top block: whole-frame inference. \sgu{} / \sgd{}: \method significantly higher / lower.}
\label{S-tab:recovery}
\end{table}

\Cref{S-tab:recovery} gives the values of \cref{fig:recovery} and adds a controlled variant.
\emph{Natural} misses are the decision-relevant objects that the draft misses on its own.
\method finds $41.5\%$, $51.1\%$, and $75.5\%$ of them, $1.6$--$2\times$ as many as the support alone, significantly more than whole-frame inference up to u512 on nuScenes and up to u640 on the two Argoverse datasets, and two to five times as many as the two object-level methods, which re-detect only what the draft detects.
The reference finds $43.3\%$, $55.5\%$, and $74.5\%$ (significantly more on Argoverse~2) with $3.7$--$4.6\times$ the pixels.
In the \emph{deleted} variant, the draft detections of all decision-relevant objects are removed before the loop, so that every such object must be found without a detection that points to it.
\method then recovers $93.3\%$, $95.9\%$, and $97.4\%$ of them, the support alone $41$--$58\%$, and the two object-level methods, which have no detection of these objects left to start from, $21$--$26\%$; whole-frame inference, which does not depend on the draft, recovers $96$--$99.6\%$.

\begin{table}[!ht]
\centering

\footnotesize
\setlength{\tabcolsep}{5pt}
\begin{tabular}{@{}lcccccc@{}}
\toprule
 & \multicolumn{2}{c}{nuScenes} & \multicolumn{2}{c}{Argoverse 2} & \multicolumn{2}{c}{Argoverse-HD} \\
\cmidrule(lr){2-3}\cmidrule(lr){4-5}\cmidrule(l){6-7}
Pipeline & Natural & Deleted & Natural & Deleted & Natural & Deleted \\
\midrule
\rf{u512} & \rf{51.3\sgu{}} & \rf{98.3\sgd{}} & \rf{62.9\sgu{}} & \rf{99.3\sgd{}} & \rf{65.5\sgu{}} & \rf{96.3} \\
\midrule
Uncertainty crops & 33.3\sgu{} & 46.9\sgu{} & 59.8\sgu{} & 62.4\sgu{} & 50.9\sgu{} & 72.5\sgu{} \\
\midrule
Support only (w/o field) & 44.4\sgu{} & 40.7\sgu{} & 48.5\sgu{} & 58.4\sgu{} & 58.9\sgu{} & 57.5\sgu{} \\
Field only (w/o support) & 65.0\sgu{} & 89.6\sgu{} & 78.3\sgu{} & 94.0\sgu{} & 93.0\sgu{} & 90.4\sgu{} \\
\midrule
\rowcolor{oursbg}\textbf{\method} & 84.6 & 93.3 & 88.4 & 95.9 & 97.2 & 97.4 \\
\bottomrule
\end{tabular}
\caption{\textbf{Recovery relative to the reference} (\%; test splits). Natural: decision-relevant objects that the draft misses and the reference detects; Deleted: decision-relevant objects whose draft detections are deleted before the loop. Uncertainty crops verify the least confident draft detections with the crop pixels of \method. Top block: whole-frame inference. \sgu{} / \sgd{}: \method significantly higher / lower.}
\label{S-tab:missed}
\end{table}

\noindent\textbf{Recovery relative to the reference.}
\Cref{S-tab:missed} restricts the natural misses to the objects that the reference detects, which separates what a region rule reaches from what the reference detector can find at all, and adds rules that do not appear in \cref{fig:recovery}.
\method then finds $84.6$--$97.2\%$ of these objects, u512 $51$--$66\%$, and uncertainty crops, which start from the draft detections, $33$--$60\%$.
When the draft detections of the relevant objects are deleted, uncertainty crops find $47$--$73\%$, and the field alone still finds $90$--$94\%$.

\begin{table}[!ht]
\centering

\scriptsize
\setlength{\tabcolsep}{2.4pt}
\begin{tabular}{@{}lcccccccccccc@{}}
\toprule
 & \multicolumn{3}{c}{Natural, vehicles} & \multicolumn{3}{c}{Natural, VRU} & \multicolumn{3}{c}{Deleted, vehicles} & \multicolumn{3}{c}{Deleted, VRU} \\
\cmidrule(lr){2-4}\cmidrule(lr){5-7}\cmidrule(lr){8-10}\cmidrule(l){11-13}
Pipeline & nuSc. & AV2 & A-HD & nuSc. & AV2 & A-HD & nuSc. & AV2 & A-HD & nuSc. & AV2 & A-HD \\
\midrule
\rf{u512} & \rf{28.4\sgu{}} & \rf{43.0\sgu{}} & \rf{60.4\sgu{}} & \rf{12.6\sgu{}} & \rf{20.8\sgu{}} & \rf{0.0} & \rf{98.4\sgd{}} & \rf{99.3\sgd{}} & \rf{96.2} & \rf{92.3} & \rf{95.1} & \rf{100.0} \\
\rf{u960 ($\full$)} & \rf{45.9} & \rf{58.0\sgd{}} & \rf{83.2} & \rf{37.9} & \rf{48.0} & \rf{20.9} & \rf{98.6\sgd{}} & \rf{99.6\sgd{}} & \rf{99.1\sgd{}} & \rf{100.0} & \rf{92.7} & \rf{100.0} \\
\midrule
AnA~\cite{ana} & 7.1\sgu{} & 31.1\sgu{} & 21.1\sgu{} & 24.1\sgu{} & 8.7\sgu{} & 1.0 & 25.4\sgu{} & 26.3\sgu{} & 26.2\sgu{} & 30.8\sgu{} & 4.9\sgu{} & 23.8\sgu{} \\
CF~\cite{cfimportance} & 6.0\sgu{} & 26.3\sgu{} & 18.1\sgu{} & 21.8\sgu{} & 8.7\sgu{} & 1.0 & 22.2\sgu{} & 22.3\sgu{} & 21.5\sgu{} & 30.8\sgu{} & 7.3\sgu{} & 22.6\sgu{} \\
\midrule
Support only (w/o field) & 19.1\sgu{} & 34.3\sgu{} & 53.2\sgu{} & 24.1\sgu{} & 7.5\sgu{} & 1.0 & 41.0\sgu{} & 58.8\sgu{} & 58.6\sgu{} & 23.1\sgu{} & 24.4\sgu{} & 23.8\sgu{} \\
\rowcolor{oursbg}\textbf{\method} & 42.6 & 53.1 & 83.8 & 39.1 & 44.5 & 23.6 & 93.2 & 96.3 & 97.3 & 100.0 & 68.3 & 98.8 \\
\bottomrule
\end{tabular}
\caption{\textbf{Recovery by class} of the decision-relevant objects that the draft lacks (\%; test splits; complements \cref{S-tab:recovery}). Natural and deleted as in \cref{S-tab:recovery}; vehicles: car, bus, truck; VRU: pedestrians and cyclists ($183$\,/\,$540$\,/\,$1{,}187$ and $87$\,/\,$173$\,/\,$191$ natural, $810$\,/\,$2{,}882$\,/\,$2{,}639$ and $13$\,/\,$41$\,/\,$84$ deleted objects). Top block: whole-frame inference. \sgu{}\,/\,\sgd{}: \method significantly higher\,/\,lower (paired bootstrap within the group).}
\label{S-tab:recovery_class}
\end{table}

\noindent\textbf{By class.}
\Cref{S-tab:recovery_class} splits the recovery into vehicles and vulnerable road users (VRU), \ie, pedestrians and cyclists.
Of the VRU that the draft misses, \method finds $24$--$45\%$, against $0$--$21\%$ for u512 and $21$--$48\%$ for the reference, and $1.6$ to $5$ times as many as the object-level rules on nuScenes and Argoverse~2; the support alone finds $1$--$24\%$, so these objects are reached mostly through the field.
On Argoverse-HD, the naturally missed VRU are few ($191$ objects in few sequences), and no difference among them is significant.
With their draft detections deleted, \method finds every one of the $13$ VRU on nuScenes and $98.8\%$ of the $84$ on Argoverse-HD.
AnA and counterfactual importance find $21$--$26\%$ of the vehicles and $5$--$31\%$ of the VRU in this variant.

\subsection{Coverage of the decisive object}
\label{S-sec:supp_coverage}

\begin{table}[!ht]
\centering

\footnotesize
\setlength{\tabcolsep}{6pt}
\begin{tabular}{@{}lcccccc@{}}
\toprule
 & \multicolumn{2}{c}{nuScenes} & \multicolumn{2}{c}{Argoverse 2} & \multicolumn{2}{c}{Argoverse-HD} \\
\cmidrule(lr){2-3}\cmidrule(lr){4-5}\cmidrule(l){6-7}
Decisive object & $\times1$ & $\times2$ & $\times1$ & $\times2$ & $\times1$ & $\times2$ \\
\midrule
In the draft & 84.8 & 71.6 & 90.2 & 84.4 & 70.4 & 66.7 \\
Box center in a verified or guard crop & 96.3 & 98.0 & 96.6 & 96.7 & 97.0 & 95.4 \\
In the final scene of \method & 92.3 & 88.7 & 96.5 & 94.6 & 96.3 & 92.5 \\
In the final scene, of those missing from the draft & 56.0 & 62.2 & 65.0 & 66.0 & 88.3 & 78.1 \\
\midrule
\emph{frames} & 928 & 2{,}481 & 2{,}854 & 5{,}758 & 3{,}593 & 6{,}976 \\
\bottomrule
\end{tabular}
\caption{\textbf{Coverage of the decisive object} on the frames that require braking (\%; test splits), at the recorded ego speed ($\times1$) and with the ego speed scaled by two ($\times2$). The decisive object is the annotated object with the largest demand.}
\label{S-tab:coverage}
\end{table}

Coverage, condition (iv) of Theorem~\ref{thm:exact}, concerns the objects of $\full$; \cref{S-tab:coverage} measures a related quantity on the annotations, namely where the decisive object of a frame that requires braking lies.
Its box center lies in a verified or guard crop on $96$--$97\%$ of these frames, and \method has it in its final scene on $92.3$--$96.5\%$, against $70.4$--$90.2\%$ for the draft; of the decisive objects that are missing from the draft, \method finds $56$--$88\%$.
With the ego speed scaled by two, the braking range and the number of frames that require braking grow, and the decisive object still lies inside a verified or guard crop on $95$--$98\%$ of them; the field reaches farther (\cref{S-tab:fieldadapt}).

\subsection{The field follows the action}
\label{S-sec:supp_fieldadapt}

\begin{table}[!ht]
\centering

\footnotesize
\setlength{\tabcolsep}{10pt}
\begin{tabular}{@{}lccc@{}}
\toprule
 & nuScenes & Argoverse 2 & Argoverse-HD \\
\midrule
\multicolumn{4}{@{}l}{\emph{Frames with a requested field, by draft action (\%)}} \\
\textsc{keep} & 59.8 & 59.7 & 65.6 \\
\textsc{decelerate} & 61.8 & 65.4 & 73.2 \\
\textsc{hard-brake} & 0.0 & 0.0 & 0.0 \\
\midrule
\multicolumn{4}{@{}l}{\emph{Pixels of \method (\% of native) as the ego speed is scaled}} \\
$\times1$ & 8.4 & 5.9 & 6.9 \\
$\times1.5$ & 9.4 & 6.5 & 7.3 \\
$\times2$ & 9.8 & 6.6 & 7.4 \\
\midrule
\multicolumn{4}{@{}l}{\emph{Farthest sensitive range of the field (m, mean over the frames with a field)}} \\
$\times1$ & 38 & 40 & 42 \\
$\times1.5$ & 50 & 48 & 53 \\
$\times2$ & 57 & 53 & 61 \\
\bottomrule
\end{tabular}
\caption{\textbf{The field depends on the action and the ego speed} (test splits). At \textsc{hard-brake}, no stronger action exists, so the field is empty and only the support is verified.}
\label{S-tab:fieldadapt}
\end{table}

The field is a function of the current action and of the ego speed (\cref{S-tab:fieldadapt}), which is why it needs few pixels.
At \textsc{keep}, it is requested on $60$--$66\%$ of the frames.
At \textsc{decelerate}, it contains only the states that would raise the action to \textsc{hard-brake}, a shorter corridor, and is requested on $62$--$73\%$ of the frames.
At \textsc{hard-brake}, no stronger action exists, the field is empty, and only the support is verified.
When the ego speed is scaled, the farthest sensitive range of the field grows from about $40$\,m to $53$--$61$\,m, and the pixels of \method grow from $5.9$--$8.4\%$ to $6.6$--$9.8\%$ of the native frame.
The cost thus follows the extent of the field: for the selector on NAVSIM, whose field spans all $15$ candidates, \method uses two thirds of the pixels of the reference (\cref{S-sec:consumers_outside}), against $22$--$27\%$ for the braking controller.

\subsection{Random and oracle crop placement}
\label{S-sec:supp_random}

\begin{table}[!ht]
\centering

\scriptsize
\setlength{\tabcolsep}{2.4pt}
\begin{tabular}{@{}lccccccccccccccc@{}}
\toprule
 & \multicolumn{5}{c}{nuScenes} & \multicolumn{5}{c}{Argoverse 2} & \multicolumn{5}{c}{Argoverse-HD} \\
\cmidrule(lr){2-6}\cmidrule(lr){7-11}\cmidrule(l){12-16}
Regions & px & Agr. & Rep. & F1 & Rec. & px & Agr. & Rep. & F1 & Rec. & px & Agr. & Rep. & F1 & Rec. \\
\midrule
\rf{u512} & 10.2 & 97.7\sgu{} & 79.6\sgu{} & 62.2\sgu{} & 23.3\sgu{} & 6.7 & 96.4\sgu{} & 84.0\sgu{} & 71.3\sgu{} & 37.6\sgu{} & 7.1 & 94.5\sgu{} & 82.6\sgu{} & 39.8\sgu{} & 52.0\sgu{} \\
\midrule
Random crops, whole image & 9.2 & 94.6\sgu{} & 29.3\sgu{} & 44.9\sgu{} & 14.3\sgu{} & 6.3 & 88.8\sgu{} & 26.7\sgu{} & 43.7\sgu{} & 14.2\sgu{} & 8.0 & 83.9\sgu{} & 26.3\sgu{} & 17.4\sgu{} & 17.8\sgu{} \\
Random crops, lower half & 8.8 & 94.5\sgu{} & 27.5\sgu{} & 43.1\sgu{} & 13.7\sgu{} & 6.2 & 87.8\sgu{} & 19.8\sgu{} & 43.5\sgu{} & 8.1\sgu{} & 7.8 & 82.9\sgu{} & 21.3\sgu{} & 18.1\sgu{} & 10.8\sgu{} \\
Crops around relevant objects & 5.1 & 94.7\sgu{} & 29.6\sgu{} & 41.3\sgu{} & 43.0 & 3.5 & 90.5\sgu{} & 37.5\sgu{} & 39.9\sgu{} & 54.4\sgd{} & 3.8 & 87.5\sgu{} & 43.2\sgu{} & 14.7\sgu{} & 75.1 \\
\midrule
\rowcolor{oursbg}\textbf{\method} & 8.4 & 98.7 & 86.9 & 69.0 & 41.5 & 5.9 & 98.2 & 91.7 & 77.5 & 51.1 & 6.9 & 97.5 & 92.0 & 46.7 & 75.5 \\
\bottomrule
\end{tabular}
\caption{\textbf{Where the pixels go matters} (\%; test splits). Random crops: the crops of \method on each frame, with their sizes, moved to uniformly random positions in the whole image or its lower half (mean of three seeds; seeds agree within $1.5$ points); crops around relevant objects: one crop (box enlarged twice) around every annotated decision-relevant object. Each row runs one pass on its crops with the truncation guard, the same-scale history, and the lift of \method. Rec.: share of the decision-relevant objects missed by the draft that the final scene contains, as in \cref{S-tab:recovery}. \sgu{}\,/\,\sgd{}: \method significantly higher\,/\,lower.}
\label{S-tab:random}
\end{table}

\Cref{S-tab:random} keeps the crops of \method on every frame, with their sizes and hence about their pixels, and moves them to random positions.
The repair rate falls from $87$--$92\%$ to $20$--$29\%$, the recovery of the objects that the draft misses from $41.5$--$75.5\%$ to $8$--$18\%$, and the F1 to the level of the draft, on all three datasets and for every seed: the gain of \method comes from where the field and the support put the pixels, not from their number.
Crops around the annotated decision-relevant objects, a region rule that knows the answer, recover about as many missed objects as \method with fewer pixels, but agree with $\policy(\full)$ on only $87.5$--$94.7\%$ of the frames and have an F1 below the draft's: they do not target the spurious or mislocalized draft detections on which a braking action rests, which is the role of the support.

\subsection{Other crop rules}
\label{S-sec:supp_unc}

\begin{table}[!ht]
\centering

\footnotesize
\setlength{\tabcolsep}{3.4pt}
\begin{tabular}{@{}lcccccccccccc@{}}
\toprule
 & \multicolumn{4}{c}{nuScenes} & \multicolumn{4}{c}{Argoverse 2} & \multicolumn{4}{c}{Argoverse-HD} \\
\cmidrule(lr){2-5}\cmidrule(lr){6-9}\cmidrule(l){10-13}
Pipeline & px & Agr. & Rep. & F1 & px & Agr. & Rep. & F1 & px & Agr. & Rep. & F1 \\
\midrule
\rf{u512} & \rf{10.2} & \rf{97.7\sgu{}} & \rf{79.6\sgu{}} & \rf{62.2\sgu{}} & \rf{6.7} & \rf{96.4\sgu{}} & \rf{84.0\sgu{}} & \rf{71.3\sgu{}} & \rf{7.1} & \rf{94.5\sgu{}} & \rf{82.6\sgu{}} & \rf{39.8\sgu{}} \\
\rf{u960 ($\full$)} & \rf{38.4} & \rf{100} & \rf{100} & \rf{65.0\sgu{}} & \rf{23.2} & \rf{100} & \rf{100} & \rf{75.9} & \rf{25.3} & \rf{100} & \rf{100} & \rf{40.9} \\
\midrule
Uncertainty crops & 9.6 & 97.7\sgu{} & 73.6\sgu{} & 60.7\sgu{} & 8.7 & 96.6\sgu{} & 81.2\sgu{} & 69.2\sgu{} & 11.5 & 94.1\sgu{} & 75.5\sgu{} & 34.0\sgu{} \\
\method, crops at native scale & 13.6 & 97.3\sgu{} & 77.3\sgu{} & 66.2 & 14.5 & 95.7\sgu{} & 84.3\sgu{} & 77.4 & 16.0 & 94.0\sgu{} & 83.3\sgu{} & 46.2 \\
\midrule
\rowcolor{oursbg}\textbf{\method} & 8.4 & 98.7 & 86.9 & 69.0 & 5.9 & 98.2 & 91.7 & 77.5 & 6.9 & 97.5 & 92.0 & 46.7 \\
\bottomrule
\end{tabular}
\caption{\textbf{Other crop rules} (\%; test splits). Uncertainty crops: the least confident draft detections, with context, selected with the crop pixels of \method on the frame (their guard re-detections add to the total), through the same crops, guard, and lift; crops at native scale: the regions of \method detected at the native resolution instead of the verification resolution, without the previous-frame re-detection. Columns as in \cref{tab:fid,tab:main}. Top block: whole-frame inference. \sgu{} / \sgd{}: \method significantly higher / lower. Against native-resolution inference instead of $\policy(\full)$, the crops at native scale agree on $98.4\%$, $97.1\%$, and $96.8\%$ of the frames, significantly more often than u960 ($97.7\%$, $96.0\%$, and $93.7\%$).}
\label{S-tab:unc}
\end{table}

\Cref{S-tab:unc} compares the regions of \method with two other ways to spend crops.
Uncertainty crops, which verify the least confident draft detections with the crop pixels of \method, have a significantly lower agreement, repair rate, and F1 than \method on all three datasets, and they repair fewer disagreements and have a lower F1 than u512.
Detecting the regions of \method at the native resolution instead of the verification resolution costs $1.6$--$2.5\times$ the pixels and lowers the agreement with $\policy(\full)$ and the repair rate, since the crops then differ from the reference in scale and not only in context; its F1 is close to that of \method, and against native-resolution inference it is the more faithful pipeline (\cref{S-sec:supp_yolo}).

\subsection{Parameters of the field}
\label{S-sec:supp_grid}

\begin{table}[!ht]
\centering

\footnotesize
\setlength{\tabcolsep}{3.4pt}
\begin{tabular}{@{}lcccccccccccc@{}}
\toprule
 & \multicolumn{4}{c}{nuScenes} & \multicolumn{4}{c}{Argoverse 2} & \multicolumn{4}{c}{Argoverse-HD} \\
\cmidrule(lr){2-5}\cmidrule(lr){6-9}\cmidrule(l){10-13}
Field & px & Agr. & F1 & Diff. & px & Agr. & F1 & Diff. & px & Agr. & F1 & Diff. \\
\midrule
\textbf{Reported} (vehicle, $3$\,m/s) & 1 & 98.75 & 69.0 & 0 & 1 & 98.25 & 77.5 & 0 & 1 & 97.52 & 46.7 & 0 \\
\midrule
Probe without closing speed & $\times$1.00 & 98.75 & 69.0 & 0 & $\times$1.00 & 98.25 & 77.5 & 0 & $\times$1.00 & 97.52 & 46.7 & 0 \\
Probe closing at $10$\,m/s & $\times$1.20 & 98.92\sgd{} & 68.3 & 24 & $\times$1.25 & 98.51\sgd{} & 76.4 & 59 & $\times$1.15 & 97.61 & 44.0 & 40 \\
Lateral spacing $\times0.5$ & $\times$1.00 & 98.75 & 69.0 & 0 & $\times$1.00 & 98.25 & 77.5 & 0 & $\times$1.00 & 97.52 & 46.7 & 0 \\
Lateral spacing $\times2$ & $\times$0.85 & 98.71 & 68.3 & 29 & $\times$0.89 & 98.20 & 77.1 & 42 & $\times$0.87 & 97.36 & 45.6 & 72 \\
Range spacing $\times0.5$ & $\times$1.00 & 98.75 & 69.0 & 0 & $\times$1.00 & 98.25 & 77.7 & 2 & $\times$1.00 & 97.52 & 46.7 & 0 \\
Range spacing $\times2$ & $\times$1.00 & 98.75 & 69.0 & 0 & $\times$1.00 & 98.25 & 77.7 & 4 & $\times$1.00 & 97.52 & 46.7 & 0 \\
Pedestrian-sized probe & $\times$0.80 & 98.50\sgu{} & 65.5\sgu{} & 55 & $\times$0.82 & 97.76\sgu{} & 77.1 & 108 & $\times$0.84 & 97.34\sgu{} & 45.0 & 115 \\
Truck-sized probe & $\times$1.30 & 98.98\sgd{} & 67.1 & 49 & $\times$1.18 & 98.47\sgd{} & 77.1 & 62 & $\times$1.22 & 97.83\sgd{} & 47.4 & 103 \\
\bottomrule
\end{tabular}
\caption{\textbf{Sensitivity to the parameters of the field} (test splits). px: detector-input pixels relative to the reported configuration; Agr.: agreement with $\policy(\full)$ (\%); F1 of the braking episodes (\%); Diff.: frames on which the action differs from that of the reported configuration. \sgu{} / \sgd{}: the reported configuration significantly higher / lower.}
\label{S-tab:grid}
\end{table}

\Cref{S-tab:grid} varies the probe and the grid of the field around the reported configuration.
The action is insensitive to the spacing of the grid: halving the lateral spacing or halving or doubling the range spacing changes the action on at most four frames of a dataset, and a static probe gives exactly the actions and pixels of the probe closing at $3$\,m/s: for the braking controller, a probe closing at $3$\,m/s reaches $60\%$ of the lowest threshold only within $9.5$\,m ($9/(2(r-r_0))\ge0.6$\,m/s$^2$), nearer than the first range of the grid ($12.5$\,m).
A faster or larger probe enlarges the footprint: a probe closing at $10$\,m/s or a truck-sized probe raises the agreement by $0.1$--$0.3$ points at $1.15$--$1.30\times$ the pixels, without a significant change of the F1, and a pedestrian-sized probe lowers both. Doubling the lateral spacing saves $11$--$15\%$ of the pixels and changes $29$--$72$ actions without a significant change of either measure.
The probe therefore sets an operating point on the trade-off between pixels and fidelity, and the grid spacing is not critical.

With the reported configuration as the incumbent, none of these variants meets the adoption rule of \cref{S-sec:supp_data}: on neither development split is its effect on the F1 significantly positive.

\subsection{Other planners}
\label{S-sec:supp_planners}

\begin{table}[!ht]
\centering

\footnotesize
\setlength{\tabcolsep}{3.4pt}
\begin{tabular}{@{}lcccccccccccc@{}}
\toprule
 & \multicolumn{4}{c}{nuScenes} & \multicolumn{4}{c}{Argoverse 2} & \multicolumn{4}{c}{Argoverse-HD} \\
\cmidrule(lr){2-5}\cmidrule(lr){6-9}\cmidrule(l){10-13}
Regions & px & P & R & F1 & px & P & R & F1 & px & P & R & F1 \\
\midrule
\multicolumn{13}{@{}l}{\emph{Curved-path braking controller}} \\
\rf{u320 (draft)} & \rf{4.3} & \rf{29.4\sgu{}} & \rf{63.5} & \rf{40.2\sgu{}} & \rf{2.6} & \rf{34.0\sgu{}} & \rf{78.9\sgd{}} & \rf{47.6\sgu{}} & \rf{3.1} & \rf{13.2\sgu{}} & \rf{64.1\sgd{}} & \rf{21.9\sgu{}} \\
\rf{u512} & \rf{10.2} & \rf{50.8\sgu{}} & \rf{61.0} & \rf{55.4} & \rf{6.7} & \rf{62.3\sgu{}} & \rf{77.7\sgd{}} & \rf{69.2\sgu{}} & \rf{7.1} & \rf{42.1\sgu{}} & \rf{51.5} & \rf{46.3} \\
\rf{u960 ($\full$)} & \rf{38.4} & \rf{54.8\sgu{}} & \rf{64.8\sgd{}} & \rf{59.4} & \rf{23.2} & \rf{65.9\sgu{}} & \rf{78.9\sgd{}} & \rf{71.8} & \rf{25.3} & \rf{43.5} & \rf{52.4} & \rf{47.6} \\
\rf{native} & \rf{103.1} & \rf{58.2} & \rf{64.8\sgd{}} & \rf{61.3} & \rf{101.2} & \rf{65.8\sgu{}} & \rf{79.4\sgd{}} & \rf{71.9} & \rf{101.3} & \rf{43.8} & \rf{55.3\sgd{}} & \rf{48.9} \\
\method, braking regions & 8.4 & 47.2\sgu{} & 64.8\sgd{} & 54.6 & 5.9 & 55.5\sgu{} & 79.8\sgd{} & 65.4\sgu{} & 6.9 & 29.6\sgu{} & 54.4 & 38.4\sgu{} \\
\rowcolor{oursbg}\textbf{\method}, own regions & 7.6 & 61.4 & 56.0 & 58.6 & 5.3 & 73.8 & 72.9 & 73.3 & 6.3 & 52.0 & 49.5 & 50.7 \\
\midrule
\multicolumn{13}{@{}l}{\emph{Time-to-collision warning}} \\
\rf{u320 (draft)} & \rf{4.3} & \rf{19.8\sgu{}} & \rf{43.5} & \rf{27.2\sgu{}} & \rf{2.6} & \rf{15.9\sgu{}} & \rf{76.8\sgd{}} & \rf{26.4\sgu{}} & \rf{3.1} & \rf{3.6\sgu{}} & \rf{87.2\sgd{}} & \rf{7.0\sgu{}} \\
\rf{u512} & \rf{10.2} & \rf{41.3\sgu{}} & \rf{35.4} & \rf{38.1\sgu{}} & \rf{6.7} & \rf{50.5\sgu{}} & \rf{69.0} & \rf{58.3} & \rf{7.1} & \rf{15.2} & \rf{38.5} & \rf{21.7} \\
\rf{u960 ($\full$)} & \rf{38.4} & \rf{52.3} & \rf{38.1} & \rf{44.1} & \rf{23.2} & \rf{56.0} & \rf{66.5} & \rf{60.8} & \rf{25.3} & \rf{13.6} & \rf{30.8} & \rf{18.9} \\
\rf{native} & \rf{103.1} & \rf{51.9} & \rf{38.1} & \rf{43.9} & \rf{101.2} & \rf{56.4} & \rf{65.2} & \rf{60.5} & \rf{101.3} & \rf{10.4} & \rf{25.6} & \rf{14.8} \\
\method, braking regions & 8.4 & 38.4\sgu{} & 38.1 & 38.2\sgu{} & 5.9 & 40.6\sgu{} & 62.6 & 49.2\sgu{} & 6.9 & 4.3\sgu{} & 35.9 & 7.6\sgu{} \\
\rowcolor{oursbg}\textbf{\method}, own regions & 8.5 & 58.9 & 36.1 & 44.7 & 5.8 & 58.7 & 65.2 & 61.8 & 7.0 & 14.5 & 30.8 & 19.7 \\
\bottomrule
\end{tabular}
\caption{\textbf{Other planners: pixels, precision, recall, and F1 of the episodes} of each planner (braking episodes of the controller, warning episodes of the warning; \%; test splits; complements \cref{tab:planners}). Braking regions: field and support of the straight-path braking controller; own regions: of the planner in use. In each block, the first rows are whole-frame inference. \sgu{} / \sgd{}: \method with its own regions significantly higher / lower.}
\label{S-tab:planners_full}
\end{table}

\Cref{S-tab:planners_full} gives the pixels, precision, and recall behind \cref{tab:planners}.

\noindent\textbf{Planners.}
The \emph{curved-path braking controller} is the controller of \cref{S-sec:supp_metric} with the ego corridor bent along the path of constant curvature given by the ego yaw rate.
The \emph{time-to-collision warning} is a second max-composable planner: its per-object demand is the inverse time to collision of an object in the ego corridor, and it warns at a time to collision of at most $4$\,s and brakes at most $2$\,s.
Each planner defines its own required action, episodes, field, and support through Corollary~\ref{cor:maxcomp}; the crops, the guard, and the lift are those of \method.
Three settings differ from the braking configuration; none was tuned for these planners.
The field of the curved-path controller is evaluated on its own grid at the true thresholds, without the margin.
The warning runs in the generic loop for planners other than the braking controller, which computes the footprint with a pinhole model from the camera height and intrinsics alone and does not re-request the crops of objects verified in the previous frame; its probe is static, so that a probe at range $r$ has the time to collision $r/v$.
The bottleneck selector on NAVSIM is specified in \cref{S-sec:consumers_outside}.

\noindent\textbf{Results.}
With the regions of the planner in use, \method has the highest precision in five of the six columns and an F1 that does not differ significantly from that of whole-frame inference at the reference resolution, with $3.6$--$5.1\times$ fewer pixels.
\Cref{S-tab:planners_fid} adds the fidelity to the reference under each planner.
Under the curved-path controller, \method with its own regions agrees with $\policy(\full)$ on $97.1$--$98.5\%$ of the frames, significantly more often than u512 and than \method with the braking regions on every dataset, with fewer pixels than both.
Under the warning, it agrees significantly more often than with the braking regions on nuScenes and Argoverse~2.

\begin{table}[!ht]
\centering

\footnotesize
\setlength{\tabcolsep}{5pt}
\begin{tabular}{@{}lccccccccc@{}}
\toprule
 & \multicolumn{3}{c}{nuScenes} & \multicolumn{3}{c}{Argoverse 2} & \multicolumn{3}{c}{Argoverse-HD} \\
\cmidrule(lr){2-4}\cmidrule(lr){5-7}\cmidrule(l){8-10}
Regions & px & Agr. & Rep. & px & Agr. & Rep. & px & Agr. & Rep. \\
\midrule
\multicolumn{10}{@{}l}{\emph{Curved-path braking controller}} \\
\rf{u512} & \rf{10.2} & \rf{97.6\sgu{}} & \rf{81.6} & \rf{6.7} & \rf{96.2\sgu{}} & \rf{82.9\sgu{}} & \rf{7.1} & \rf{94.7\sgu{}} & \rf{83.2\sgu{}} \\
\method, braking regions & 8.4 & 97.1\sgu{} & 71.9\sgu{} & 5.9 & 94.6\sgu{} & 75.5\sgu{} & 6.9 & 94.7\sgu{} & 82.4\sgu{} \\
\rowcolor{oursbg}\textbf{\method}, own regions & 7.6 & 98.5 & 85.4 & 5.3 & 97.5 & 87.8 & 6.3 & 97.1 & 90.2 \\
\midrule
\multicolumn{10}{@{}l}{\emph{Time-to-collision warning}} \\
\rf{u512} & \rf{10.2} & \rf{98.3} & \rf{86.8\sgd{}} & \rf{6.7} & \rf{97.2} & \rf{89.4} & \rf{7.1} & \rf{96.3} & \rf{91.6} \\
\method, braking regions & 8.4 & 97.4\sgu{} & 67.2\sgu{} & 5.9 & 96.4\sgu{} & 81.5\sgu{} & 6.9 & 88.4 & 61.7 \\
\rowcolor{oursbg}\textbf{\method}, own regions & 8.5 & 98.1 & 80.6 & 5.8 & 97.4 & 89.2 & 7.0 & 89.9 & 68.6 \\
\bottomrule
\end{tabular}
\caption{\textbf{Other planners: fidelity} to $\policy(\full)$ under the planner in use (\%; test splits; complements \cref{S-tab:planners_full}). px: detector-input pixels (\% of native); Agr., Rep.: agreement and repair rate as in \cref{tab:fid}, with the draft and the reference under the same planner. \sgu{} / \sgd{}: \method with its own regions significantly higher / lower.}
\label{S-tab:planners_fid}
\end{table}

Reusing the regions of the straight-path braking controller instead loses precision, by $14$--$22$ points under the curved-path controller and by $10$--$21$ points under the warning: the regions then follow the wrong planner, as they do for the selector on NAVSIM (\cref{S-sec:consumers_outside}).

\section{Controlled closed loop}
\label{S-sec:supp_embedded}
\label{S-sec:supp_aeb}

This section specifies the controlled closed loop of \cref{sec:exp_planning} and reports all of its metrics.
To isolate the effect of where the detector runs on longitudinal control, the closed loop of \cref{sec:exp_planning} runs in CARLA~0.9.15~\cite{carla} on the $155$ Bench2Drive~\cite{bench2drive} evaluation routes whose scenarios require braking for vehicles, pedestrians, or bicycles and need no lateral maneuver around an actor, with three traffic seeds per route.

\noindent\textbf{Loop.}
A route follower controls the steering and the target speed without seeing any actor, and the braking controller of \cref{S-sec:supp_metric} acts on the scene that each perception pipeline estimates, with a hold at standstill.
All pipelines see the same front camera ($1600\times900$ pixels, $70^{\circ}$ horizontal field of view) and use the same YOLOv8s and the geometric lift with a time to collision from the growth of the box over $0.5$\,s; the looming estimate and the range refiner, which were trained on real images, are not used in simulation. The action of a frame is applied in the same simulation step.

\noindent\textbf{Metrics.}
Required braking episodes are defined on the privileged states of the simulator as in \cref{S-sec:supp_data}; an episode is missed if the pipeline does not brake during it, and a hard-braking onset is unnecessary if these states do not require \textsc{hard-brake}.
A collision is \emph{preventable} unless it is a side or rear impact or the braking demand on the privileged states arises less than $v/(6\,\mathrm{m/s^2})+0.4$\,s before the contact, where $v$ is the ego speed; most collisions on these routes are not preventable by longitudinal control.
Differences are tested with a paired bootstrap over routes, with the seeds averaged per route.

\begin{table}[!ht]
\centering

\footnotesize
\setlength{\tabcolsep}{6pt}
\begin{tabular}{@{}lcccc@{}}
\toprule
Pipeline & Miss. & Hard & Prev. & px \\
\midrule
\rf{Privileged} & \rf{0.00\sgd{}} & \rf{0.00\sgd{}} & \rf{0.2} & \na \\
\rf{u320 (draft)} & \rf{0.44\sgu{}} & \rf{4.25\sgu{}} & \rf{1.5} & \rf{4.3} \\
\rf{u512} & \rf{0.35\sgu{}} & \rf{3.90\sgu{}} & \rf{1.7\sgu{}} & \rf{10.2} \\
\rf{u960 ($\full$)} & \rf{0.33} & \rf{3.97\sgu{}} & \rf{1.1} & \rf{38.4} \\
\rf{native} & \rf{0.32} & \rf{4.77\sgu{}} & \rf{1.3} & \rf{103.1} \\
\rowcolor{oursbg}\textbf{\method} & 0.28 & 3.07 & 0.4 & 8.1 \\
\bottomrule
\end{tabular}
\caption{\textbf{Controlled closed loop, all metrics} ($155$ routes $\times$ three seeds; complements \cref{tab:closedloop}). Per rollout: Miss.: missed required braking episodes; Hard: unnecessary hard-braking onsets; Prev.: rollouts with a preventable collision (\%); px: detector-input pixels (\% of native). Lower is better. Privileged: the controller on the simulator states. \sgu{} / \sgd{}: \method significantly better / worse (paired route bootstrap).}
\label{S-tab:aeb_full}
\end{table}

\noindent\textbf{Results.}
\cref{tab:closedloop} and \cref{S-tab:aeb_full} report the rollouts of the $155$ routes with three seeds each.
\method has the fewest unnecessary hard-braking onsets, significantly fewer than every perception pipeline, and the fewest missed required episodes, significantly fewer than u320 and u512.
Preventable collisions are rare ($0.4$--$1.7\%$ of the rollouts): \method has $2$ in $465$ rollouts, against $7$ for the draft, $8$ for u512, $5$ for the reference, and $6$ for native-resolution inference.
The experiment characterizes braking behavior, not driving safety.

\section{Qualitative results}
\label{S-sec:supp_qual}

\subsection{Selection rule and examples}
\label{S-sec:supp_qualitative}

The frames of \cref{fig:qual} and of \cref{S-fig:qual_n} are selected by a fixed rule on the stored test results, with categories defined by the actions:
\begin{itemize}[leftmargin=14pt]
\item \emph{unnecessary braking removed}: the required action is \textsc{keep}, the draft brakes, and \method and the reference keep;
\item \emph{unseen object recovered}: a decision-relevant object is missed by the draft and by the support-only ablation and found by the reference and by \method, which takes the required action;
\end{itemize}
For each category and dataset, the first qualifying frame in time order is shown; none is chosen by inspection.
\cref{fig:qual}(b) shows the first category on nuScenes, and \cref{S-fig:qual_n} the second on nuScenes and both on Argoverse~2.

\cref{fig:qual}(a) is meant to show the field alone reaching an object, which the second category does not ensure: there, the object can also lie in the crop of an object verified in the previous frame or of another detection's support.
Its frame therefore meets a stricter rule, applied to every nuScenes and Argoverse~2 test frame on which the draft keeps the speed while braking is required and \method takes the required action: a decision-relevant object that no draft box overlaps lies in the field footprint of the first round, in no crop of an object verified in the previous frame and in no first-round support crop, was not verified in the previous frame, and is found by \method; any other crop that contains it is built around it after it is found.
\cref{fig:qual}(a) shows the qualifying daylight frame with the object beyond $25$\,m in which \method finds every decision-relevant object that the draft misses and the support alone would keep the speed (Argoverse~2, a pedestrian crossing $37$\,m ahead).

\cref{fig:method} is meant to show the support and the field on one daylight frame.
Its frame meets the following rule: in daylight, the draft takes \textsc{decelerate}, the first round verifies only the support crop and lowers the action to \textsc{keep}, the second round verifies the field of \textsc{keep}, the two crops overlap with an IoU below $0.3$ and each covers at least $3\%$ of the image, and \method returns the required \textsc{keep} within two rounds.
\cref{fig:method} shows one of the qualifying nuScenes test frames, chosen for legibility (a single vehicle ahead and two clearly separated crops); its top view is cropped to the near range, where the field of \textsc{keep} lies at this ego speed.

\definecolor{qB}{HTML}{1F77B4}\definecolor{qO}{HTML}{FF7F0E}\definecolor{qG}{HTML}{7F7F7F}\definecolor{qR}{HTML}{D62728}\definecolor{qN}{HTML}{2CA02C}
\newcommand{\qualhead}{\noindent\makebox[0.318\textwidth]{\footnotesize (a) draft}\hfill\makebox[0.318\textwidth]{\footnotesize (b) verified crops}\hfill\makebox[0.318\textwidth]{\footnotesize (c) final scene}\par\vspace{2pt}}
\newcommand{\qualrow}[6]{
\noindent{\small\textbf{(#1)~#2}}\par\vspace{1pt}
\noindent\includegraphics[width=0.318\textwidth]{#3}\hfill\includegraphics[width=0.318\textwidth]{#4}\hfill\includegraphics[width=0.318\textwidth]{#5}\par\vspace{2pt}
\noindent{\small #6}\par\vspace{6pt}}
\newcommand{\qlg}[2]{\tikz[baseline=-0.6ex]\draw[#1, line width=1.1pt] (0,-0.08) rectangle (0.30,0.10);\,{\footnotesize #2}\quad}
\newcommand{\qualkey}[1]{\noindent\parbox{\textwidth}{\centering\qlg{qB}{draft box in (a), support crop in (b)}\qlg{qB, dash pattern=on 2pt off 1.3pt}{replaced draft box in (a), crop of an object verified in the previous frame in (b)}\\[2pt]
\qlg{qO}{field footprint}\qlg{qG}{guard crop}\qlg{qR}{verified detection}\qlg{qR, dash pattern=on 2pt off 1.3pt}{missed by the draft}\qlg{qN}{annotation}\qlg{qN, dash pattern=on 2pt off 1.3pt}{relevant annotation}#1}\par\vspace{2pt}}

\begin{figure}[t]
\qualhead
\qualrow{1}{Unseen object recovered (nuScenes)}{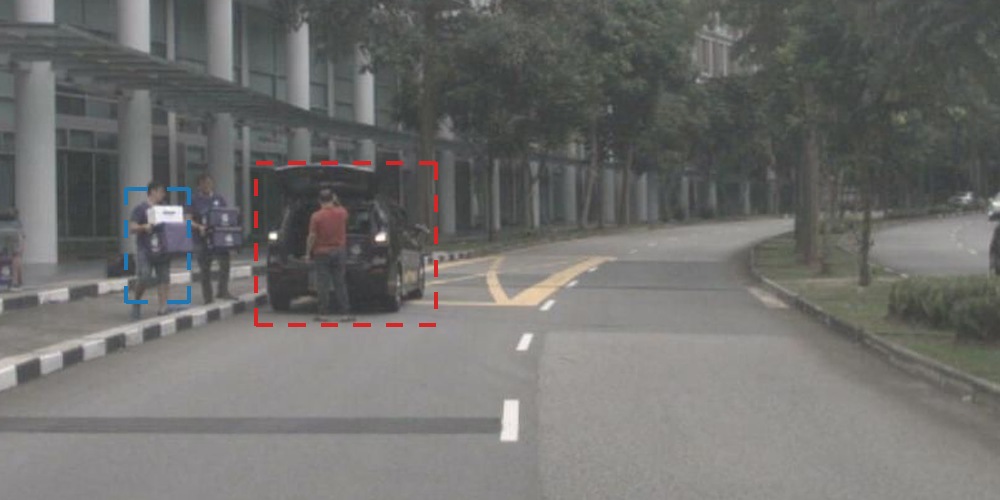}{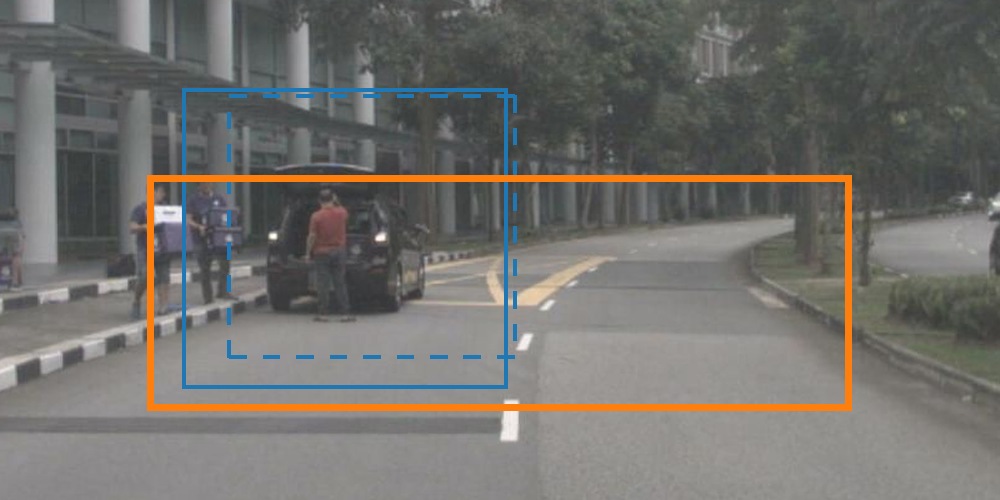}{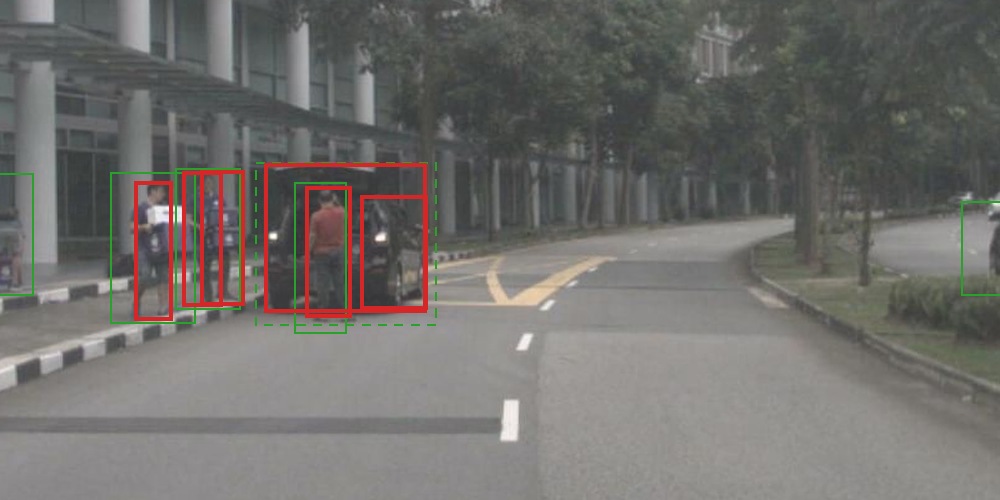}{The draft misses the car parked partly in the lane, with a person at its open trunk, and keeps the speed. The field footprint covers the car, as does the crop of an object verified in the previous frame (dashed blue); the reference detector finds it in the crop, and \method decelerates as required.}
\qualrow{2}{Unnecessary braking removed (Argoverse~2)}{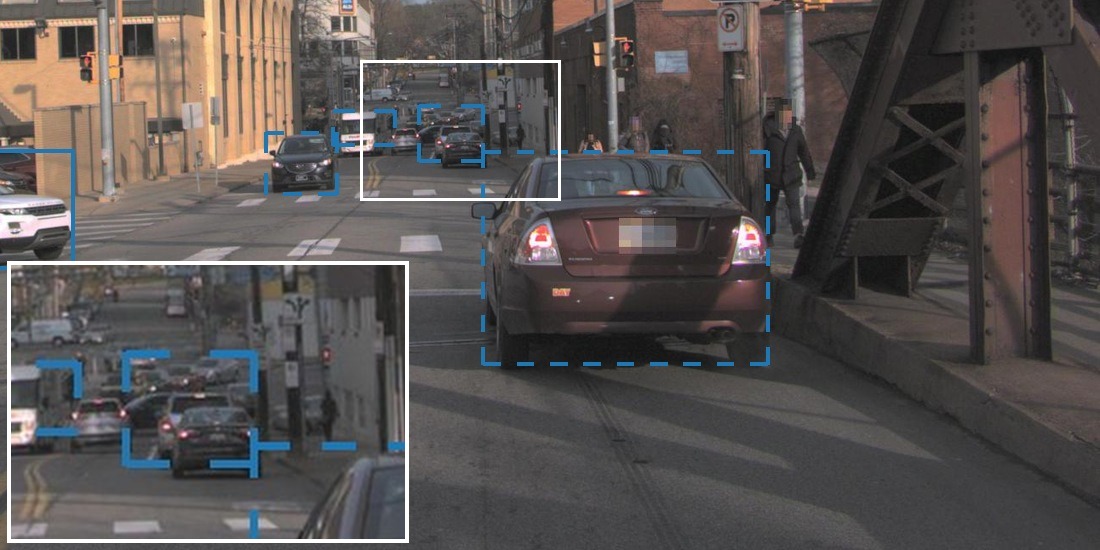}{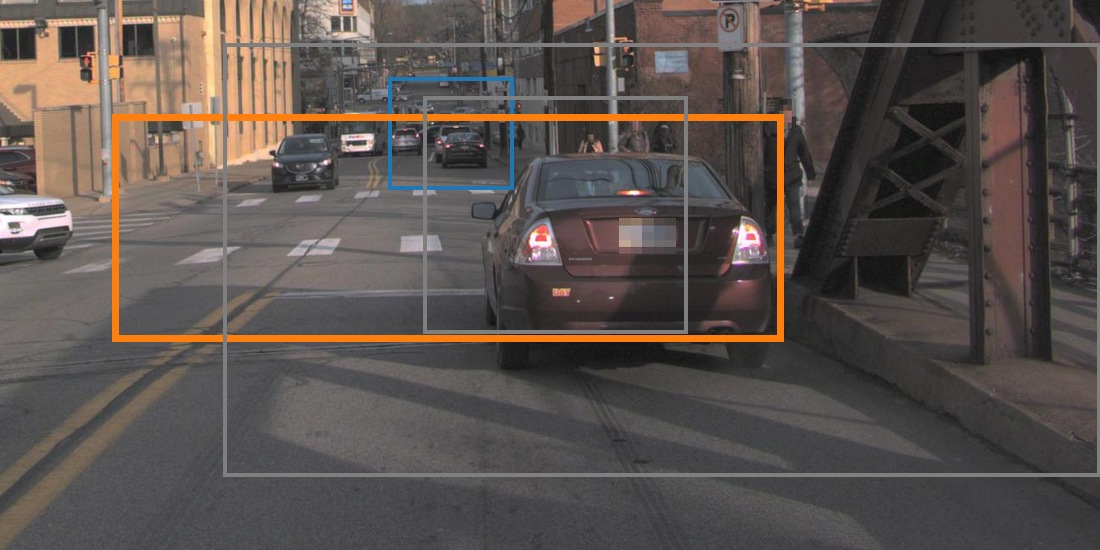}{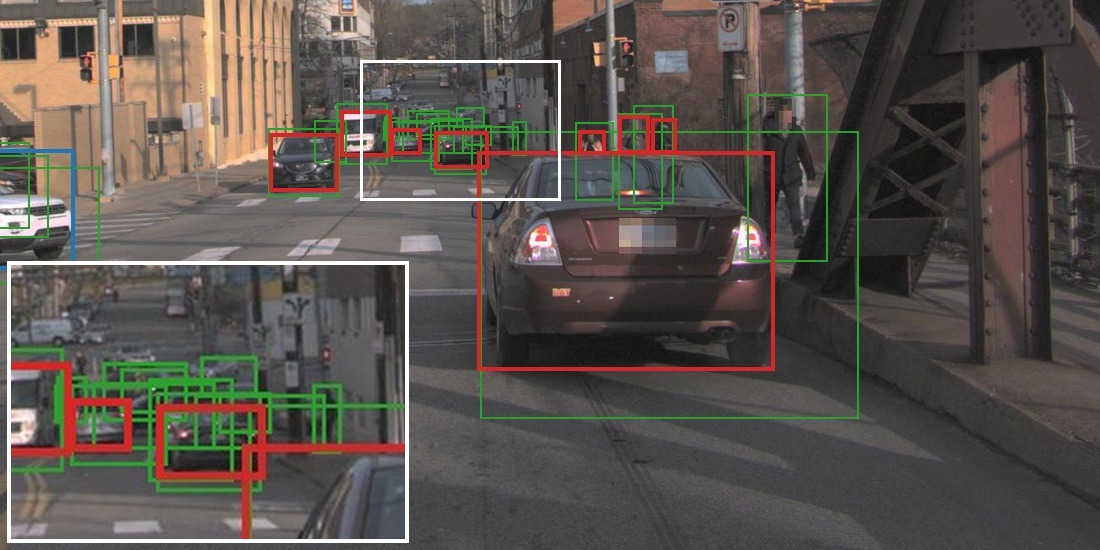}{The draft brakes hard for an oversized box around distant vehicles (zoomed). Verifying its support (blue crop) replaces it by the boxes of the individual vehicles, the action falls to \textsc{keep}, and the field of \textsc{keep} requests its footprint in a second round, in which nothing relevant is found: \textsc{keep}, as required.}
\qualrow{3}{Unseen object recovered (Argoverse~2)}{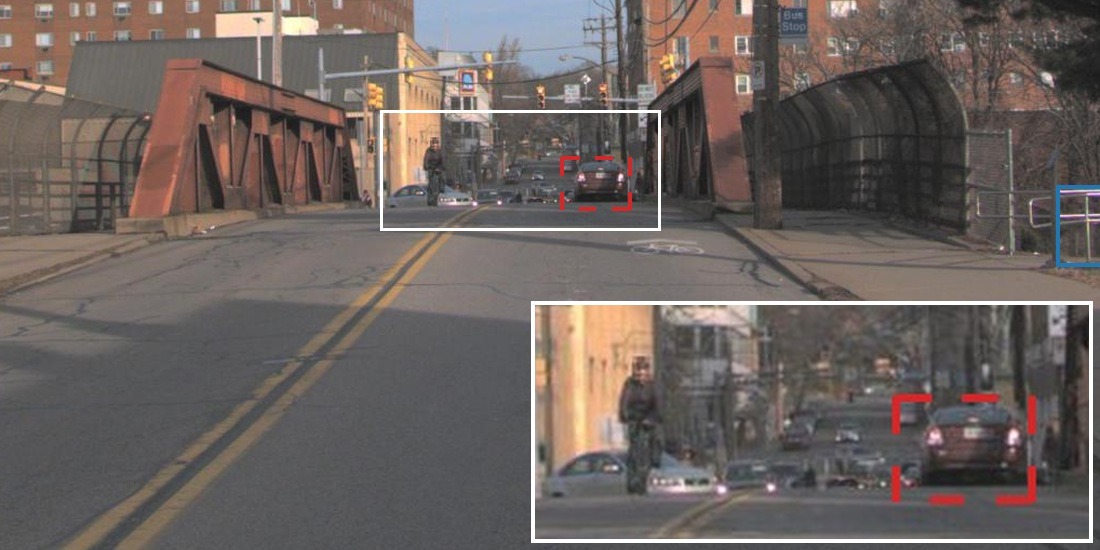}{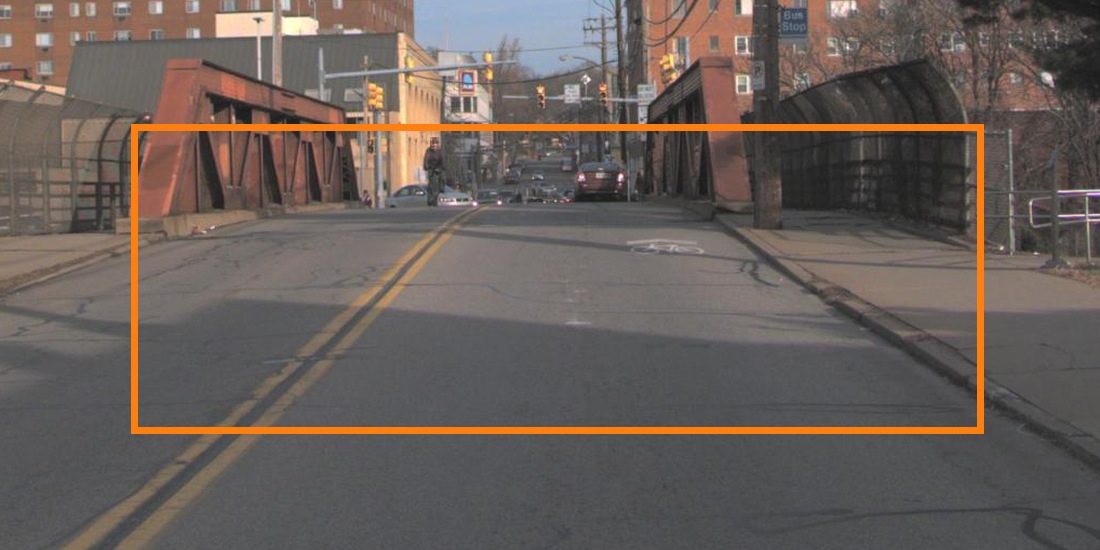}{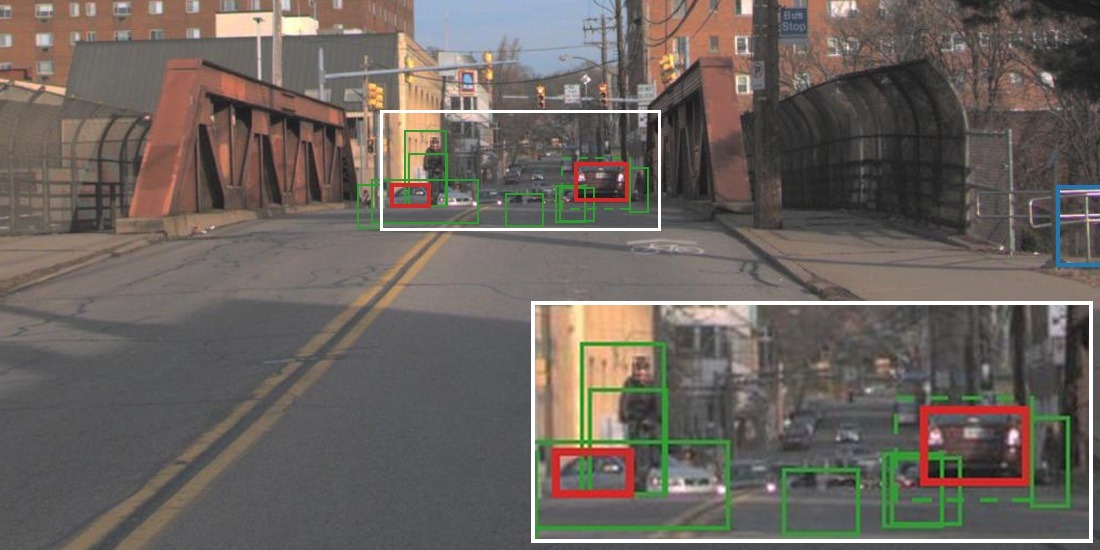}{The draft misses the vehicle ahead (zoomed) and keeps the speed. The field footprint covers the vehicle, the reference detector finds it in the crop, and \method decelerates as required.}
\qualkey{\\[2pt]\qlg{black!25, line width=0.6pt}{zoomed region (white frame) and its enlargement}}
\caption{\textbf{Qualitative results selected by the fixed rule} (test frames; the categories not shown in \cref{fig:qual}(b)). Images are cropped to the union of the crops and the decision-relevant objects.}
\label{S-fig:qual_n}
\end{figure}

\end{document}